\pdfoutput=1

\documentclass[12pt]{article}
\usepackage{a4wide}

\usepackage{amsfonts, amssymb, amsmath} 
\usepackage[english]{babel}
\usepackage{multirow}
\usepackage{bm} 
\usepackage{cancel}

\usepackage{caption}
\usepackage{natbib}

\usepackage{customCommands} 
\usepackage{eufrak}

\usepackage{tikz}
\renewcommand{\pm}{\mathbin{\tikz [x=1.4ex,y=1.4ex,line width=.1ex] \draw (0.0,0) -- (1.0,0) (0.5,0.1) -- (0.5,1.0) (0.0,0.55) -- (1.0,0.55);}}%

\renewcommand\Im{\operatorname{Im}}

\usepackage{empheq}
\newcommand*\widefbox[1]{\fbox{\hspace{1em}#1\hspace{1em}}}

\newcommand{\pv}{\bfp_v}
\renewcommand{\qv}{\bfq_v}
\newcommand{\qI}{\bfq_{{\,}_{\bf 1}}}
\newcommand{\q}{\,\bfq}
\newcommand{\dq}{\,\dot\bfq}
\newcommand{\ddq}{\,\ddot\bfq}
\newcommand{\dddq}{\,\dddot\bfq}
\newcommand{\ddddq}{\,\ddddot\bfq}
\newcommand{\w}{\,\bfomega}
\newcommand{\aw}{\,\ol\bfomega}
\newcommand{\dw}{\,\dot\bfomega}
\newcommand{\e}{&=}
\newcommand{\nw}{\norm{\bfomega}}
\newcommand{\Q}[1]{[#1]}
\newcommand{\QL}[1]{\Q{#1}_L}
\newcommand{\QR}[1]{\Q{#1}_R}
\newcommand{\bth}{\bftheta}
\newcommand{\dth}{\delta\bftheta}
\newcommand{\nth}{\norm{\bftheta}}
\newcommand{\te}{\triangleq}

\usepackage[	bookmarks,%
			colorlinks = true,%
			linkcolor = black,%
			citecolor = blue,%
			pdfauthor = {Joan\ Sola},%
			pdftitle = {Quaternion\ kinematics\ for\ the\ error-state\ KF},%
			pdftex
			]{hyperref} 
			
\title{Quaternion kinematics for the error-state Kalman filter}
\author{Joan Sol\`a}

\begin{document}
\maketitle

\tableofcontents



\begin{abstract}

This article is an exhaustive revision of concepts and formulas related to quaternions and rotations in 3D space, and their proper use in estimation engines such as the error-state Kalman filter.

The paper includes an in-depth study of the rotation group and its Lie structure, with formulations using both quaternions and rotation matrices.
It makes special attention in the definition of rotation perturbations, derivatives and integrals.
It provides numerous intuitions and geometrical interpretations to help the reader grasp the inner mechanisms of 3D rotation.

The whole material is used to devise precise formulations for error-state Kalman filters suited for real applications using integration of signals from an inertial measurement unit (IMU).

\end{abstract}


\section{Quaternion definition and properties}

\subsection{Definition of quaternion}

One introduction to the quaternion that I find particularly attractive  is given by the Cayley-Dickson construction: 
If we have two complex numbers $A=a+bi$ and $C=c+di$, then constructing $Q=A+Cj$ and defining $k\triangleq ij$ yields a number in the space of quaternions $\bbH$,
\begin{align}
Q = a + bi + cj + dk \in\bbH ~, \label{equ:ijkQuat}
\end{align}%
where $\{a,b,c,d\}\in\bbR$, and $\{i,j,k\}$ are three imaginary unit numbers defined so that
\begin{subequations}
\label{equ:quatAlgebra}
\begin{align}
i^2=j^2=k^2=ijk=-1~,
\end{align}%
from which we can derive
\begin{align}
ij = -ji = k ~, \quad jk=-kj=i~, \quad ki=-ik=j~.
\end{align}
\end{subequations}
From \eqRef{equ:ijkQuat} we see that we can embed complex numbers, and thus real and imaginary numbers, in the quaternion definition, in the sense that real, imaginary and complex numbers are indeed quaternions,
\begin{align}
Q = a \in \bbR \subset \bbH~,
\Quad 
Q=bi \in \bbI \subset \bbH~,
\Quad 
Q=a+bi \in \bbZ \subset \bbH~.
\end{align}
Likewise, and for the sake of completeness, we may define numbers in the tri-dimensional imaginary subspace of $\bbH$.
We refer to them as \emph{pure quaternions}, and may note $\bbH_p=\Im(\bbH)$ the space of pure quaternions,
\begin{align}
Q=bi+cj+dk \in\bbH_p \subset\bbH~.
\end{align}

It is noticeable that, while regular complex numbers of unit length $\bfz=e^{i\theta}$ can encode rotations in the 2D plane (with one complex product, $\bfx'=\bfz\tdot\bfx$), ``extended complex numbers" or quaternions of unit length $\bfq=e^{(u_xi+u_yj+u_zk)\theta/2}$ encode rotations in the 3D space (with a double quaternion product, $\bfx'=\bfq\otimes\bfx\otimes\bfq^*$, as we explain later in this document).

\bigskip

{\bf CAUTION:} Not all quaternion definitions are the same. 
Some authors write the products as $ib$ instead of $bi$, and therefore they get the property $k = ji = -ij$, which results in $ijk=1$ and a left-handed quaternion. 
Also, many authors place the real part at the end position, yielding $Q = ia + jb + kc + d$. 
These choices have no fundamental implications but make the whole formulation different in the details. 
Please refer to \secRef{sec:conventions} for further explanations and disambiguation.

\bigskip

{\bf CAUTION:} There are additional conventions that also make the formulation different in details. 
They concern the ``meaning'' or ``interpretation'' we give to the rotation operators, either rotating vectors or rotating reference frames --which, essentially, constitute opposite operations. 
Refer also to \secRef{sec:conventions} for further explanations and disambiguation.

\bigskip 

{\bf NOTE:} Among the different conventions exposed above, this document concentrates on the Hamilton convention, whose most remarkable property is the definition \eqRef{equ:quatAlgebra}. A proper and grounded disambiguation requires to first develop a significant amount of material; therefore, this disambiguation is relegated to the aforementioned \secRef{sec:conventions}.

\subsubsection{Alternative representations of the quaternion}
\label{sec:altQuat}

The real + imaginary notation $\{1,i,j,k\}$ is not always convenient for our purposes. 
Provided that the algebra \eqRef{equ:quatAlgebra} is used, a quaternion can be posed as a sum scalar + vector,
\begin{align}
Q=q_w+q_xi+q_yj+q_zk
\qquad
\Leftrightarrow
\qquad
Q = q_w + \qv~,
\end{align}
where $q_w$ is referred to as the \emph{real} or \emph{scalar} part, and $\qv=q_x i+q_y j+q_z k=(q_x,q_y,q_z)$ as the \emph{imaginary} or \emph{vector} part.\footnote{\label{ftn:quatComponents}Our choice for the $(w,x,y,z)$ subscripts notation comes from the fact that we are interested in the geometric properties of the quaternion in the 3D Cartesian space. 
Other texts often use alternative subscripts such as $(0,1,2,3)$ or $(1,i,j,k)$, perhaps better suited for mathematical interpretations.} 
It can be also defined as an ordered pair scalar-vector 
\begin{align}
Q = \langle q_w,\qv\rangle ~.
\end{align}
We mostly represent a quaternion $Q$ as a 4-vector $\bfq$~,
\begin{align}
\bfq \triangleq 
\begin{bmatrix}
q_w\\\qv
\end{bmatrix}=
\begin{bmatrix}
q_w\\q_x\\q_y\\q_z
\end{bmatrix}~,
\end{align}%
which allows us to use matrix algebra for operations involving quaternions.
At certain occasions, we may allow ourselves to mix notations by abusing of the sign ``$=$". Typical examples are \emph{real quaternions} and \emph{pure quaternions},
\begin{align}
\textrm{general: }
\bfq
=q_w+\qv=\begin{bmatrix}
q_w\\\qv
\end{bmatrix} \in \bbH
~,\quad
\textrm{real: }
q_w=\begin{bmatrix}
q_w\\{\bf0}_v
\end{bmatrix} \in \bbR
~,\quad
\textrm{pure: }
\qv=\begin{bmatrix}
0\\\qv
\end{bmatrix} \in \bbH_p
~.
\end{align}

\subsection{Main quaternion properties}

\subsubsection{Sum}

The sum is straightforward,
\begin{align}
\bfp\pm\bfq = \begin{bmatrix}
p_w \\ \pv
\end{bmatrix} \pm \begin{bmatrix}
q_w \\ \qv
\end{bmatrix}
  = \begin{bmatrix}
p_w \pm q_w \\ \pv \pm \qv
\end{bmatrix}~.
\end{align}
By construction, the sum is \textbf{commutative} and \textbf{associative},
\begin{align}
\bfp+\bfq&=\bfq+\bfp \\
\bfp+(\bfq+\bfr)&=(\bfp+\bfq)+\bfr
~.
\end{align}%
%

\subsubsection{Product}

Denoted by $\otimes$, the quaternion product requires using the original form \eqRef{equ:ijkQuat} and the quaternion algebra \eqRef{equ:quatAlgebra}. 
Writing the result in vector form gives
\begin{align}
\bfp\otimes\bfq = \begin{bmatrix}
p_wq_w - p_{x}q_{x} - p_{y}q_{y} - p_{z}q_{z} \\
p_wq_{x} + p_{x}q_w + p_{y}q_{z} - p_{z}q_{y} \\
p_wq_{y} - p_{x}q_{z} + p_{y}q_w + p_{z}q_{x} \\
p_wq_{z} + p_{x}q_{y} - p_{y}q_{x} + p_{z}q_w  
\end{bmatrix}~. \label{equ:quatProd}
\end{align}
This can be posed also in terms of the scalar and vector parts,
\begin{align}
\bfp\otimes\bfq = \begin{bmatrix}
p_wq_w - \pv\tr\qv \\
p_w\qv+q_w\pv+\pv\!\times\!\qv
\end{bmatrix}~, \label{equ:quatProdVec}
\end{align}
where 
the presence of the cross-product
reveals that the quaternion product is \textbf{not commutative} in the general case,
\begin{align}
\bfp\otimes\bfq\neq\bfq\otimes\bfp~.
\end{align}
Exceptions to this general non-commutativity are limited to the cases where $\pv\!\times\!\qv=0$, which happens whenever one quaternion is real, $\bfp=p_w$ or $\bfq=q_w$, or when both vector parts are parallel, $\pv \| \qv$. Only in these cases the quaternion product is commutative.

The quaternion product is however \textbf{associative},
\begin{align}
(\bfp\otimes\bfq)\ot\bfr = \bfp\otimes(\bfq\ot\bfr)~,
\end{align}
and \textbf{distributive over the sum},
\begin{align}
\bfp\ot(\bfq+\bfr) = \bfp\ot\bfq + \bfp\ot\bfr
\qquad \textrm{and} \qquad
(\bfp+\bfq)\ot\bfr = \bfp\ot\bfr + \bfq\ot\bfr~.
\end{align}

The product of two quaternions is bi-linear and can be expressed as two equivalent matrix products, namely
\begin{align}
\bfq_1\ot\bfq_2 = \QL{\bfq_1}\,\bfq_2 
\qquad \textrm{and} \qquad
\bfq_1\ot\bfq_2 = \QR{\bfq_2}\,\bfq_1 ~, \label{equ:quatMatProd}
\end{align}%
where $\QL{\bfq}$ and $\QR{\bfq}$ are respectively the left- and right- quaternion-product matrices, which are derived from \eqRef{equ:quatProd} and \eqRef{equ:quatMatProd} by simple inspection,
\begin{align}
\QL{\bfq} = \begin{bmatrix}
q_w &-q_x &-q_y &-q_z\\
q_x & q_w &-q_z & q_y\\
q_y & q_z & q_w &-q_x\\
q_z &-q_y & q_x & q_w\\
\end{bmatrix}, \qquad
\QR{\bfq} = \begin{bmatrix}
q_w &-q_x &-q_y &-q_z\\
q_x & q_w & q_z &-q_y\\
q_y &-q_z & q_w & q_x\\
q_z & q_y &-q_x & q_w\\
\end{bmatrix} ,
\label{equ:quatMatrixComponents}
\end{align}%
or more concisely, from \eqRef{equ:quatProdVec} and \eqRef{equ:quatMatProd}, 
\begin{align}
\QL{\bfq} = q_w\,\bfI + \begin{bmatrix}
0 & -\qv\tr \\
\qv & \hatx{\qv}
\end{bmatrix}, \qquad
\QR{\bfq} = q_w\,\bfI + \begin{bmatrix}
0 & -\qv\tr \\
\qv & -\hatx{\qv}
\end{bmatrix}~.
\label{equ:quatMatrix}
 ~
\end{align}%
Here, the \emph{skew operator}\footnote{The skew-operator can be found in the literature in a number of different names and notations, either related to the cross operator $\times$, or to the `hat' operator $^\wedge$, so that all the forms below are equivalent,
$$
\hatx{\bfa} \equiv [\bfa_\times] \equiv \bfa\!\times \equiv \bfa_\times \equiv [\bfa] \equiv \widehat{\bfa} \equiv \bfa^\wedge~.
$$
} 
$\hatx{\bullet}$ produces the cross-product matrix,
\begin{align}
\hatx{\bfa} \triangleq \begin{bmatrix}
0 & -a_z & a_y \\
a_z & 0 & -a_x \\
-a_y & a_x & 0
\end{bmatrix}
\label{equ:skew}
~,
\end{align}
which is a skew-symmetric matrix, $\hatx{\bfa}\tr=-\hatx{\bfa}$, equivalent to the cross product, \ie, 
\begin{align}
\hatx{\bfa}\bfb = \bfa\tcross\bfb~,\quad \forall\, \bfa,\bfb\in\bbR^3 ~.  
\end{align}

Finally, since
\begin{align*}
\bfq\ot\bfx\ot\bfp 
&= (\bfq\ot\bfx)\ot\bfp = \QR{\bfp}\,\QL{\bfq}\,\bfx 
\\
&= \bfq\ot(\bfx\ot\bfp) = \QL{\bfq}\,\QR{\bfp}\,\bfx
~,
\end{align*}
we have the relation
\begin{align}
\QR{\bfp}\,\QL{\bfq} = \QL{\bfq}\,\QR{\bfp}~,
\label{equ:PQ_commute}
\end{align}
that is, left- and right- quaternion product matrices commute.
Further properties of these matrices are provided in \secRef{sec:isoclinic}.

Quaternions endowed with the product operation $\otimes$ form a non-commutative group. 
The group's elements identity, $\bfq_1=1$, and inverse, $\bfq\inv$,  are explored below.

\subsubsection{Identity}

The identity quaternion $\qI$ \wrt the product is such that $\qI\otimes\bfq=\bfq\otimes\qI=\bfq$. 
It corresponds to the real product identity `1' expressed as a quaternion,
\begin{align*}
\qI = 1 = \begin{bmatrix}
1 \\ {\bf0}_v
\end{bmatrix} ~.
\end{align*}

\subsubsection{Conjugate}

The conjugate of a quaternion is defined by
\begin{align}
\bfq^*\triangleq q_w-\qv=\begin{bmatrix}
q_w \\ -\qv
\end{bmatrix}
~.
\end{align}
This has the properties
\begin{align}
\bfq\otimes\bfq^* = \bfq^*\otimes\bfq 
=q_w^2 +q_x^2 +q_y^2 +q_z^2
= \begin{bmatrix}
q_w^2 +q_x^2 +q_y^2 +q_z^2 \\ {\bf0}_v
\end{bmatrix}
~,
\end{align}
and
\begin{align}
(\bfp\ot\bfq)^* = \bfq^*\ot\bfp^* 
~.
\end{align}%

\subsubsection{Norm}

The norm of a quaternion is defined by
\begin{align}\label{equ:q_norm}
\norm{\bfq} \triangleq \sqrt{\bfq\otimes\bfq^*} = \sqrt{\bfq^*\otimes\bfq} = \sqrt{q_w^2 +q_x^2 +q_y^2 +q_z^2 } ~\in \bbR ~.
\end{align}
It has the property
\begin{align}
\norm{\bfp\ot\bfq} = \norm{\bfq\ot\bfp} = \norm{\bfp}\norm{\bfq}~. \label{equ:norm_prod}
\end{align}•

\subsubsection{Inverse}

The inverse quaternion $\bfq\inv$ is such that the quaternion times its inverse gives the identity,
\begin{align}
\bfq\otimes\bfq\inv = \bfq\inv\otimes\bfq = \qI~.
\end{align}
It can be computed with
\begin{align}
\bfq\inv = \bfq^*/\norm{\bfq}^2~.
\end{align}

\subsubsection{Unit or normalized quaternion}

For unit quaternions, $\norm{\bfq}=1$, and therefore
\begin{align}
\bfq\inv = \bfq^*~.
\end{align}

When interpreting the unit quaternion as an orientation specification, or as a rotation operator, this property implies that the inverse rotation can be accomplished with the conjugate quaternion. Unit quaternions can always be written in the form,
\begin{align}
\bfq = \begin{bmatrix}
\cos\theta \\ \bfu\sin\theta
\end{bmatrix}
~,
\end{align}
where $\bfu = u_x i + u_y j + u_z k$ is a unit vector and $\theta$ is a scalar. 

From \eqRef{equ:norm_prod}, unit quaternions endowed with the product operation $\otimes$ form a non commutative group, where the inverse coincides with the conjugate.

\subsection{Additional quaternion properties}

\subsubsection{Quaternion commutator}

The quaternion \emph{commutator} is defined as $[\bfp,\bfq]\triangleq\bfp\ot\bfq-\bfq\ot\bfp$. We have from \eqRef{equ:quatProdVec},
\begin{align}
\bfp\ot\bfq-\bfq\ot\bfp = 2\,\pv\tcross\qv
~.
\label{equ:quatCommutator}
\end{align}
This has as a trivial consequence,
\begin{align}
\pv\ot\qv-\qv\ot\pv = 2\,\pv\tcross\qv
~.
\label{equ:quatCommutatorPure}
\end{align}
We will use this property later on.

\subsubsection{Product of pure quaternions}

Pure quaternions are those with null real or scalar part, $Q=\qv$ or $\bfq=[0,\qv]$. We have from \eqRef{equ:quatProdVec},
\begin{align}
\pv\ot\qv 
= -\pv\tr\qv + \pv\tcross\qv
= \begin{bmatrix}
-\pv\tr\qv \\
\pv\tcross\qv
\end{bmatrix}~.
\label{equ:quatProdPure}
\end{align}
This implies
\begin{align}
\qv\ot\qv = -\qv\tr\qv = -\norm{\qv}^2~,
\label{equ:quatPureSquared}
\end{align}
and for pure unitary quaternions $\bfu\in\bbH_p,~ \norm{\bfu}=1$,
\begin{align}
\bfu\ot\bfu = -1
~,
\end{align}
which is analogous to the standard imaginary case, $i\cdot i=-1$.

\subsubsection{Natural powers of pure quaternions}

Let us define $\bfq^n, ~n\in\bbN$, as the $n$-th power of $\bfq$ using the quaternion product $\ot$. 
Then, if $\bfv$ is a pure quaternion and we let $\bfv=\bfu\,\theta$, with $\theta=\norm{\bfv}\in\bbR$ and $\bfu$ unitary, we get from \eqRef{equ:quatPureSquared} the cyclic pattern
\begin{align}
\bfv^2 = -\theta^2 \quad,\quad
\bfv^3 = -\bfu\,\theta^3 \quad,\quad
\bfv^4 = \theta^4 \quad,\quad
\bfv^5 = \bfu\,\theta^5 \quad,\quad
\bfv^6 = -\theta^6 \quad,\quad
\cdots
\label{equ:qvPowers}
\end{align}
and for pure unitary quaternions $\bfu$, this reduces to the pattern
\begin{align}
\bfu^2 = -1 	\quad,\quad
\bfu^3 = -\bfu 	\quad,\quad
\bfu^4 = 1 		\quad,\quad
\bfu^5 = \bfu 	\quad,\quad
\bfu^6 = -1 	\quad,\quad
\cdots
\label{equ:uPowers}
\end{align}

\subsubsection{Exponential of pure quaternions}

The quaternion exponential is a function on quaternions analogous to the ordinary exponential function. 
Exactly as in the real exponential case, it is defined as the absolutely convergent power series,
\begin{align}
e^\bfq
\triangleq \sum_{k=0}^\infty \frac{1}{k!}\bfq^k \quad \in \bbH
~.
\label{equ:quatExpSeries}
\end{align}
Clearly, the exponential of  a real quaternion coincides exactly with the ordinary exponential function. 

More interestingly,  the exponential of a pure quaternion $\bfv=v_xi+v_yj+v_zk$~ is a new quaternion defined by,
\begin{align}
e^\bfv
= \sum_{k=0}^\infty \frac{1}{k!}\bfv^k \quad \in \bbH
~.
\label{equ:pureQuatExpSeries}
\end{align}
Letting $\bfv=\bfu\,\theta$, with $\theta=\norm{\bfv}\in\bbR$ and $\bfu$ unitary, and considering \eqRef{equ:qvPowers}, we group the scalar and vector terms in the series, 
\begin{align}
e^{\bfu\theta} 
&= \left(1-\frac{\theta^2}{2!}+\frac{\theta^4}{4!}+\cdots\right) + \left(\bfu\theta - \frac{\bfu\theta^3}{3!}+\frac{\bfu\theta^5}{5!}+\cdots\right)
\end{align}
and recognize in them, respectively, the series of $\cos\theta$ and $\sin\theta$.%
\footnote{We remind that $\cos \theta = 1 - \theta^2/2! + \theta^4/4! - \cdots$, and $\sin \theta =  \theta - \theta^3/3! + \theta^5/5! - \cdots$.}
 This results in
\begin{align}
e^\bfv
= e^{\bfu\,\theta} 
= \cos\theta + \bfu\sin\theta = \begin{bmatrix}
\cos\theta \\ \bfu\sin\theta 
\end{bmatrix} ~,
\label{equ:EulerFormulaQuat}
\end{align}
which constitutes a beautiful extension of the Euler formula, $e^{i\theta}=\cos\theta+i\sin\theta$, defined for imaginary numbers. 
Notice that since $\norm{e^{\bfv}}^2=\cos^2\theta+\sin^2\theta=1$, the exponential of a pure quaternion is a unit quaternion.
Notice also the property,
\begin{align}
e^{-\bfv} = \left(e^{\bfv}\right)^*
~.
\end{align}

For small angle quaternions we avoid the division by zero in $\bfu=\bfv/\norm{\bfv}$ by expressing the Taylor series of $\sin\theta$ and $\cos\theta$ and truncating, obtaining varying degrees of the approximation,
\begin{align}
e^\bfv 
\approx
\begin{bmatrix}
1-\theta^2/2 \\ \bfv\big(1-\theta^2/6\big) 
\end{bmatrix}
\approx
\begin{bmatrix}
1 \\ \bfv 
\end{bmatrix}
\xrightarrow[\theta\to 0]{}
\begin{bmatrix}
1 \\ {\bf0}
\end{bmatrix}
~.
\end{align}

\subsubsection{Exponential of general quaternions}

Due to the non-commutativity property of the quaternion product, we cannot write for general quaternions $\bfp$ and $\bfq$ that $e^{\bfp+\bfq}=e^\bfp e^\bfq $. However, commutativity holds when any of the product members is a scalar, and therefore,
\begin{align}
e^\bfq = e^{q_w+\qv} = e^{q_w}\,e^{\qv} ~.
\end{align}
Then, using \eqRef{equ:EulerFormulaQuat} with $\bfu\theta=\qv$ we get
\begin{align}\label{equ:expGeneralQuat}
e^\bfq 
= e^{q_w}\begin{bmatrix}
\cos\norm{\qv} \\ \frac{\qv}{\norm{\qv}}\sin\norm{\qv} 
\end{bmatrix}~.
\end{align}

\subsubsection{Logarithm of unit quaternions}
\label{sec:qlog}

It is immediate to see that, if $\norm{\bfq}=1$,
\begin{align}\label{equ:qlog}
\log \bfq = \log (\cos \theta + \bfu \sin\theta) = \log(e^{\bfu\,\theta}) = \bfu\,\theta = \begin{bmatrix}
0 \\ \bfu\,\theta
\end{bmatrix}
~,
\end{align}
that is, the logarithm of a unit quaternion is a pure quaternion. The angle-axis values are obtained easily by inverting \eqRef{equ:EulerFormulaQuat},
\begin{align}
\bfu &= \qv / \norm{\qv} \\
\theta &= \arctan(\norm{\qv},q_w)
~.
\end{align}
For small angle quaternions, we avoid division by zero by expressing the Taylor series of $\arctan(x)$ and truncating,\footnote{We remind that $\arctan x = x - x^3/3 + x^5/5 - \cdots$, and $\arctan(y,x)\equiv\arctan(y/x)$.} obtaining varying degrees of the approximation,
\begin{align}
\log(\bfq) 
= \bfu\theta 
&= \qv\frac{\arctan({\norm{\qv},q_w})}{\norm{\qv}} 
\approx \frac{\qv}{q_w} \left(1 - \frac{\norm{\qv}^2}{3q_w^2}\right)
\approx \bfq_v
\xrightarrow[\theta\to 0]{}
{\bf 0}
~.
\end{align}
%

\subsubsection{Logarithm of general quaternions}

By extension, if $\bfq$ is a general quaternion,
\begin{align}
\log\bfq = \log(\norm{\bfq}\frac{\bfq}{\norm{\bfq}}) = \log\norm{\bfq} + \log\frac{\bfq}{\norm{\bfq}} = \log\norm{\bfq} + \bfu\,\theta = \begin{bmatrix}
\log\norm{\bfq} \\ \bfu\,\theta
\end{bmatrix}
~.
\end{align}

\subsubsection{Exponential forms of the type $\bfq^t$}

We have, for $\bfq\in\bbH$ and $t\in\bbR$,
\begin{align}
\bfq^t = \exp(\log(\bfq^t)) = \exp(t\log(\bfq))
~.
\end{align}
If $\norm{\bfq}=1$, we can write $\bfq=[\cos\theta,~\bfu\sin\theta]$, thus $\log(\bfq)=\bfu\theta$, which gives
\begin{align}\label{equ:qa}
\bfq^t = \exp(t\,\bfu\theta)=\begin{bmatrix}
\cos t\theta \\
\bfu\sin t\theta
\end{bmatrix}
~.
\end{align}
Because the exponent $t$ has ended up as a linear multiplier of the angle $\theta$, it can be seen as a linear angular interpolator. We will develop this idea in \secRef{sec:slerp}.

\section{Rotations and cross-relations}
\label{sec:rotations}

\subsection{The 3D vector rotation formula}

\begin{figure}[htbp]
\centering
\includegraphics{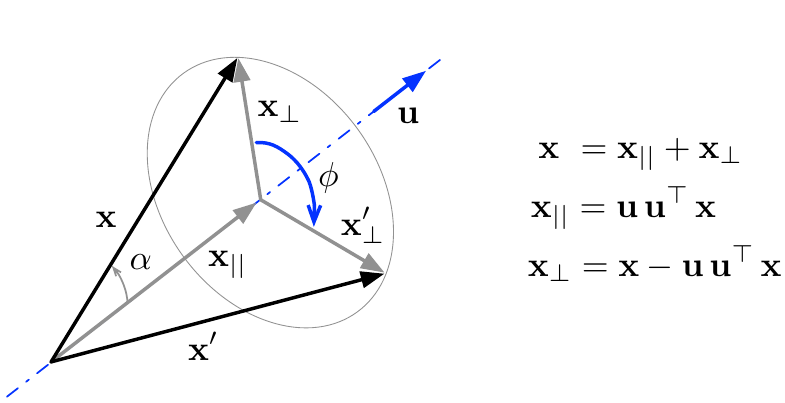}
\caption{Rotation of a vector $\bfx$, by an angle $\phi$, around the axis $\bfu$. See text for details.}
\label{fig:rotation3d}
\end{figure}

We illustrate in~\figRef{fig:rotation3d} the rotation, following the right-hand rule, of a general 3D vector $\bfx$, by an angle $\phi$, around the axis defined by the unit vector $\bfu$. 
This is accomplished by decomposing the vector $\bfx$ into a part $\bfx_{||}$ parallel to $\bfu$, and a part $\bfx_\bot$ orthogonal to $\bfu$, so that %
\begin{align*}
\bfx=\bfx_{||}+\bfx_\bot~. 
\end{align*}
These parts can be computed easily ($\alpha$ is the angle between the vector $\bfx$ and the axis $\bfu$),
\begin{align*}
\bfx_{||} &= \bfu \, (\norm{\bfx}\cos\alpha)  = \bfu\,\bfu\tr\,\bfx 
\\
\bfx_\bot &= \bfx - \bfx_{||} = \bfx - \bfu\,\bfu\tr\,\bfx~.
\end{align*}%
Upon rotation, the parallel part does not rotate, 
\begin{align*}
\bfx_{||}' = \bfx_{||}~,
\end{align*}
and the orthogonal part experiences a planar rotation in the plane normal to $\bfu$. That is, if we create an orthogonal base $\{\bfe_1,\bfe_2\}$ of this plane with
\begin{align*}
\bfe_1 &= \bfx_\bot \\
\bfe_2 &= \bfu \tcross \bfx_\bot = \bfu \tcross \bfx  ~, 
\end{align*}%
satisfying $\norm{\bfe_1} = \norm{\bfe_2}$, then $\bfx_\bot=\bfe_1\tdot1+\bfe_2\tdot0$. A rotation of $\phi$\,rad  on this plane produces,
\begin{align*}
\bfx_\bot' = \bfe_1\cos\phi + \bfe_2\sin\phi~,
\end{align*}
which develops as,
\begin{align*}
\bfx_\bot' = \bfx_\bot\cos\phi + (\bfu\tcross\bfx)\sin\phi~.
\end{align*}
Adding the parallel part yields the expression of the rotated vector, $\bfx'=
\bfx'_{||}+\bfx'_\bot$~, which is known as the \emph{vector rotation formula},
\begin{align}
\eqbox{
\bfx'=
\bfx_{||}+\bfx_\bot\cos\phi+(\bfu\times\bfx)\sin\phi}~.
\label{equ:vecRotFormula}
\end{align}
%

\subsection{The rotation group $SO(3)$}

In $\bbR^3$, the rotation group $SO(3)$ is the group of rotations around the origin under the operation of composition. Rotations are linear transformations that preserve vector length and relative vector orientation (\ie, handedness). Its importance in robotics is that it represents rotations of rigid bodies in 3D space: a \emph{rigid motion} requires precisely that distances, angles and relative orientations within a rigid body be preserved upon motion ---otherwise, if norms, angles or relative orientations are not kept, the body could not be considered rigid.

Let us then define rotations through an operator that satisfies these properties. 
A rotation operator $r:\bbR^3\to\bbR^3; \bfv\mapsto r(\bfv)$ acting on vectors $\bfv\in\bbR^3$ can be defined from the metrics of Euclidean space, constituted by the dot and cross products, as follows.
%
%
\begin{itemize}
\item Rotation preserves the vector norm,
\begin{subequations}
\begin{align}
\norm{r(\bfv)}=
\sqrt{\langle r(\bfv), r(\bfv)\rangle}=\sqrt{\langle\bfv,\bfv\rangle}\triangleq\norm{\bfv}~,\quad\forall\bfv \in\bbR^3~. \label{eq:keepnorm}
\end{align}
\item Rotation preserves angles between vectors, 
\begin{align}
\langle r(\bfv), r(\bfw)\rangle  = \langle \bfv,\bfw\rangle = \norm{\bfv}\norm{\bfw}\cos\alpha~,\quad\forall\bfv,\bfw\in\bbR^3~.
\end{align}
\end{subequations}
\item Rotation preserves the relative orientations of vectors,
\begin{align}
\bfu\times\bfv=\bfw 
\iff 
r(\bfu)\times r(\bfv)=r(\bfw) 
~. 
\label{equ:keeporientation}
\end{align}
\end{itemize}
It is easily proved that the first two conditions are equivalent. We can thus define the rotation group $SO(3)$ as,
%
\begin{align}
SO(3):\{r:\bbR^3\to\bbR^3\,/\,\forall\, \bfv,\bfw\in\bbR^3~,~ \norm{r(\bfv)}=\norm{\bfv}~,~ r(\bfv)\tcross r(\bfw)=r(\bfv\tcross\bfw)\} 
~.
\end{align}

The rotation group is typically represented by the set of rotation matrices. 
However, quaternions constitute also a good representation of it. 
The aim of this chapter is to show that both representations are equally valid. 
They exhibit a lot of similarities, both conceptually and algebraically, as the reader will appreciate in \tabRef{tab:Rq}.
\begin{table}[htp]
\renewcommand{\arraystretch}{1.5}
\begin{center}
\caption{The rotation matrix and the quaternion for representing $SO(3)$.}
\label{tab:Rq}
\begin{tabular}{|c|c|c|}
\hline
& Rotation matrix, $\bfR$ & Quaternion, $\bfq$ \\
\hline
\hline
Parameters & $3\times3=9$ & $1+3=4$ \\
Degrees of freedom & 3 & 3 \\
Constraints & $9-3=6$ & $4-3=1$ \\
Constraints & $\bfR\bfR\tr=\bfI~~;~~\det(\bfR)=+1$ & $\bfq\ot\bfq^* = 1$ \\
\hline
\hline
ODE & $\dot\bfR=\bfR\hatx{\bfomega}$ & $\dot\bfq=\frac12\bfq\ot\bfomega$ \\
Exponential map & $\bfR=\exp(\hatx{\bfu\phi})$ & $\bfq=\exp(\bfu\phi/2)$ \\
Logarithmic map & $\log(\bfR) = \hatx{\bfu\phi}$ & $\log(\bfq) = \bfu\phi/2$ \\
Relation to $SO(3)$ & Single cover & Double cover \\
\hline
\hline
Identity & $\bfI$ & $1$ \\
Inverse & $\bfR\tr$ & $\bfq^*$ \\
Composition & $\bfR_1\,\bfR_2$  & $\bfq_1\ot\bfq_2$ \\
\hline
Rotation operator & $\bfR = \bfI + \sin\phi\hatx{\bfu} + (1-\cos\phi)\hatx{\bfu}^2$ & $\bfq = \cos\phi/2 + \bfu\sin\phi/2$ \\
Rotation action & $\bfR\,\bfx$ & $\bfq\ot\bfx\ot\bfq^*$ \\
\hline
\multirow{3}{*}{Interpolation} & $\bfR^t=\bfI + \sin t\phi\hatx{\bfu} \!+ (1\!-\!\cos t\phi)\hatx{\bfu}^2$ & $\bfq^t=\cos t\phi/2+\bfu\sin t\phi/2$\\
 & $\bfR_1(\bfR_1\tr\bfR_2)^t$ & $\bfq_1\ot(\bfq_1^*\ot\bfq_2)^t$ \\
 & & $\bfq_1\frac{\sin((1-t)\Delta\theta)}{\sin(\Delta\theta)}+\bfq_2\frac{\sin(t\Delta\theta)}{\sin(\Delta\theta)}$ \\
\hline
\hline
Cross relations & \multicolumn{2}{|c|}
{$\begin{aligned}
\rule{0pt}{2.7ex} 
\bfR\{\bfq\} &= (q_w^2-\qv\tr\qv)\,\bfI + 2\,\qv\qv\tr + 2\,q_w\hatx{\qv} \\
\bfR\{-\bfq\}&=\bfR\{\bfq\} ~~~~~~~~~~~~~~~~~~~~\, \text{double cover} \\
\bfR\{1\}&=\bfI ~~~~~~~~~~~~~~~~~~~~~~~~~~~ \text{identity}\\
\bfR\{\bfq^*\}&=\bfR\{\bfq\}\tr ~~~~~~~~~~~~~~~~~~~ \text{inverse}\\
\bfR\{\bfq_1\ot\bfq_2\}&=\bfR\{\bfq_1\}\,\bfR\{\bfq_2\} ~~~~~~~~~~ \text{composition}\\
\bfR\{\bfq^t\} &= \bfR\{\bfq\}^t ~~~~~~~~~~~~~~~~~~~\, \text{interpolation} 
\end{aligned}$} \\
\hline
\end{tabular}
\end{center}
\end{table}%
Perhaps, the most important difference is that the unit quaternion group constitutes a double cover of $SO(3)$ (thus technically not being $SO(3)$ itself), something that is not critical in most of our applications.\footnote{The effect of the double cover needs to be considered when performing interpolation in the space of rotations. This is however easy, as we will see in \secRef{sec:slerp}.}
The table is inserted upfront for the sake of a rapid comparison and evaluation.
The rotation matrix and quaternion representations of $SO(3)$ are explored in the following sections.

\subsection{The rotation group and the rotation matrix}


The operator $r()$ is linear, since it is defined from the scalar and vector products, which are linear. 
It can therefore be represented by a matrix $\bfR\in\bbR^{3\times3}$, which produces rotations to vectors $\bfv\in\bbR^3$ through the matrix product,
\begin{align}
r(\bfv) = \bfR\,\bfv~.
\end{align}
Injecting it in \eqRef{eq:keepnorm}, using the dot product $\langle\bfa,\bfb\rangle=\bfa\tr\bfb$ and developing we have that for all $\bfv$,
\begin{align}
(\bfR\bfv)\tr(\bfR\bfv) = \bfv\tr\bfR\tr\bfR\bfv = \bfv\tr\bfv
~,
\end{align}
yielding the \emph{orthogonality} condition on $\bfR$,
\begin{align}
\eqbox{
\bfR\tr\bfR = \bfI = \bfR\,\bfR\tr
}
~. 
\label{equ:Rorthogonal}
\end{align}
The condition above is indeed a condition of orthogonality, since we can observe from it that, by writing $\bfR=[\bfr_1,\bfr_2,\bfr_3]$ and substituting above, the column vectors $\bfr_i$ of $\bfR$, with $i\in\{1,2,3\}$, are of unit length and orthogonal to each other,
\begin{align*}
\langle\bfr_i,\bfr_i\rangle &= \bfr_i\tr\bfr_i = 1 \\
\langle\bfr_i,\bfr_j\rangle &= \bfr_i\tr\bfr_j = 0 ~,\quad \textrm{if } i\neq j~.
\end{align*}
The set of transformations keeping vector norms and angles is for this reason called the \emph{Orthogonal group}, denoted $O(3)$. 
The orthogonal group includes rotations (which are rigid motions) and reflections (which are not rigid).
The notion of \emph{group} here means essentially (and informally) that the product of two orthogonal matrices is always an orthogonal matrix,%
\footnote{\label{ftn:O3}%
Let $\bfQ_1$ and $\bfQ_2$ be orthogonal,
and build $\bfQ=\bfQ_1\,\bfQ_2$.
Then $\bfQ\tr\bfQ=
\bfQ_2\tr\bfQ_1\tr\bfQ_1\bfQ_2=\bfQ_2\tr\bfI\bfQ_2=\bfI$.}
and that each orthogonal matrix admits an inverse.
In effect, the orthogonality condition \eqRef{equ:Rorthogonal} implies that the inverse rotation is achieved with the transposed matrix,
\begin{align}
\bfR\inv=\bfR\tr~.
\end{align}

Adding the relative orientations condition \eqRef{equ:keeporientation} guarantees rigid body motion (hence discarding reflections), and results in one additional constraint on $\bfR$,%
\footnote{Notice that reflections satisfy $|\bfR|=\det(\bfR)=-1$, and do not form a group since $|\bfR_1\bfR_2|=1\neq-1$.}
\begin{align}\label{equ:unitDet}
\eqbox{
\det(\bfR)=1
}
~.
\end{align}
Orthogonal matrices with positive unit determinant are commonly referred to as \emph{proper} or \emph{special}. The set of such \emph{special orthogonal matrices} is a subgroup of $O(3)$ named the \emph{Special Orthogonal group} $SO(3)$.
Being a group, the product of two rotation matrices is always a rotation matrix.%
\footnote{%
See footnote \ref{ftn:O3} for $O(3)$ and add this for $SO(3)$: let $|\bfR_1|=|\bfR_2|=1$, then $|\bfR_1\bfR_2|=|\bfR_1|\,|\bfR_2|=1$.}

\subsubsection{The exponential map}

The exponential map (and the logarithmic map, which we see in the next section) is a powerful mathematical tool for working in the rotational 3D space with ease and rigor. It represents the entrance door to a corpus of infinitesimal calculus suited for the rotational space. The exponential map allows us to properly define derivatives, perturbations, and velocities, and to manipulate them. It is therefore essential in estimation problems in the space of rotations or orientations.

Rotations constitute rigid motions. This rigidity implies that it is possible to define a continuous trajectory or \emph{path} $r(t)$ in $SO(3)$ that continuously rotates the rigid body from its initial orientation, $r(0)$, to its current orientation, $r(t)$.
Being continuous, it is legitimate to investigate the time-derivatives of such transformations.
We do so by deriving the properties \eqRef{equ:Rorthogonal} and \eqRef{equ:unitDet} that we have just seen.

First of all, we notice that it is impossible to continuously escape the unit determinant condition~\eqRef{equ:unitDet} while satisfying~\eqRef{equ:Rorthogonal}, because this would imply a jump of the determinant from $+1$ to $-1$.\footnote{Put otherwise: a rotation cannot become a reflection through a continuous transformation.}
Therefore we only need to investigate the time-derivative of the orthogonality condition \eqRef{equ:Rorthogonal}. This reads
\begin{align}
\dif{}{t}{(\bfR\tr\bfR)} = \dot\bfR\tr\bfR+\bfR\tr\dot\bfR = 0
~,
\end{align}
which results in
\begin{align}
\bfR\tr\dot\bfR 
= -(\bfR\tr\dot\bfR)\tr
~,
\end{align}
meaning that the matrix $\bfR\tr\dot\bfR$ is skew-symmetric (\ie, it is equal to the negative of its transpose). 
The set of skew-symmetric $3\times3$ matrices is denoted $\so(3)$, and receives the name of the \emph{Lie algebra} of $SO(3)$. 
Skew-symmetric $3\times3$ matrices have the form,
%
\begin{align}
\hatx{\bfomega} \triangleq \begin{bmatrix}
0 & -\omega_z & \omega_y \\
\omega_z & 0 & -\omega_x \\
-\omega_y & \omega_x & 0
\end{bmatrix}
~;
\end{align}
they have 3\,DOF, and correspond to cross-product matrices, as we introduced already in \eqRef{equ:skew}. This establishes a one-to-one mapping $\bfomega\in\bbR^3\leftrightarrow\hatx{\bfomega}\in\so(3)$.
%
Let us then take a vector $\bfomega=(\omega_x,\omega_y,\omega_z)\in\bfR^3$ and write
\begin{align}
\bfR\tr\dot\bfR = \hatx{\bfomega}
~.
\end{align}
%
This leads to the ordinary differential equation (ODE),
\begin{align}
\label{equ:Rdot}
\dot\bfR=\bfR\hatx{\bfomega}~.
\end{align}
Around the origin, we have $\bfR=\bfI$ and the equation above reduces to $\dot\bfR=\hatx{\bfomega}$. 
Thus, we can interpret the Lie algebra $\so(3)$ as the space of the derivatives of $r(t)$ at the origin;
it constitutes the \emph{tangent space} to $SO(3)$, or the \emph{velocity space}. 
Following these facts, we can very well call $\bfomega$ the vector of instantaneous angular velocities. 

If $\bfomega$ is constant, the differential equation above can be time-integrated as
\begin{align}
\bfR(t) = \bfR(0)\,e^{\hatx{\bfomega}t} = \bfR(0)\,e^{\hatx{\bfomega t}} 
\end{align}
where the exponential $e^{\hatx{x}}$ is defined by its Taylor series, as we see in the following section.
Since $\bfR(0)$ and $\bfR(t)$ are rotation matrices, then clearly $e^{\hatx{\bfomega t}}=\bfR(0)\tr\bfR(t)$ is a rotation matrix. 
Defining the vector $\bfphi\triangleq\bfomega\Dt$ as the rotation vector encoding the full rotation over the period $\Dt$, we have
\begin{align}
\eqbox{
\bfR = e^{\hatx{\bfphi}} \label{equ:vectomat}
}~.
\end{align}
This is known as the exponential map, an application from  $\so(3)$ to $SO(3)$,
\begin{align}
\exp: \so(3) \to SO(3) ~;~ \hatx{\bfphi} \mapsto \exp(\hatx{\bfphi})=e^{\hatx{\bfphi}}
~.
\end{align}

\subsubsection{The capitalized exponential map}

\begin{figure}[tb]
\begin{center}
\includegraphics{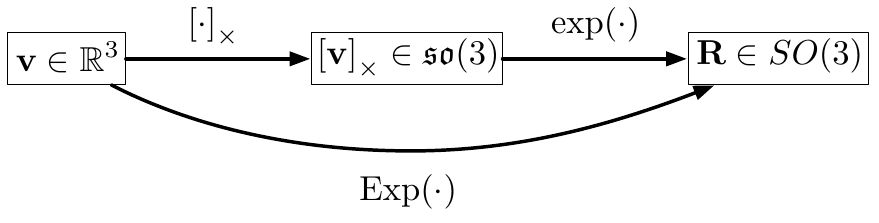}
\caption{Exponential maps of the rotation matrix.}
\label{fig:exp_map_R}
\end{center}
\end{figure}

The exponential map above is sometimes expressed with some abuse of notation, \ie, confounding $\bfphi\in\bbR^3$ with $\hatx{\bfphi}\in\so(3)$.
To avoid possible ambiguities, we opt for writing this new application $\bbR^3\to SO(3)$ with an explicit notation using a capitalized $\Exp$, having (see \figRef{fig:exp_map_R})
\begin{align}
\Exp: \bbR^3 \to SO(3) ~;~ \bfphi \mapsto \Exp(\bfphi) = e^{\hatx{\bfphi}}
~.
\end{align}
Its relation with the exponential map is trivial,
\begin{align}
\Exp(\bfphi) \triangleq \exp(\hatx{\bfphi})
~.
\end{align}

In the following sections we'll see that the vector $\bfphi$, called the rotation vector or the angle-axis vector, encodes through $\bfphi=\bfomega\Dt=\phi\bfu$ the angle $\phi$ and axis $\bfu$ of rotation.

\subsubsection{Rotation matrix and rotation vector: the Rodrigues rotation formula}

The rotation matrix is defined from the rotation vector $\bfphi=\phi\bfu$ through the exponential map \eqRef{equ:vectomat},
with the cross-product matrix $\hatx{\bfphi}=\phi\hatx{\bfu}$ as defined in \eqRef{equ:skew}.
The Taylor expansion of \eqRef{equ:vectomat} with $\bfphi=\phi\bfu$ reads, 
\begin{align}
\bfR=e^{\phi\hatx{\bfu}} = 
	  \bfI 
	+ 			\phi\hatx{\bfu} 
	+ \frac12	\phi^2\hatx{\bfu}^2
	+ \frac1{3!}\phi^3\hatx{\bfu}^3 
	+ \frac1{4!}\phi^4\hatx{\bfu}^4 
	+ \dots
\end{align}
When applied to unit vectors, $\bfu$, the matrix $\hatx{\bfu}$ satisfies
\begin{align}
\hatx{\bfu}^2 &= \bfu\bfu\tr-\bfI
\label{equ:prop1}
\\
\hatx{\bfu}^3 &= -\hatx{\bfu}
~, \label{equ:prop2}
\end{align}%
and thus all powers of $\hatx{\bfu}$ can be expressed in terms of $\hatx{\bfu}$ and $\hatx{\bfu}^2$ in a cyclic pattern,
\begin{align}
\hatx{\bfu}^4 &= -\hatx{\bfu}^2 
& \hatx{\bfu}^5 &= \hatx{\bfu} 
& \hatx{\bfu}^6 &= \hatx{\bfu}^2 
& \hatx{\bfu}^7 &=-\hatx{\bfu} 
~~\cdots 
~.
\end{align}
Then, grouping the Taylor series in terms of $\hatx{\bfu}$ and $\hatx{\bfu}^2$, and identifying in them, respectively, the series of $\sin\phi$ and $\cos\phi$, leads to a closed form to obtain the rotation matrix from the rotation vector, the so called \emph{Rodrigues rotation formula},
\begin{align}
\eqbox{
\bfR 
= \bfI + \sin\phi\hatx{\bfu} + (1-\cos\phi)\hatx{\bfu}^2
}~, \label{equ:rodrigues}
\end{align}%
which we denote $\bfR\{\bfphi\}\triangleq\Exp(\bfphi)$. 
This formula admits some variants, \eg, using \eqRef{equ:prop1},
\begin{align}
\bfR &= \bfI\cos\phi + \hatx{\bfu}\sin\phi + \bfu\bfu\tr(1-\cos\phi)
~.
\end{align}%

\subsubsection{The logarithmic maps}

We define the logarithmic map as the inverse of the exponential map,
\begin{align}
\log : SO(3)\to\so(3)~;~ \bfR \mapsto \log(\bfR)=\hatx{\bfu\,\phi}
~,
\end{align}
with
\begin{align}
\phi &= \arccos\left(\frac{\trace(\bfR)-1}{2}\right) 
\\
\bfu &= \frac{(\bfR-\bfR\tr)^\vee}{2\sin\phi} 
~,
\end{align}
where $\bullet^\vee$ is the inverse of $\hatx{\bullet}$, that is, $(\hatx{\bfv})^\vee=\bfv$ and $\hatx{\bfV^\vee}=\bfV$.

We also define a capitalized version $\Log$, which allows us to recover the rotation vector $\bfphi=\bfu\phi\in\bbR^3$ directly from the rotation matrix, 
\begin{subequations}
\begin{align}
\Log: SO(3) \to \bbR^3 ~;~ \bfR\mapsto\Log(\bfR) = \bfu\,\phi 
~.
\end{align}
\end{subequations}
Its relation with the logarithmic map is trivial,
\begin{align}
\Log(\bfR) \triangleq (\log(\bfR))^\vee
~.
\end{align}

\subsubsection{The rotation action}

Rotating a vector $\bfx$ by an angle $\phi$ around the unit axis $\bfu$ 
is performed with the linear product
\begin{align}
\bfx'=\bfR\,\bfx
~, 
\label{equ:rotWithMat}
\end{align}
where $\bfR=\Exp(\bfu\phi)$.
This can be shown by developing \eqRef{equ:rotWithMat}, 
using \eqRef{equ:rodrigues}, \eqRef{equ:prop1} and \eqRef{equ:prop2}, 
%
\begin{align}
\begin{split}
\bfx' &= \bfR\,\bfx  \\
&= (\bfI + \sin\phi\hatx{\bfu} + (1-\cos\phi)\hatx{\bfu}^2)\,\bfx  \\
&= \bfx + \sin\phi\hatx{\bfu}\bfx + (1-\cos\phi)\hatx{\bfu}^2\bfx  \\
&= \bfx + \sin\phi(\bfu\tcross\bfx) + (1-\cos\phi)(\bfu\bfu\tr-\bfI)\,\bfx  \\
&= \bfx_\| + \bfx_\bot + \sin\phi(\bfu\tcross\bfx) - (1-\cos\phi)\,\bfx_\bot  \\
&= \bfx_\| + (\bfu\tcross\bfx)\sin\phi + \bfx_\bot\cos\phi
~,
\end{split}
\end{align}%
which is precisely the vector rotation formula \eqRef{equ:vecRotFormula}.

\subsection{The rotation group and the quaternion}

For didactical purposes, we are interested in highlighting the connections between quaternions and rotation matrices as representations of the rotation group $SO(3)$. 
For this, the well-known formula of the quaternion rotation action, which reads,
%
\begin{align} \label{equ:qrot}
r(\bfv)=\bfq\ot\bfv\ot\bfq^*
~,
\end{align}
is here taken initially as an hypothesis.
This allows us to develop the full quaternion section with a discourse that retraces the one we used for the rotation matrix. 
The exactness of this hypothesis will be proved a little later, in \secRef{sec:qRotAction}, thus validating the approach. 

Let us then inject the rotation above into the orthogonality condition \eqRef{eq:keepnorm}, and develop it using \eqRef{equ:norm_prod} as
\begin{align}
\norm{\bfq\ot\bfv\ot\bfq^*}=\norm{\bfq}^2\norm{\bfv} = \norm{\bfv}
~.
\end{align}
This yields $\norm{\bfq}^2=1$, that is, the unit norm condition on the quaternion, which reads,
\begin{align} \label{equ:q_unit}
\eqbox{
\bfq^*\ot\bfq = 1 = \bfq\ot\bfq^*
}
~.
\end{align}
This condition is akin to the one we encountered for rotation matrices, see \eqRef{equ:Rorthogonal}, which reads $\bfR\tr\bfR=\bfI=\bfR\bfR\tr$. We encourage the reader to stop at their similarities for a second.

Similarly, we show that the relative orientation condition \eqRef{equ:keeporientation} is satisfied by construction (we use \eqRef{equ:quatCommutatorPure} twice, as indicated below),
\begin{align}
\begin{split}
r(\bfv)\times r(\bfw) 
&= (\bfq\ot\bfv\ot\bfq^*) \times (\bfq\ot\bfw\ot\bfq^*) \\
\eqRef{equ:quatCommutatorPure}~~
&= \frac12\big((\bfq\ot\bfv\ot\bfq^*) \ot (\bfq\ot\bfw\ot\bfq^*) - (\bfq\ot\bfw\ot\bfq^*) \ot (\bfq\ot\bfv\ot\bfq^*) \big) \\
&= \frac12(\bfq\ot\bfv\ot\bfw\ot\bfq^* - \bfq\ot\bfw\ot\bfv\ot\bfq^*) \\
&= \frac12(\bfq\ot(\bfv\ot\bfw - \bfw\ot\bfv)\ot\bfq^*) \\
\eqRef{equ:quatCommutatorPure}~~
&= \bfq\ot(\bfv\times\bfw)\ot\bfq^* \\
&= r(\bfv\times\bfw)
~.
\end{split}
\end{align}

The set of unit quaternions forms a group under the operation of multiplication. This group is topologically a 3-sphere, that is, the 3-dimensional surface of the unit sphere of $\bbR^4$, and is commonly noted as $S^3$. 

\subsubsection{The exponential map}

Let us consider a unit quaternion $\bfq\in S^3$, that is, $\bfq^*\ot\bfq=1$, and let us proceed as we did for the orthogonality condition of the rotation matrix, $\bfR\tr\bfR=\bfI$. 
Taking the time derivative,
\begin{align}
\dif{(\bfq^*\ot\bfq)}{t} = \dot\bfq^*\ot\bfq+\bfq^*\ot\dot\bfq=0
~,
\end{align}
it follows that
\begin{align}
\bfq^*\ot\dot\bfq = -(\dot\bfq^*\ot\bfq) = -(\bfq^*\ot\dot\bfq)^*~,
\end{align}
which means that $\bfq^*\ot\dot\bfq$ is a pure quaternion (\ie, it is equal to minus its conjugate, therefore its real part is zero). 
We thus take a pure quaternion $\bfOmega\in\bbH_p$ and write,
\begin{align}
\bfq^*\ot\dot\bfq = \bfOmega = \begin{bmatrix}
0\\\bfOmega
\end{bmatrix} 
\in\bbH_p
~.
\end{align}
Left-multiplication by $\bfq$ yields the differential equation,
\begin{align}
\label{equ:qdotOmega}
\dot\bfq = \bfq\ot\bfOmega~.
\end{align}
Around the origin, we have $\bfq=1$ and the equation above reduces to $\dot\bfq=\bfOmega\in\bbH_p$. 
Thus, the space $\bbH_p$ of pure quaternions constitutes the \emph{tangent space}, or the Lie Algebra, of the unit sphere $S^3$ of quaternions. 
In the quaternion case, however, this space is not directly the velocity space, but rather the space of the half-velocities, as we will see soon.

If $\bfOmega$ is constant, the differential equation can be integrated as
\begin{align}\label{equ:qexpWt}
\bfq(t) = \bfq(0)\ot e^{\bfOmega\, t}
~,
\end{align}
where, since $\bfq(0)$ and $\bfq(t)$ are unit quaternions, the exponential $e^{\bfOmega t}$ is also a unit quaternion ---something we already knew from the quaternion exponential~\eqRef{equ:EulerFormulaQuat}.
Defining $\bfV\triangleq\bfOmega\Dt$ we have
\begin{align}\label{equ:qexpV}
\eqbox{
\bfq = e^{\bfV}
}~.
\end{align}
This is again an exponential map: an application from the space of pure quaternions to the space of rotations represented by unit quaternions,
\begin{align}\label{equ:q_expmap}
\exp:\bbH_p\to S^3~;~ \bfV\mapsto \exp(\bfV) = e^{\bfV}
\end{align}

\subsubsection{The capitalized exponential map}

As we will see, the pure quaternion $\bfV$ in the exponential map~\eqRef{equ:q_expmap} encodes, through $\bfV=\theta\bfu = \phi\bfu/2$, the axis of rotation $\bfu$ and the half of the rotated angle, $\theta=\phi/2$. 
We will provide ample explanations to this half-angle fact very soon, mainly in Sections \ref{sec:qRotAction}, \ref{sec:double_cover} and \ref{sec:isoclinic}. By now, let it suffice to say that, since the rotation action is accomplished by the double product $\bfx'=\bfq\ot\bfx\ot\bfq^*$, the vector $\bfx$ experiences a rotation which is `twice' the one encoded in $\bfq$, or equivalently, the quaternion $\bfq$ encodes `half' the intended rotation on~$\bfx$.

\begin{figure}[tb]
\begin{center}
\includegraphics{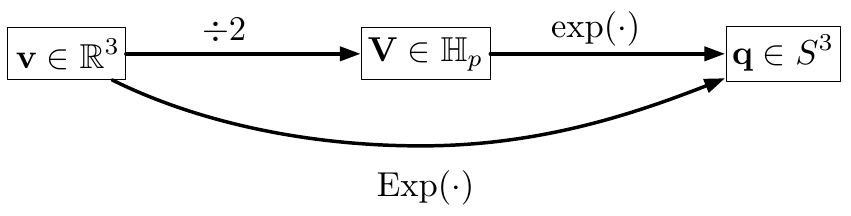}
\caption{Exponential maps of the quaternion.}
\label{fig:exp_map_q}
\end{center}
\end{figure}

In order to express a direct relation between the angle-axis rotation parameters, $\bfphi=\phi\bfu\in\bbR^3$, and the quaternion, we define a capitalized version of the exponential map, which captures the half-angle effect (see \figRef{fig:exp_map_q}),
\begin{align}
\Exp:\bfR^3\to S^3~;~\bfphi\mapsto\Exp(\bfphi)=e^{\bfphi/2}
\end{align}
Its relation to the exponential map is trivial,
\begin{align}
\Exp(\bfphi) \triangleq \exp(\bfphi/2)
~.
\end{align}

It is also convenient to introduce the vector of angular velocities $\bfomega=2\bfOmega\in\bbR^3$, so that \eqRef{equ:qdotOmega} and \eqRef{equ:qexpWt} become,
\begin{align}
\dot\bfq &= \frac12\bfq\ot\bfomega \label{equ:qdot} \\ 
\bfq &= e^{\bfomega t/2}
~.
\end{align}
%

\subsubsection{Quaternion and rotation vector}
\label{sec:quatAndVector}

Let $\bfphi=\phi\bfu$ be a rotation vector representing a rotation of $\phi$\,rad around the axis $\bfu$.
Then,
the exponential map can be developed using an extension of the \emph{Euler formula} (see \eqsRef{equ:qvPowers}{equ:EulerFormulaQuat} for a complete development),
\begin{align}
\eqbox{
\bfq \triangleq \Exp(\phi\bfu) = e^{\phi\bfu/2} = \cos \frac{\phi}{2} + \bfu\sin\frac{\phi}{2}=\begin{bmatrix}
\cos(\phi/2) \\
\bfu\sin(\phi/2)
\end{bmatrix}
}
~.   \label{equ:vectoquat}
\end{align}
We call this the \emph{rotation vector to quaternion} conversion formula, and will be denoted in this document by 
$\bfq=\bfq\{\bfphi\}\triangleq\Exp(\bfphi)$.

\subsubsection{The logarithmic maps}

We define the logarithmic map as the inverse of the exponential map,
\begin{align}
\log:S^3\to\bbH_p ~;~ \bfq\mapsto \log(\bfq) = \bfu\theta
~,
\end{align}
which is of course the definition we gave for the quaternion logarithm in \secRef{sec:qlog}.

We also define the capitalized logarithmic map, which directly provides the angle $\phi$ and axis $\bfu$ of rotation in Cartesian 3-space,
\begin{align}
\Log:S^3\to\bbR^3 ~;~ \bfq\mapsto \Log(\bfq) = \bfu\phi
~.
\end{align}
Its relation with the logarithmic map is trivial,
\begin{align}
\Log (\bfq) \triangleq 2\log(\bfq)
~.
\end{align}

For its implementation we use the 4-quadrant version of $\arctan(y,x)$. 
From \eqRef{equ:vectoquat},
\begin{subequations}
\begin{align}
\phi &= 2\arctan(\norm{\qv},q_w) \\
\bfu &= \qv / \norm{\qv} \label{equ:qvec}
~.
\end{align}
\end{subequations}
For small-angle quaternions, \eqRef{equ:qvec} diverges. We then use the a truncated Taylor series for the $\arctan()$ function, getting,
\begin{equation}
\Log(\bfq) = \theta\bfu 
\approx 2\,\frac{\qv}{q_w} \left(1 - \frac{\norm{\qv}^2}{3q_w^2}\right) \label{equ:log_q_small}
~.
\end{equation}

\subsubsection{The rotation action}
\label{sec:qRotAction}

We are finally in the position of proving our hypothesis~\eqRef{equ:qrot} for the vector rotation  using quaternions,  
thus validating all the material presented so far.
%
Rotating a vector $\bfx$ by an angle $\phi$ around the axis $\bfu$ is performed with the double quaternion product, also known as the sandwich product,
\begin{align}
\bfx' = \bfq\otimes\bfx\otimes\bfq^* ~, \label{equ:sandwichProd}
\end{align}
where $\bfq=\Exp(\bfu\phi)$, and
where the vector $\bfx$ has been written in quaternion form, that is, 
\begin{align}
\bfx= x i + y j + z k = \begin{bmatrix}
0 \\ \bfx
\end{bmatrix} \in \bbH_p
~. \label{equ:quatvec}
\end{align}%
To show that this double product does perform the desired vector rotation, we use \eqRef{equ:quatProdVec}, 
\eqRef{equ:vectoquat}, and basic vector and trigonometric identities, to develop~\eqRef{equ:sandwichProd} as follows,
\begin{align}\label{equ:quatRotFormula}
\begin{split}
\bfx'
&= \bfq \ot \bfx \ot \bfq^* \\
&= \Big(\cos \frac{\phi}{2} + \bfu \sin \frac{\phi}{2}\Big)
 \ot (0+\bfx)
 \ot \Big(\cos \frac{\phi}{2} - \bfu \sin \frac{\phi}{2}\Big)
 \\
&= \bfx \cos^2 \frac{\phi}{2} + (\bfu\ot\bfx - \bfx\ot\bfu) \sin \frac{\phi}{2} \cos \frac{\phi}{2} - \bfu\ot\bfx\ot\bfu \sin^2 \frac{\phi}{2} \\
&= \bfx \cos^2 \frac{\phi}{2} + 2 (\bfu \tcross \bfx) \sin \frac{\phi}{2} \cos \frac{\phi}{2} - (\bfx (\bfu \tr \bfu) - 2 \bfu (\bfu \tr \bfx)) \sin^2 \frac{\phi}{2} \\
&= \bfx (\cos^2 \frac{\phi}{2} - \sin^2 \frac{\phi}{2}) + (\bfu \tcross \bfx) (2\sin\frac{\phi}{2} \cos\frac{\phi}{2}) + \bfu (\bfu \tr \bfx) (2\sin^2 \frac{\phi}{2}) \\
&= \bfx \cos \phi + (\bfu \tcross \bfx) \sin \phi + \bfu (\bfu \tr \bfx) (1 - \cos \phi) \\
&= (\bfx - \bfu \,\bfu \tr \bfx) \cos \phi + (\bfu \tcross \bfx) \sin \phi + \bfu \,\bfu \tr \bfx \\
&= \bfx_{\bot} \cos \phi + (\bfu \tcross \bfx) \sin \phi + \bfx_{||} ~,
\end{split}
\end{align}%
which is precisely the vector rotation formula~\eqRef{equ:vecRotFormula}.

\subsubsection{The double cover of the manifold of $SO(3)$.}
\label{sec:double_cover}

Consider a unit quaternion $\bfq$. When regarded as a regular 4-vector, the angle $\theta$ between $\bfq$ and the identity quaternion $\bfq_1=[1,0,0,0]$ representing the origin of orientations is,
\begin{align}
\cos\theta = \bfq_1\tr\bfq = \bfq(1) = q_w
~.
\end{align}
At the same time, the angle $\phi$ rotated by the quaternion $\bfq$ on objects in 3D space satisfies
\begin{align}
\bfq = \begin{bmatrix}
q_w \\ \qv
\end{bmatrix} = \begin{bmatrix}
\cos\phi/2 \\ \bfu\sin\phi/2
\end{bmatrix}
~.
\end{align}
That is, we have $q_w = \cos\theta = \cos\phi/2$, 
so the angle between a quaternion vector and the identity in 4D space is half the angle rotated by the quaternion in 3D space,
\begin{align}
\theta = \phi/2
~.
\end{align}

\begin{figure}[htbp]
\begin{center}
\includegraphics{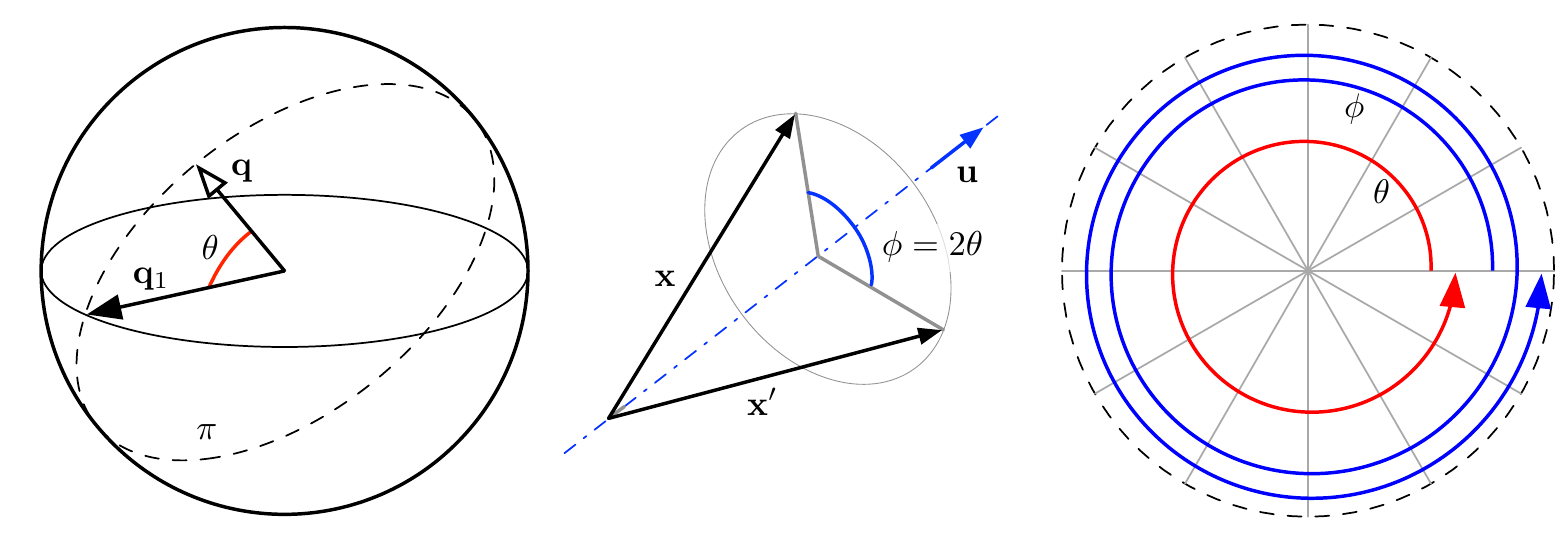}
\caption{Double cover of the rotation manifold. Left: the quaternion $\bfq$ in the unit 3-sphere defines an angle $\theta$ with the identity quaternion $\bfq_1$. Center: the resulting 3D rotation $\bfx'=\bfq\ot\bfx\ot\bfq^*$ has double angle $\phi$ than that of the original quaternion. Right: Superposing the 4D and 3D rotation planes, observe how one turn of the quaternion $\bfq$ over the 3-sphere (red) represents two turns of the rotated vector $\bfx$ in 3D space (blue).}
\label{fig:double_cover}
\end{center}
\end{figure}

We illustrate this double cover in \figRef{fig:double_cover}. 
By the time the angle between the two quaternion vectors is $\theta=\pi/2$, the 3D rotation has already achieved $\phi=\pi$, which is half a turn. 
And by the time the quaternion vector has made a half turn, $\theta=\pi$, the 3D rotation has completed a full turn. 
The second half turn of the quaternion vector, $\pi<\theta<2\pi$, represents a second full turn of the 3D rotation, $2\pi<\phi<4\pi$, that is, a second cover of the rotation manifold.

\subsection{Rotation matrix and quaternion}

As we have just seen, given a rotation vector $\bfphi=\bfu\,\phi$, the exponential maps for 
the unit quaternion 
and 
the rotation matrix 
produce rotation operators 
$\bfq=\Exp(\bfu\,\phi)$ 
and 
$\bfR=\Exp(\bfu\,\phi)$ 
that rotate vectors $\bfx$ exactly the same angle $\phi$ around the same axis $\bfu$.%
\footnote{The obvious notation ambiguity between the exponential maps $\bfR=\Exp(\bfphi)$ and $\bfq = \Exp(\bfphi)$ is easily resolved by the context: 
at occasions it is just the type of the returned value, $\bfR$ or $\bfq$; 
other times it is the presence or absence of the quaternion product $\ot$.}
That is, if
\begin{align}
\forall \bfphi,\bfx \in \bbR^3,~ 
\bfq = \Exp(\bfphi) 
,~ 
\bfR=\Exp(\bfphi)
\end{align}
%
%
%
%
then,
\begin{align}
\bfq\otimes\bfx\otimes\bfq^* = \bfR\,\bfx~.
\end{align}
As both sides of this identity are linear in $\bfx$, an expression of the rotation matrix equivalent to the quaternion is found by developing the left hand side and identifying terms on the right, yielding the \emph{quaternion to rotation matrix} formula,
\begin{align}
\eqbox{
\bfR = \begin{bmatrix}
q_w^2+q_x^2-q_y^2-q_z^2 & 2(q_xq_y-q_wq_z) & 2(q_xq_z+q_wq_y) \\ 
2(q_xq_y+q_wq_z) & q_w^2-q_x^2+q_y^2-q_z^2 & 2(q_yq_z-q_wq_x) \\
2(q_xq_z-q_wq_y) & 2(q_yq_z+q_wq_x) & q_w^2-q_x^2-q_y^2+q_z^2
\end{bmatrix}
}~,
\end{align}%
denoted throughout this document by $\bfR=\bfR\{\bfq\}$. 
The matrix form of the quaternion product \eqsRef{equ:quatMatProd}{equ:quatMatrix} provides us with an alternative formula
, since
\begin{align}
\bfq\otimes
\bfx\otimes\bfq^*
&= \QR{\bfq^*}\,\QL{\bfq}\begin{bmatrix}
0 \\ \bfx
\end{bmatrix} 
= \begin{bmatrix}
0 \\ \bfR\,\bfx
\end{bmatrix} 
\label{equ:quatRotMatrixForm}
~,
\end{align}
which leads after some easy developments to
\begin{align}
\eqbox{\bfR = (q_w^2-\qv\tr\qv)\,\bfI + 2\,\qv\qv\tr + 2\,q_w\hatx{\qv}}~.
\end{align}

The rotation matrix $\bfR$ has the following properties \wrt the quaternion,
\begin{align}
\bfR\{[1,0,0,0]\tr\} &= \bfI \label{equ:rotident}\\
\bfR\{-\bfq\} &= \bfR\{\bfq\} \label{equ:rotneg} \\
\bfR\{\bfq^*\} &= \bfR\{\bfq\}\tr \label{equ:rotconj} \\
\bfR\{\bfq_1\ot\bfq_2\} &= \bfR\{\bfq_1\}\bfR\{\bfq_2\} \label{equ:rotprod}
~, 
\end{align}%
where we observe that: 
\eqRef{equ:rotident}~the identity quaternion encodes the null rotation;  
\eqRef{equ:rotneg}~a quaternion and its negative encode the same rotation, defining a double cover of $SO(3)$; 
\eqRef{equ:rotconj}~the conjugate quaternion encodes the inverse rotation; and  
\eqRef{equ:rotprod}~the quaternion product composes consecutive rotations in the same order as rotation matrices do. 

Additionally, we have the property
\begin{align}
\bfR\{\bfq^t\}=\bfR\{\bfq\}^t
~,
\end{align}
which relates the spherical interpolations  of the quaternion and rotation matrix over a running scalar $t$.

\subsection{Rotation composition}

\begin{figure}[htbp]
\begin{center}
\includegraphics{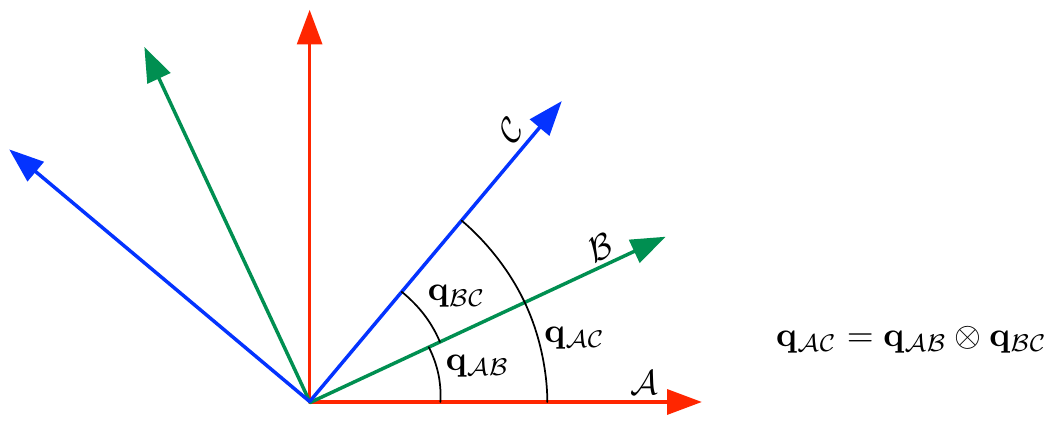}
\caption{Rotation composition. In $\bbR^2$, we would simply do $\theta_{\cA\cC} = \theta_{\cA\cB}+\theta_{\cB\cC}$, with an operation `sum' that is commutative. 
In $\bbR^3$ composition satisfies $\bfq_{\cA\cC} = \bfq_{\cA\cB}\ot\bfq_{\cB\cC}$ and, in matrix form, $\bfR_{\cA\cC} = \bfR_{\cA\cB}\,\bfR_{\cB\cC}$. 
These operators are not commutative and one must respect the order strictly ---a proper notation helps: `AB' chains with `BC' to create `AC'.}
\label{fig:composition}
\end{center}
\end{figure}

Quaternion composition is done similarly to rotation matrices, \ie, with appropriate quaternion- and matrix- products, and in the same order (\figRef{fig:composition}),
\begin{align}
\bfq_{\cA\cC} &= \bfq_{\cA\cB}\ot\bfq_{\cB\cC} ~,
&
\bfR_{\cA\cC} &= \bfR_{\cA\cB}\,\bfR_{\cB\cC} ~.\label{equ:rotComposition}
\end{align}%
%
%
%
This comes immediately from the associative property of the involved products,
\begin{align*}
\bfx_\cA 
&= \bfq_{\cA\cB}\ot\bfx_\cB\ot\bfq_{\cA\cB}^* 
& \bfx_\cA
&= \bfR_{\cA\cB}\,\bfx_\cB
\\
&= \bfq_{\cA\cB}\ot(\bfq_{\cB\cC}\ot\bfx_\cC\ot\bfq_{\cB\cC}^*)\ot\bfq_{\cA\cB}^* 
&&= \bfR_{\cA\cB}\,(\bfR_{\cB\cC}\,\bfx_\cC) 
\\
&= (\bfq_{\cA\cB}\ot\bfq_{\cB\cC})\ot\bfx_\cC\ot(\bfq_{\cB\cC}^*\ot\bfq_{\cA\cB}^*) 
&&= (\bfR_{\cA\cB}\,\bfR_{\cB\cC})\,\bfx_\cC 
\\
&= (\bfq_{\cA\cB}\ot\bfq_{\cB\cC})\ot\bfx_\cC\ot(\bfq_{\cA\cB}\ot\bfq_{\cB\cC})^* 
&&= \bfR_{\cA\cC}\,\bfx_\cC ~.
\\
&= \bfq_{\cA\cC}\ot\bfx_\cC\ot\bfq_{\cA\cC}^* 
~,
\end{align*}

\paragraph{A comment on notation}
A proper notation helps determining the right order of the factors in the composition, especially for compositions of several rotations (see \figRef{fig:composition}).
For example, let $\bfq_{ji}$ (resp. $\bfR_{ji}$) represent a rotation from situation $i$ to situation $j$, that is, $\bfx_j=\bfq_{ji}\ot\bfx_i\ot\bfq_{ji}^*$ (resp. $\bfx_j=\bfR_{ji}\bfx_i$).
Then, given a number of rotations represented by the quaternions $\bfq_{OA},\bfq_{AB},\bfq_{BC},\bfq_{OX},\bfq_{XZ}$, we just have to chain the indices and get:
\begin{align*}
\bfq_{OC} &= \bfq_{OA}\ot\bfq_{AB}\ot\bfq_{BC}
&
\bfR_{OC} &= \bfR_{OA}\,\bfR_{AB}\,\bfR_{BC}
~,
\end{align*}
and knowing that the opposite rotation corresponds to the conjugate, $\bfq_{ji}=\bfq_{ij}^*$, or  the transpose, $\bfR_{ji}=\bfR_{ij}\tr$, we also have
\begin{align*}
\bfq_{ZA} &= \bfq_{XZ}^*\ot\bfq_{OX}^*\ot\bfq_{OA} 
&
\bfR_{ZA} &= \bfR_{XZ}\tr\,\bfR_{OX}\tr\,\bfR_{OA} 
\\
&= \bfq_{ZX}\ot\bfq_{XO}\ot\bfq_{OA}
&
&= \bfR_{ZX}\,\bfR_{XO}\,\bfR_{OA}
~.
\end{align*}

\subsection{Spherical linear interpolation (SLERP)}
\label{sec:slerp}

Quaternions are very handy for computing proper orientation interpolations. 
Given two orientations represented by quaternions $\bfq_0$ and $\bfq_1$, we want to find a quaternion function $\bfq(t),~ t\in[0,1]$, that linearly interpolates from $\bfq(0)=\bfq_0$ to $\bfq(1)=\bfq_1$. 
This interpolation is such that, as $t$ evolves from $0$ to $1$, a body will continuously rotate from orientation $\bfq_0$ to orientation $\bfq_1$, at constant speed along a fixed axis.

\paragraph{Method 1}
A first approach uses quaternion algebra, and follows a geometric reasoning in $\bbR^3$ that should be easily related to the material presented so far.
First, compute the orientation increment $\Delta\bfq$ from $\bfq_0$ to $\bfq_1$ such that $\bfq_1=\bfq_0\ot\Delta\bfq$,
\begin{align}
\Delta\bfq = \bfq_0^*\ot\bfq_1
~.
\end{align}
Then obtain the associated rotation vector, $\Delta\bfphi=\bfu\Delta\phi$,
using the logarithmic map,\footnote{We can use here either the maps $\log()$ and $\exp()$, or their capitalized forms $\Log()$ and $\Exp()$. The involved factor 2 in the resulting angles is finally irrelevant as it cancels out in the final formula.}
\begin{align}\label{equ:LogDq}
\bfu\,\Delta\phi = \Log(\Delta\bfq)
~.
\end{align}
Finally, keep the rotation axis $\bfu$ and take a linear fraction of the rotation angle, $\delta\phi=t\Delta\phi$. 
Put it in quaternion form through the exponential map, $\delta\bfq=\Exp(\bfu\,\delta\phi)$, and compose it with the original quaternion to get the interpolated result,
\begin{align}
\bfq(t) = \bfq_0\ot\Exp(t\,\bfu\,\Delta\phi)
~.
\end{align}
The whole process can be written as $\bfq(t)=\bfq_0\ot \Exp(t\Log(\bfq_0^*\ot\bfq_1))$, which reduces to
\begin{align}
\eqbox{
\bfq(t)=\bfq_0\ot(\bfq_0^*\ot\bfq_1)^t
}
~,
\end{align}
and which is usually implemented (see \eqRef{equ:qa}) as,
\begin{align}
\bfq(t)=\bfq_0\ot
\begin{bmatrix}
\cos (t\,\Delta\phi/2) \\ \bfu \sin (t\,\Delta\phi/2)
\end{bmatrix}
~.
\end{align}

\begin{figure}[htbp]
\begin{center}
\includegraphics{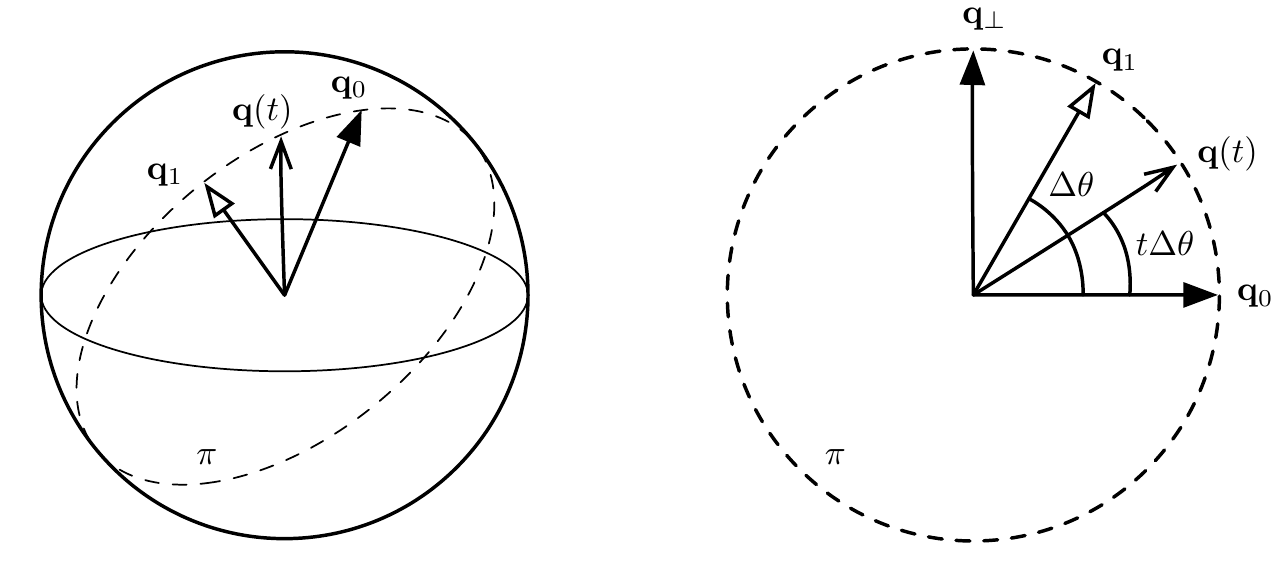}
\caption{Quaternion interpolation in the unit sphere of $\bbR^4$, and a frontal view of the situation on the rotation plane $\pi$ of $\bbR^4$.}
\label{fig:slerp_S4}
\end{center}
\end{figure}

\paragraph{Note:} An analogous procedure may be used to define Slerp for rotation matrices, yielding
\begin{align}\label{equ:Rslerp}
\bfR(t)= \bfR_0\Exp(t\Log(\bfR_0\tr\bfR_1))=\bfR_0(\bfR_0\tr\bfR_1)^t 
~,
\end{align}
where the matrix exponential $\bfR^t$ can be implemented using Rodrigues \eqRef{equ:rodrigues}, leading to
\begin{align}
\bfR(t)= \bfR_0 \left(\bfI + \sin(t\Delta\phi)\hatx{\bfu} + (1-\cos(t\Delta\phi))\hatx{\bfu}^2\right)
~.
\end{align}

\paragraph{Method 2}
Other approaches to Slerp can be developed that are independent of the inners of quaternion algebra, and even independent of the dimension of the space in which the arc is embedded. 
In particular, see \figRef{fig:slerp_S4}, we can treat quaternions $\bfq_0$ and $\bfq_1$ as two unit vectors in the unit sphere, and interpolate in this same space. 
The interpolated $\bfq(t)$ is the unit vector that follows at a constant angular speed
the shortest spherical path joining $\bfq_0$ to $\bfq_1$.
This path is the planar arc resulting from intersecting the unit sphere with the plane defined by $\bfq_0$, $\bfq_1$ and the origin (dashed circumference in the figure).
For a proof that these approaches are equivalent to the above, see \cite{DAM-1998}.

The first of these approaches uses vector algebra and follows literally the ideas above. 
Consider $\bfq_0$ and $\bfq_1$ as two unit vectors; 
%
the angle%
\footnote{The angle $\Delta\theta=\arccos(\bfq_0\tr\bfq_1)$ is the angle between the two quaternion vectors in Euclidean 4-space, not the real rotated angle in 3D space, which from \eqRef{equ:LogDq} is $\Delta\phi=\norm{\Log(\bfq_0^*\ot\bfq_1)}$. See \secRef{sec:double_cover} for further details.}
between them is derived from the scalar product,
\begin{align}\label{equ:slerp_angle}
\cos(\Delta\theta)&=\bfq_0\tr\bfq_1 & \Delta\theta&=\arccos(\bfq_0\tr\bfq_1)
~.
\end{align}
We proceed as follows. 
We identify the plane of rotation, that we name here $\pi$, 
and build its ortho-normal basis $\{\bfq_0,\bfq_\bot\}$, where $\bfq_\bot$ comes from ortho-normalizing $\bfq_1$ against $\bfq_0$,
\begin{align}
\bfq_\bot &= \frac{\bfq_1-(\bfq_0\tr\bfq_1)\bfq_0}{\norm{\bfq_1-(\bfq_0\tr\bfq_1)\bfq_0}}
~,
\end{align}
so that (see \figRef{fig:slerp_S4} -- right)
\begin{align} \label{equ:q1}
\bfq_1 = \bfq_0 \cos \Delta\theta + \bfq_\bot \sin \Delta\theta
~.
\end{align}
Then, we just need to rotate $\bfq_0$ a fraction of the angle, $t\Delta\theta$, over the plane $\pi$,
to yield the spherical interpolation,
\begin{align}\label{equ:slerp_rot}
\eqbox{
\bfq(t) = \bfq_0 \cos(t\Delta\theta) + \bfq_\bot\sin(t\Delta\theta)
}
~.
\end{align}

\paragraph{Method 3}
A similar approach, credited to Glenn Davis in \cite{SHOEMAKE-1985}, draws from the fact that any point on the great arc joining $\bfq_0$ to $\bfq_1$ must be a linear combination of its ends (since the three vectors are coplanar). 
Having computed the angle $\Delta\theta$ using \eqRef{equ:slerp_angle},  
%
%
we can isolate $\bfq_\bot$ from \eqRef{equ:q1} and inject it in \eqRef{equ:slerp_rot}. Applying the identity $\sin(\Delta\theta-t\Delta\theta)=\sin \Delta\theta\cos t\Delta\theta-\cos \Delta\theta\sin t\Delta\theta$, we obtain the Davis' formula (see \cite{EBERLY-2010} for an alternative derivation),
\begin{align}
\eqbox{
\bfq(t)=\bfq_0\frac{\sin((1-t)\Delta\theta)}{\sin(\Delta\theta)}+\bfq_1\frac{\sin(t\Delta\theta)}{\sin(\Delta\theta)}
}
~.
\end{align}
This formula has the benefit of being symmetric: defining the reverse interpolator  $s=1-t$ yields
\begin{align*}
\bfq(s)=
\bfq_1\frac{\sin((1-s)\Delta\theta)}{\sin(\Delta\theta)}
+
\bfq_0\frac{\sin(s\Delta\theta)}{\sin(\Delta\theta)}
~.
\end{align*}
which is exactly the same formula with the roles of $\bfq_0$ and $\bfq_1$ swapped.

\begin{figure}[htbp]
\begin{center}
\includegraphics{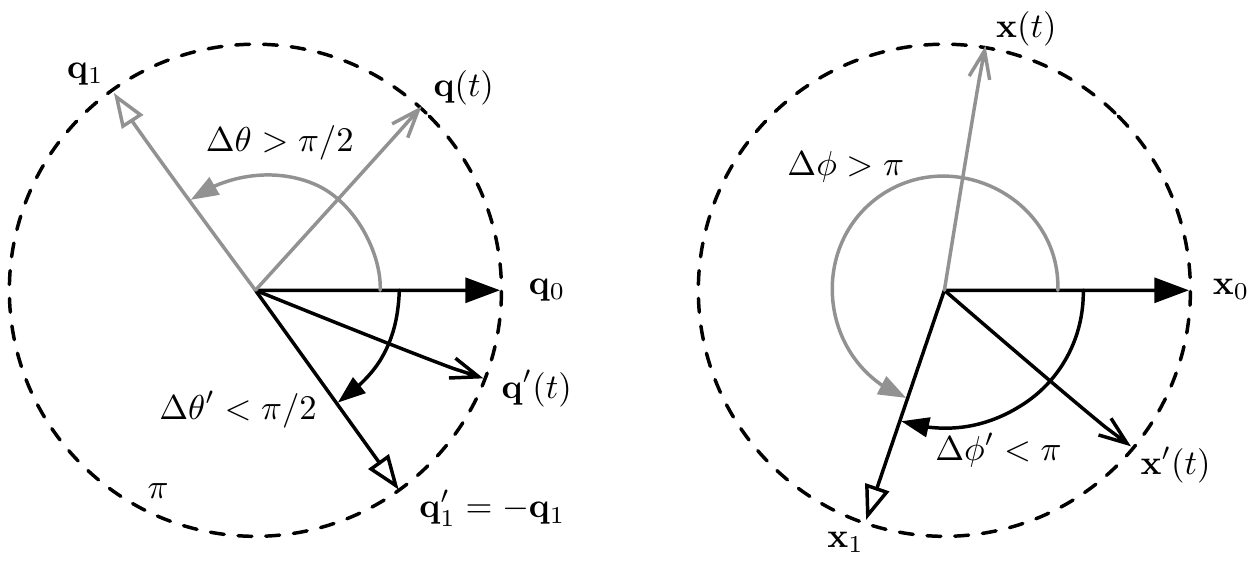}
\caption{Ensuring Slerp along the shortest path between the orientations represented by $\bfq_0$ and $\bfq_1$. Left: quaternion rotation plane in  4D space, showing initial and final orientation quaternions, and two possible interpolations, $\bfq(t)$ from $\bfq_0$ to $\bfq_1$, and $\bfq'(t)$ from $\bfq_0$ to $-\bfq_1$. Right: vector rotation plane in 3D space: 
since $\bfq_1=-\bfq_1$, we have $\bfx_1=\bfq_1\ot\bfx_0\ot\bfq_1^*=\bfq_1'\ot\bfx_0\ot\bfq_1'^*$, that is, both quaternions produce the same rotation.
However, the interpolated quaternion $\bfq(t)$ produces the vector $\bfx(t)$ which takes the long path from $\bfx_0$ to $\bfx_1$, 
while the corrected $\bfq_1'=-\bfq_1$ yields $\bfq'(t)$, producing the vector $\bfx'(t)$ along the shortest path from $\bfx_0$ to $\bfx_1$.}
\label{fig:slerp_fix}
\end{center}
\end{figure}

All these quaternion-based SLERP methods require some care to ensure proper interpolation along the shortest path, that is, with rotation angles $\phi\leq\pi$. 
Due to the quaternion double cover of $SO(3)$ (see \secRef{sec:double_cover}) only the interpolation between quaternions in acute angles $\Delta\theta\leq\pi/2$ is done following the shortest path (\figRef{fig:slerp_fix}). 
Testing for this situation and solving it is simple: if $\cos(\Delta\theta)=\bfq_0\tr\bfq_1<0$, then replace \eg~$\bfq_1$ by $-\bfq_1$ and start over.

\subsection{Quaternion and isoclinic rotations: explaining the magic}
\label{sec:isoclinic}

This section provides geometrical insights to the two intriguing questions about quaternions, what we call the `magic':
\begin{itemize}
\item
How is it that the product $\bfq\ot\bfx\ot\bfq^*$ rotates the vector $\bfx$? 
\item
Why do we need to consider half-angles when constructing the quaternion through $\bfq=e^{\bfphi/2}=[\cos \phi/2 , \bfu\sin\phi/2]$?
\end{itemize}
We want a geometrical explanation, that is, some rationale that goes beyond the algebraic demonstration \eqRef{equ:quatRotFormula} and the double cover facts in \secRef{sec:double_cover}.

To start, let us reproduce here equation \eqRef{equ:quatRotMatrixForm} expressing the quaternion rotation action through the quaternion product matrices $\QL{\bfq}$ and $\QR{\bfq^*}$, defined in \eqRef{equ:quatMatrix},
\begin{align*}
\bfq\otimes
\bfx\otimes\bfq^*
&= \QR{\bfq^*}\,\QL{\bfq}\begin{bmatrix}
0 \\ \bfx
\end{bmatrix} 
= \begin{bmatrix}
0 \\ \bfR\,\bfx
\end{bmatrix} 
~.
\end{align*}
For unit quaternions $\bfq$, the quaternion product matrices $\QL{\bfq}$ and $\QR{\bfq^*}$ satisfy two remarkable properties,
\begin{align}
 \Q{\bfq}\,\Q{\bfq}\tr &= \bfI_4 \\
 \det(\Q{\bfq}) &= +1 
 ~, 
\end{align}%
and are therefore elements of $SO(4)$, that is, proper rotation matrices in the $\bbR^4$ space. 
To be more specific, they represent a particular type of rotation, named \emph{isoclinic rotation}, as we explain hereafter.
Thus, according to \eqRef{equ:quatRotMatrixForm}, a quaternion rotation corresponds to two chained isoclinic rotations in $\bbR^4$.

In order to explain the insights of quaternion rotation, 
we need to understand isoclinic rotations in $\bbR^4$. 
For this, we first need to understand general rotations in $\bbR^4$.
And to understand rotations in $\bbR^4$, we need to go back to $\bbR^3$, whose rotations are in fact planar rotations.
Let us walk all these steps one by one.

\begin{figure}[tb]
\begin{center}
\includegraphics{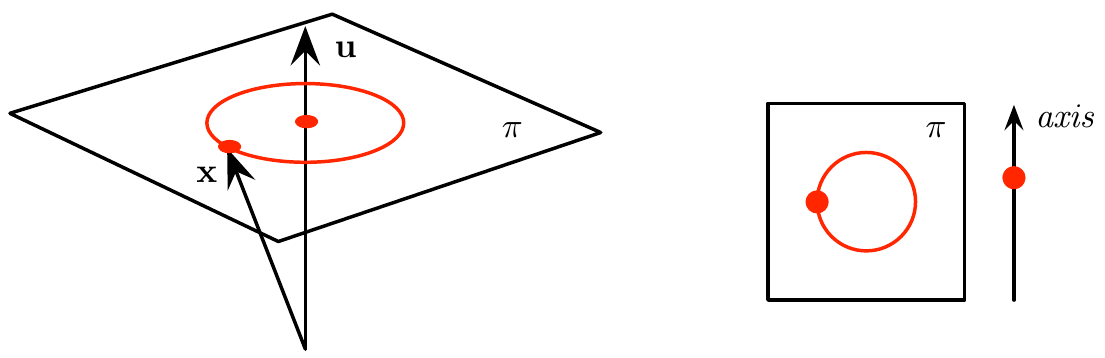}
\caption{%
Rotation in $\bbR^3$. 
A rotation of a vector $\bfx$ around an axis $\bfu$ describes a circumference in a plane orthogonal to the axis. 
The component of $\bfx$ parallel to the axis does not move, and is represented by the small red dot on the axis.
The sketch on the right illustrates the radically different behaviors of the rotating point on the plane and axis subspaces. 
}
\label{fig:isoclinic3}
\end{center}
\end{figure}

\paragraph{Rotations in $\bbR^3$:} 
In $\bbR^3$, let us consider the rotations of a vector $\bfx$ around an arbitrary axis represented by the vector $\bfu$ ---see \figRef{fig:isoclinic3}, and recall  \figRef{fig:rotation3d}. 
Upon rotation, vectors parallel to the axis of rotation $\bfu$ do not move, and  vectors perpendicular to the axis rotate in the plane $\pi$ perpendicular to the axis. 
For general vectors $\bfx$, the two components of the vector in the plane rotate in this plane, while the axial component remains static.

\begin{figure}[tb]
\begin{center}
\includegraphics{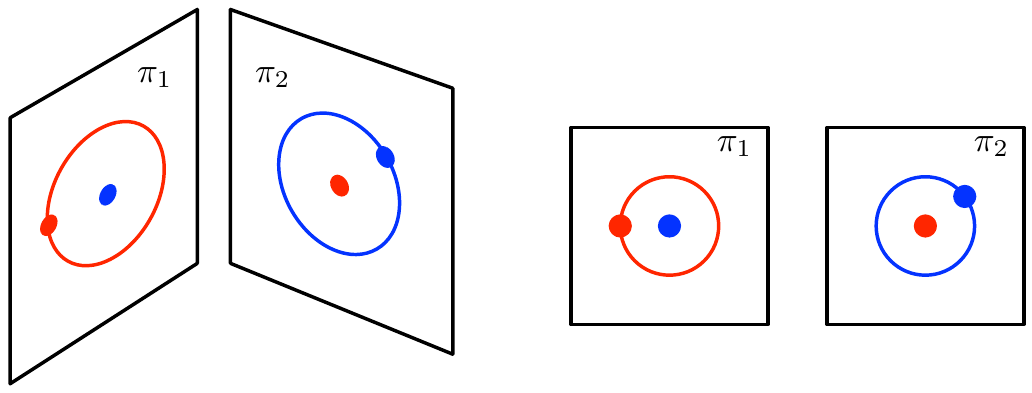}
\caption{%
Rotations in $\bbR^4$. 
Two orthogonal rotations are possible, on two orthogonal planes $\pi_1$ and $\pi_2$. 
Rotations of a vector $\bfx$ (not drawn) in the plane $\pi_1$ cause the two components of the vector parallel to this plane (red dot on $\pi_1$) to describe a circumference (red circle), 
leaving the other two components in $\pi_2$ unchanged (the red dot). 
Conversely, rotations in the plane $\pi_2$ (blue dot on blue circle in $\pi_2$) leave the components in $\pi_1$ unchanged (blue dot).
The sketch on the right better illustrates the situation by resigning to draw unrepresentable perspectives in $\bbR^4$, which might be misleading.
}
\label{fig:isoclinic4}
\end{center}
\end{figure}

\paragraph{Rotations in $\bbR^4$:} 
In $\bbR^4$, see \figRef{fig:isoclinic4}, due to the extra dimension, the one-dimensional axis of rotation in $\bbR^3$ becomes a new two-dimensional plane.
This second plane provides room for a second rotation.
Indeed, rotations in $\bbR^4$ encompass two independent rotations in two orthogonal planes of the 4-space. 
This means that every 4-vector of each of these planes rotates in its own plane, and that rotations of general 4-vectors \wrt one plane leave unaffected the vector components in the other plane. 
These planes are for this reason called \emph{`invariant'}.

\paragraph{Isoclinic rotations in $\bbR^4$:} 
Isoclinic rotations (from Greek, \emph{iso:} ``equal", \emph{klinein:} ``to incline") are those rotations in $\bbR^4$ where the angles of rotation in the two invariant planes have the same magnitude.
Then, when the two angles have also the same sign,%
\footnote{Given the two invariant planes, we arbitrarily select their orientations so that we can associate positive and negative rotation angles in them.}
we speak of \emph{left-isoclinic rotations}. 
And when they have opposite signs, we speak of \emph{right-isoclinic rotations}.
A remarkable property of isoclinic rotations, that we had already seen in  \eqRef{equ:PQ_commute}, is that left- and right- isoclinic rotations commute,
\begin{align}
 \QR{\bfp}\,\QL{\bfq} = \QL{\bfq}\,\QR{\bfp}~. \label{equ:isoclinic_commute}
\end{align}

\begin{figure}[tb]
\begin{center}
\includegraphics{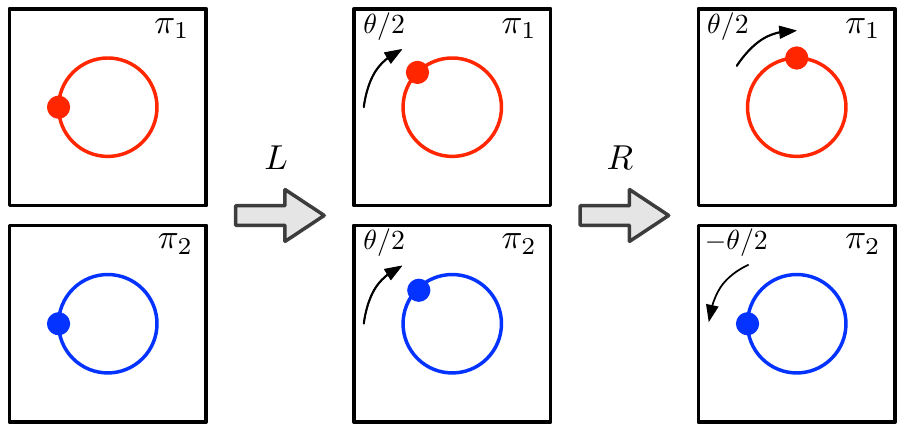}
\caption{%
Quaternion rotation in $\bbR^4$.
Two chained isoclinic rotations, one left (with equal half-angles), and one right (with opposing half-angles), produce a pure rotation by the full angle in only one of the invariant planes.
}
\label{fig:isoclinicQ}
\end{center}
\end{figure}

\paragraph{Quaternion rotations in $\bbR^4$ and $\bbR^3$:}
Given a unit quaternion $\bfq=e^{\bfu\,\theta/2}$, representing a rotation in $\bbR^3$ of an angle $\theta$ around the axis $\bfu$, 
the matrix $\QL{\bfq}$ is a left-isoclinic rotation in $\bbR^4$ corresponding to the left-multiplication by the quaternion $\bfq$, 
and  $\QR{\bfq^*}$ is a right-isoclinic rotation corresponding to the right-multiplication by the quaternion $\bfq^*$. 
The angles of these isoclinic rotations are exactly of magnitude $\theta/2$,%
\footnote{This can be checked by extracting the eigenvalues of the isoclinic rotation matrices: they are formed by pairs of conjugate complex numbers with a phase equal to $\pm\theta/2$.}
and the invariant planes are the same.
Then, the rotation expression \eqRef{equ:quatRotMatrixForm}, reproduced once again here,
\begin{align*}
\begin{bmatrix}
0 \\ \bfx'
\end{bmatrix}=\bfq\otimes\bfx\otimes\bfq^*
&= \QR{\bfq^*}\,\QL{\bfq}\begin{bmatrix}
0 \\ \bfx
\end{bmatrix} 
~,
\end{align*}
represents two chained isoclinic rotations to the 4-vector $(0,\bfx)\tr$, one left- and one right-, each by half the desired rotation angle in $\bbR^3$. 
In one of the invariant planes of $\bbR^4$ (see \figRef{fig:isoclinicQ}), the two half angles cancel out, because they have opposite signs. 
In the other plane, they sum up to yield the total rotation angle $\theta$. 
If we define from  \eqRef{equ:quatRotMatrixForm} the resulting rotation matrix, $\bfR_4$, one easily realizes that (see also \eqRef{equ:isoclinic_commute}),
\begin{align}\label{equ:R4}
\bfR_4 \triangleq \QR{\bfq^*}\QL{\bfq}=\QL{\bfq}\QR{\bfq^*}= \begin{bmatrix}
1 & \bf0 \\
\bf0 & \bfR
\end{bmatrix}
~,
\end{align}
where $\bfR$ is the rotation matrix in $\bbR^3$, which clearly rotates vectors in the $\bbR^3$ subspace of $\bbR^4$, leaving the fourth dimension unchanged.

This discourse is somewhat beyond the scope of the present document. 
It is also incomplete, for it does not provide, beyond the result in \eqRef{equ:R4}, an intuition or geometrical explanation for why we need to do $\bfq\ot\bfx\ot\bfq^*$ instead of \eg~$\bfq\ot\bfx\ot\bfq$.\footnote{Let it suffice to say that $\bfq\ot\bfx\ot\bfq^*$ works for rotations if $\bfq$ is a unit quaternion. In fact, the product $\qv\ot\bfx\ot\qv$ produces reflections (not rotations!) in $\bbR^3$ if $\qv$ is a unit \emph{pure} quaternion. Finally, the product $\bfq\ot\bfx\ot\bfq$, with $\bfq$ a unit \emph{non-pure} quaternion, exhibits no remarkable properties.}
We include it here just as a means for providing yet another way to interpret rotations by quaternions, with the hope that the reader grasps more intuition about its mechanisms.
The interested reader is suggested to consult the appropriate literature on isoclinic rotations in $\bbR^4$.

\section{Quaternion conventions. My choice.}
\label{sec:conventions}

\subsection{Quaternion flavors}

There are several ways to determine the quaternion. They are basically related to four binary choices:
\begin{itemize}
\item
The order of its elements --- real part first or last:
\begin{align}
\bfq = \begin{bmatrix}
q_w \\ \qv
\end{bmatrix} 
\qquad \textit{vs.} \qquad 
\bfq = \begin{bmatrix}
\qv \\ q_w
\end{bmatrix}~.
\label{equ:quatOrder}
\end{align}

\item
The multiplication formula --- definition of the quaternion algebra:
\begin{subequations}
\label{equ:quatAlg}
\begin{align}
ij=-ji=k 
\qquad \textit{vs.} \qquad 
ji=-ij=k~,
\label{equ:quatAlgDef}
\end{align}
which correspond to different handedness, respectively:
\begin{align}
\textit{right-handed
\qquad vs. \qquad
left-handed}~.
\label{equ:quatHand}
\end{align}
\end{subequations}
This means that, given a rotation axis $\bfu$, one quaternion $\bfq_{right}\{\bfu\,\theta\}$ rotates vectors an angle $\theta$ around $\bfu$ using the right hand rule, while the other quaternion $\bfq_{left}\{\bfu\,\theta\}$ uses the left hand rule.

\item
The function of the rotation operator --- rotating frames or rotating vectors:
\begin{align}
\textit{Passive
\qquad vs. \qquad
Active.}
\label{equ:quatAlibi}
\end{align}

\item
In the passive case, the direction of the operation --- local-to-global or global-to-local:
\begin{align}
\bfx_{global} = \bfq\otimes\bfx_{local}\otimes\bfq^*
\qquad vs. \qquad 
\bfx_{local} = \bfq\otimes\bfx_{global}\otimes\bfq^*
\label{equ:quatInterpret}
\end{align}

\end{itemize}

This variety of choices leads to 12 different combinations. Historical developments have favored some conventions over others~\citep{CHOU-92,yazell-09}. 
Today, in the available literature, we find many quaternion flavors such as 
the Hamilton, 
the STS\footnote{Space Transportation System, commonly known as NASA's Space Shuttle.}, 
the JPL\footnote{Jet Propulsion Laboratory.}, 
the ISS\footnote{International Space Station.}, 
the ESA\footnote{European Space Agency.}, 
the Engineering, 
the Robotics, 
and possibly a lot more denominations. 
Many of these forms might be identical, others not, but this fact is rarely explicitly stated, 
and many works simply lack a sufficient description of their quaternion with regard to the four choices above.

These differences impact the respective formulas for rotation, composition, etc., in non-obvious ways. 
The formulas are thus not compatible, and we need to make a clear choice from the very start.

The two most commonly used conventions, which are also the best documented, are Hamilton (the options on the left in \eqsRef{equ:quatOrder}{equ:quatInterpret}) and JPL  (the options on the right, with the exception of \eqRef{equ:quatAlibi}).  
\tabRef{tab:Hamilton_vs_JPL} shows a summary of their characteristics. 
JPL is mostly used in the aerospace domain, while Hamilton is more common to other engineering areas such as robotics ---though this should not be taken as a rule.

\begin{table*}
\renewcommand{\arraystretch}{1.3}
\centering
\caption{Hamilton vs. JPL quaternion conventions \wrt the 4 binary choices}
\vspace{1ex}
\begin{tabular}{|cl|c|c|}
\hline
& Quaternion type & Hamilton & JPL \\
\hline\hline
1 & Components order & $(q_w \,,\, \qv)$ & $(\qv \,,\, q_w)$ \\
\hline
\multirow{2}{*}{2} & Algebra & $ij=k$ & $ij=-k$ \\
& Handedness & Right-handed & Left-handed \\
\hline
3 & Function & Passive & Passive
\\
\hline
\multirow{3}{*}{4} & Right-to-left products mean & Local-to-Global & Global-to-Local \\
& Default notation, $\bfq$ & $\bfq \triangleq \bfq_{\cG\cL}$ & $\bfq \triangleq \bfq_{\cL\cG}$ \\
& Default operation & $\bfx_\cG = \bfq\otimes\bfx_\cL\otimes\bfq^*$ & $\bfx_\cL = \bfq\otimes\bfx_\cG\otimes\bfq^*$ \\
\hline
\end{tabular}
\label{tab:Hamilton_vs_JPL}
\end{table*}

My choice, that has been taken as early as in equation \eqRef{equ:quatAlgebra}, is to take the Hamilton convention, 
which is right-handed and coincides with many software libraries of widespread use in robotics, such as Eigen, ROS, Google Ceres, 
and with a vast amount of literature on Kalman filtering for attitude estimation using IMUs~\citep[and many others]{CHOU-92,KUIPERS-99,PINIES-07,ROUSSILLON-11a,MARTINELLI-12}.

The JPL convention is possibly less commonly used, at least in the robotics field. 
It is extensively described in~\citep{TRAWNY-05-QUAT}, a reference work that has an aim and scope very close to the present one, but that concentrates exclusively in the JPL convention. 
The JPL quaternion is used in the JPL literature (obviously) and in key papers by Li, Mourikis, Roumeliotis, and colleagues (see \eg~\citep{LI-2012,LI-14}), which draw from \citeauthor{TRAWNY-05-QUAT}' document. 
These works are a primary source of inspiration when dealing with visual-inertial odometry and SLAM ---which is what we do. 

In the rest of this section we analyze these two quaternion conventions with a little more depth.

\subsubsection{Order of the quaternion components}

Though not the most fundamental, the most salient difference between Hamilton and JPL quaternions is in the order of the components, with the scalar part being either in first (Hamilton) or last (JPL) position. 
The implications of such change are quite obvious and should not represent a great challenge of interpretation. 
In fact, some works with the quaternion's real component at the end (\eg, the C++ library Eigen) are still considered as using the Hamilton convention, as long as the other three aspects are maintained.

We have used the subscripts $(w, x, y, z)$ for the quaternion components for increased clarity, instead of the other commonly used $(0, 1, 2, 3)$. 
When changing the order, $q_w$ will always denote the real part, while it is not clear whether $q_0$ would also do 
---in some occasions, one might find things such as $\bfq=(q_1, q_2, q_3, q_0)$, with $q_0$ real and last, but in the general case of $\bfq=(q_0, q_1, q_2, q_3)$, the real part at the end would be $q_3$.%
\footnote{See also footnote \ref{ftn:quatComponents}.} 
When passing from one convention to the other, we must be careful of formulas involving full $4\times4$ or $3\times4$ quaternion-related matrices, 
for their rows and/or columns need to be swapped. 
This is not difficult to do, but it might be difficult to detect and therefore prone to error.

Two curiosities about the components' order are:
\begin{itemize}
\item
With real part first, the quaternion is naturally interpreted as an extended complex number, of the familiar form \emph{real+imaginary}. 
Some of us are comfortable with this representation probably because of this.
\item
With real part last, the quaternion expressed in vector form,
$\bfq=\begin{bmatrix}
x,y,z,w
\end{bmatrix}\in\bbH$, 
has a format absolutely equivalent to the homogeneous vector in the projective 3D space, 
$\bfp=\begin{bmatrix}
x,y,z,w
\end{bmatrix}\in\bbP^3$, 
where in both cases $x,y,z$ are clearly identified with the three Cartesian axes. 
When dealing with geometric problems in 3D, this makes the algebra for operating on quaternions and homogeneous vectors more uniform, 
especially (but not only) if the homogeneous vector is constrained to the unit sphere $\norm{\bfp}=1$.
\end{itemize}

\subsubsection{Specification of the quaternion algebra}

The Hamilton convention defines $ij=k$ and therefore,
\begin{align}
i^2 = j^2 = k^2 = ijk = -1~,\quad ij = -ji = k~, \quad jk = -kj = i~, \quad ki = -ik = j~,
\end{align}
whereas the JPL convention defines $ji=k$ and hence its quaternion algebra becomes,
\begin{align}
i^2 = j^2 = k^2 = -ijk = -1~,\quad -ij = ji = k~, \quad -jk = kj = i~, \quad -ki = ik = j~.
\end{align}

Interestingly, these subtle sign changes preserve the basic properties of quaternions as rotation operators. 
Mathematically, the key consequence is the change of the sign of the cross-product in \eqRef{equ:quatProdVec}, which induces a change in the quaternion handedness~\citep{SHUSTER-93}: 
Hamilton uses $ij=k$ and is therefore right-handed, \ie, it turns vectors following the right-hand rule; JPL uses $ji=k$ and is left-handed~\citep{TRAWNY-05-QUAT}. 
Being left- and right- handed rotations of opposite signs, we can say that their quaternions $\bfq_{left}$ and $\bfq_{right}$ are related by,
\begin{align}
\bfq_{\textit{left}}= \bfq_{\textit{right}}^*~.
\end{align}

\subsubsection{Function of the rotation operator}

We have seen how to rotate vectors in 3D. This is referred to in~\citep{SHUSTER-93} as the \emph{active} interpretation, 
because operators (this affects all rotation operators) actively rotate vectors,
\begin{align}
\bfx' &=\bfq_{\textit{active}}\otimes\bfx\otimes\bfq_{\textit{active}}^* ~,
& 
\bfx' &= \bfR_{\textit{active}}\,\bfx~.
\end{align}

Another way of seeing the effect of $\bfq$ and $\bfR$ over a vector $\bfx$ is to consider that the vector is steady but it is us who have rotated our point of view by an amount specified by $\bfq$ or $\bfR$. 
This is called here \emph{frame transformation} and it is referred to in~\citep{SHUSTER-93} as the \emph{passive} interpretation, because vectors do not move,
\begin{align}
\bfx_\cB &= \bfq_{\textit{passive}}\ot\bfx_\cA\ot\bfq_{\textit{passive}}^*~,
&
\bfx_\cB&=\bfR_{\textit{passive}}\,\bfx_\cA~,
\end{align}
where $\cA$ and $\cB$ are two Cartesian reference frames, and $\bfx_\cA$ and $\bfx_\cB$ are expressions of the same vector $\bfx$ in these frames. 
See further down for explanations and proper notations.

The active and passive interpretations are governed by operators inverse of each other, that is, 
\begin{align*}
\bfq_{\textit{active}} &= \bfq_{\textit{passive}}^* ~,
& 
\bfR_{active} &= \bfR_{passive}\tr ~.
\end{align*}
Both Hamilton and JPL use the passive convention. 

\paragraph{Direction cosine matrix}
A few authors understand the passive operator as not being a rotation operator, 
but rather an orientation specification, named the \emph{direction cosine matrix},
\begin{align}
\bfC = \begin{bmatrix}
c_{xx} & c_{xy} & c_{zx} \\
c_{xy} & c_{yy} & c_{zy} \\
c_{xz} & c_{yz} & c_{zz} 
\end{bmatrix}
~,
\end{align}
where each component $c_{ij}$ is the cosine of the angle between the axis $i$ in the source frame and the axis $j$ in the target frame. We have the identity,
\begin{align}
\bfC \equiv \bfR_{\textit{passive}}
~.
\end{align}

\subsubsection{Direction of the rotation operator}

In the passive case, a second source of interpretation is related to the direction in which the rotation matrix and quaternion operate, 
either converting from local to global frames, or from global to local. 

Given two Cartesian frames $\cG$ and $\cL$, we identify $\cG$ and $\cL$ as being the global and local frames.
``Global'' and ``local'' are relative definitions, \ie, $\cG$ is global \wrt $\cL$, and $\cL$ is local \wrt $\cG$ -- in other words, $\cL$ is a frame specified in the reference frame $\cG$.\footnote{Other common denominations for the \{global, local\} frames are \{parent, child\} and \{world, body\}. The first one is convenient when more than two frames are involved in a system (\eg~the frames of each moving link in a humanoid robot); the second one is convenient for a solid vehicle body (\eg~a plane, a car) moving in a unique reference frame identified as the world.} 
We specify $\bfq_{\cG\cL}$ and $\bfR_{\cG\cL}$ as being respectively the quaternion and rotation matrix transforming vectors from frame $\cL$ to frame $\cG$, 
in the sense that a vector $\bfx_\cL$ in frame $\cL$ is expressed in frame $\cG$ with the quaternion- and matrix- products
\begin{align}
\bfx_\cG &= \bfq_{\cG\cL}\otimes\bfx_\cL\otimes\bfq_{\cG\cL}^*~, &
\bfx_\cG &= \bfR_{\cG\cL}\,\bfx_\cL~.
\label{equ:local_to_global}
\end{align}
The opposite conversion, from $\cG$ to $\cL$, is done with
\begin{align}
\bfx_\cL &= \bfq_{\cL\cG}\otimes\bfx_\cG\otimes\bfq_{\cL\cG}^*
~,
&
\bfx_\cL &= \bfR_{\cL\cG}\,\bfx_\cG~,
\end{align}
where
\begin{align}
\bfq_{\cL\cG} &= \bfq_{\cG\cL}^*~,
& 
\bfR_{\cL\cG} &= \bfR_{\cG\cL}\tr~. \label{equ:localVsGlobal}
\end{align}

Hamilton uses local-to-global as the default specification of a frame $\cL$ expressed in frame $\cG$, 
\begin{align}
\bfq_{\textit{Hamilton}} \triangleq \bfq_{[\textit{with~respect~to}][\textit{of\,}]}=\bfq_{[\textit{to}][\textit{from}]}=\bfq_{\cG\cL}~, 
\end{align}
while JPL uses the opposite, global-to-local conversion,
\begin{align}
\bfq_{\textit{JPL}} \triangleq \bfq_{[\textit{of}\,][\textit{with~respect~to}]}=\bfq_{[\textit{to}][\textit{from}]}=\bfq_{\cL\cG}~.
\end{align}

Notice that
\begin{align}
\bfq_{\textit{JPL}}
\triangleq    \bfq_{\cL\cG,left}
=    \bfq_{\cL\cG,\textit{right}}^*
=    \bfq_{\cG\cL,\textit{right}}
\triangleq    \bfq_{\textit{Hamilton}}
~,
\label{equ:quatEquivalences}
\end{align}
which is not particularly useful, but illustrates how easy it is to get confused when mixing conventions.
Notice also that we can conclude that $\bfq_{JPL} = \bfq_{Hamilton}$, but this, far from being a beautiful result, is just the source of great confusion, 
because the equality is only present in the quaternion values, 
but the two quaternions, when employed in formulas,  mean and represent different things.

\section{Perturbations, derivatives and integrals}


\subsection{The additive and subtractive operators in $SO(3)$}

\begin{figure}[tb]
\centering
\includegraphics{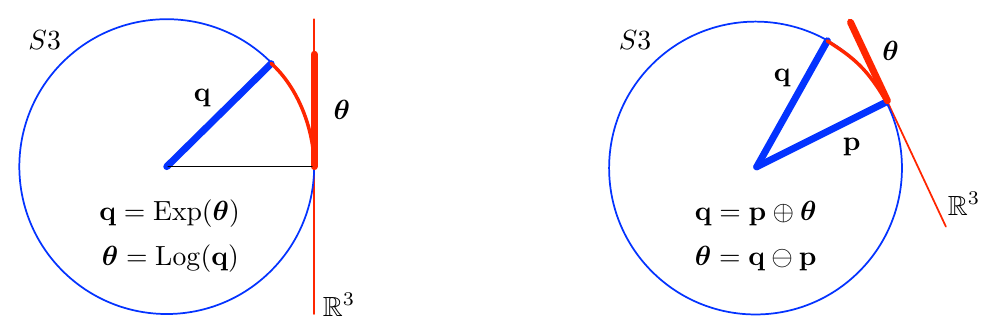}
\caption{The S3 manifold is a unit sphere in $\bbR^4$, here represented by a unit circle (blue),  where all unit quaternions live. 
The tangent space to the manifold is the hyperplane $\bbR^3$, here represented by a line (red). 
\emph{Left}: The $\Exp()$ and $\Log()$ operators map elements of $\bbR^3$ to/from elements of $S3$. 
\emph{Right}: The $\oplus$ and $\ominus$ operators relate elements of the manifold with elements in the tangent space. (Likewise, these figures illustrate the $SO(3)$ manifold.)}
\label{fig:manifold}
\end{figure}

In vector spaces $\bbR^n$, the addition and subtraction operations are performed with the regular sum `$+$' and minus `$-$' operations.
In $SO(3)$ this is not possible, but equivalent operators can be defined for establishing a proper calculus corpus. 

We thus define the plus and minus operators, $\oplus,\ominus$, between elements $\sR\in SO(3)$, and elements $\bth\in\bbR^3$ of the tangent space at $\sR$, as follows.

\paragraph{The plus operator.}
The `plus' operator $\oplus:SO(3)\times\bbR^3\to SO(3)$ produces an element $\sS$ of $SO(3)$ which is the result of composing a reference element $\sR$ of $SO(3)$ with a (often small) rotation. This rotation is specified by a vector of $\bth\in\bbR^3$ in the vector space tangent to the $SO(3)$ manifold at the reference element $\sR$, that is,
\begin{align}
\sS = \sR\oplus \bth &\te \sR\circ\Exp(\bth) && \sR,\sS\in SO(3),~ \bth\in\bbR^3 
~.
\end{align}
Notice that this operator may be defined for any representation of $SO(3)$. In particular, for the quaternion and rotation matrix we have,
\begin{align}
\bfq_\sS &= \,\bfq_\sR\oplus\bth = \bfq_\sR\ot\Exp(\bth) \\
\bfR_\sS &= \bfR_\sR\oplus \bth = \bfR_\sR\tdot\Exp(\bth) 
~.
\end{align}

\paragraph{The minus operator.}
The `minus' operator $\ominus:SO(3)\times SO(3)\to\bbR^3$ is the inverse of the above. It returns the vectorial angular difference $\bth\in\bbR^3$ between two elements of $SO(3)$. This difference is expressed in the  vector space tangent to the reference element $\sR$, 
\begin{align}
\bth=\sS\ominus \sR
&\te \Log(\sR\inv \circ \sS)     && \sR,\sS\in SO(3),~ \bth\in\bbR^3  
~,
\end{align}
which for the quaternion and rotation matrix reads,
\begin{align}
\bth &= \,\,\bfq_\sS\ominus\bfq_\sR\, = \Log(\bfq_\sR^*\ot\bfq_\sS)                      \\
\bth &= \bfR_\sS\ominus\bfR_\sR = \Log(\bfR_\sR\tr\,\bfR_\sS)
~.
\end{align}

\bigskip
In both cases, notice that even though the vector difference $\bftheta$ is typically supposed to be small, the definitions above hold for any value of $\bftheta$ (up to the first coverage of the $SO(3)$ manifold, that is, for angles $\theta<\pi$).

\subsection{The four possible derivative definitions}

\subsubsection{Functions from vector space to vector space}

The scalar and vector cases follow the classical definition of the derivative: given a function $f:\bbR^m\to\bbR^n$, we use $\{+,-\}$ to define the derivative as
\begin{align}
\dpar{f(\bfx)}{\bfx} &\te \lim_{\delta\bfx\to0}\frac{f(\bfx+\delta\bfx)-f(\bfx)}{\delta\bfx} &&\in \bbR^{n\times m} \label{equ:derivative_vector}
\end{align}
Euler integration produces linear expressions of the form
\begin{align*}
f(\bfx+\Delta\bfx) &\approx f(\bfx) + \dpar{f(\bfx)}{\bfx}\Delta\bfx
& \in \bbR^n
\end{align*}

\subsubsection{Functions from $SO(3)$ to $SO(3)$}

Given a function $f:SO(3) \to SO(3)$ with $\sR\in SO(3)$ and a local, small angular variation $\bth\in\bbR^3$, we use $\{\oplus,\ominus\}$ to define the derivative as
\begin{align}
\dpar{f(\sR)}{\bth} 
&\te \lim_{\delta\bth\to0}\frac{f(\sR\oplus\delta\bth)\ominus f(\sR)}{\delta\bth}  && \in \bbR^{3\times 3}\\
&= \lim_{\delta\bth\to0}\frac{\Log\big(f\inv(\sR)\,f(\sR\Exp(\delta\bth))\big)}{\delta\bth} \label{equ:derivative_SO3}
\end{align}
Euler integration produces expressions of the form,
\begin{align*}
f(\sR\oplus\Delta\bth) &\approx f(\sR)\,\oplus\,\dpar{f(\sR)}{\bth}\,\Delta\bth
 \te f(\sR)\Exp\left(\dpar{f(\sR)}{\bth}\Delta\bth\right)
 & \in SO(3)
\end{align*}

\subsubsection{Functions from vector space to $SO(3)$}

For the case of a function $f:\bbR^m\to SO(3)$, we use `+' for the vector perturbations, and `$\ominus$' for the $SO(3)$ difference,
\begin{align}
\dpar{f(\bfx)}{\bfx} &\te \lim_{\delta\bfx\to0} \frac{ f(\bfx+\delta\bfx)\ominus f(\bfx)}{\delta\bfx} && \in \bbR^{3\times m} \label{equ:dif_RtoSO3}\\
&= \lim_{\delta\bfx\to0} \frac{\Log(f\inv(\bfx) f(\bfx+\delta\bfx))}{\delta\bfx}
\end{align}
Euler integration produces expressions of the form,
\begin{align*}
f(\bfx+\Delta\bfx) &\approx f(\bfx)\,\oplus\,\dpar{f(\bfx)}{\bfx}\,\Delta\bfx
 \te f(\bfx)\,\Exp\left(\dpar{f(\bfx)}{\bfx}\Delta\bfx\right)
 & \in SO(3)
\end{align*}

\subsubsection{Functions from $SO(3)$ to vector space}

For the case of a function $f: SO(3)\to\bbR^n$, we use `$\oplus$' for the $SO(3)$ perturbations, and `$-$' for the vector difference,
\begin{align}
\dpar{f(\sR)}{\bth} &\te \lim_{\delta\bth\to0} \frac{f(\sR\oplus\delta\bth) - f(\sR)}{\delta\bth} && \in \bbR^{n\times 3} \label{equ:jacobian_SO3_Rn}\\
&= \lim_{\delta\bth\to0} \frac{f(\sR\Exp(\delta\bth)) - f(\sR)}{\delta\bth}
\end{align}
Euler integration produces expressions of the form,
\begin{align*}
f(\sR\oplus\delta\bth) &\approx f(\sR)+\dpar{f(\sR)}{\bth}\,\Delta\bth
 \te f(\sR)+\Exp\left(\dpar{f(\sR)}{\bth}\Delta\bth\right)
 & \in SO(3)
\end{align*}


\subsection{Useful, and very useful, Jacobians of the rotation}

Let us consider a rotation to a vector $\bfa$, of $\theta$ radians around the unit axis $\bfu$. Let us express the rotation specification in three equivalent forms, namely $\bftheta=\theta\bfu$, $\bfq=\bfq\{\bftheta\}$ and $\bfR=\bfR\{\bftheta\}$. 
We are interested in the Jacobians of the rotated result \wrt different magnitudes.

\subsubsection{Jacobian \wrt the vector}

The derivative of the rotation of a vector $\bfa$ \wrt this vector is trivial,
\begin{align}
\eqbox{
\dpar{(\bfq\ot\bfa\ot\bfq*)}{\bfa} = \dpar{(\bfR\,\bfa)}{\bfa} = \bfR
}
~.
\end{align}
%

\subsubsection{Jacobian \wrt the quaternion}

On the contrary, the derivative of the rotation \wrt the quaternion $\bfq$ is tricky. 
For convenience, we use a lighter notation for the quaternion, $\bfq=[w~\bfv] = w+\bfv$.
We make use of \eqRef{equ:quatProdPure}, \eqRef{equ:quatCommutatorPure}, and the identity $\bfa \times (\bfb \times \bfc) = (\bfc \times \bfb) \times \bfa = (\bfa \tr \bfc)\,\bfb - (\bfa \tr \bfb)\,\bfc$, to develop the quaternion-based rotation \eqRef{equ:sandwichProd} as follows,
\begin{align} \label{equ:drot_dtheta}
\begin{split}
\bfa' &= \bfq\ot\bfa\ot\bfq* \\
&= (w+\bfv)\ot\bfa\ot(w-\bfv) \\
&= w^2\bfa + w(\bfv\ot\bfa - \bfa\ot\bfv) - \bfv\ot\bfa\ot\bfv \\
&= w^2\bfa + 2w(\bfv\tcross\bfa) - \big[(-\bfv\tr\bfa+\bfv\tcross\bfa)\ot\bfv \big]\\
&= w^2\bfa + 2w(\bfv\tcross\bfa) - \big[(-\bfv\tr\bfa)\,\bfv+(\bfv\tcross\bfa)\ot\bfv \big]\\
&= w^2\bfa + 2w(\bfv\tcross\bfa) - \big[(-\bfv\tr\bfa)\,\bfv - \cancel{(\bfv\tcross\bfa)\tr\bfv}+(\bfv\tcross\bfa)\tcross\bfv \big]\\
&= w^2\bfa + 2w(\bfv\tcross\bfa) - \big[(-\bfv\tr\bfa)\,\bfv + (\bfv\tr\bfv)\,\bfa - (\bfv\tr\bfa)\,\bfv \big]\\
&= w^2\bfa + 2w(\bfv\tcross\bfa) + 2(\bfv\tr\bfa)\,\bfv - (\bfv\tr\bfv)\,\bfa
~.
\end{split}
\end{align}%
With this, we can extract the derivatives $\dparil{\bfa'}{w}$ and $\dparil{\bfa'}{\bfv}$,
\begin{align}
\dpar{\bfa'}{w} &= 2(w\bfa + \bfv\tcross\bfa) \\
\begin{split}
\dpar{\bfa'}{\bfv} &= -2w\hatx{\bfa} + 2(\bfv\tr\bfa\,\bfI+\bfv\,\bfa\tr) - 2\bfa\,\bfv\tr 
\\
&= 2(\bfv\tr\bfa\,\bfI+\bfv\,\bfa\tr - \bfa\,\bfv\tr - w\hatx{\bfa} )
~,
\end{split}
\end{align}%
yielding
\begin{align} \label{equ:drot_dq}
\eqbox{
\dpar{(\bfq\ot\bfa\ot\bfq*)}{\bfq} = 
2\begin{bmatrix}
~w\,\bfa + \bfv\tcross\bfa ~~ \big| ~  \bfv\tr\bfa\,\bfI_3+\bfv\,\bfa\tr - \bfa\,\bfv\tr -w\hatx{\bfa}
~\end{bmatrix} \in \bbR^{3\times 4}
}~.
\end{align}

\subsubsection{Right Jacobian of $SO(3)$ }

\begin{figure}[tbp]
\begin{center}
\includegraphics{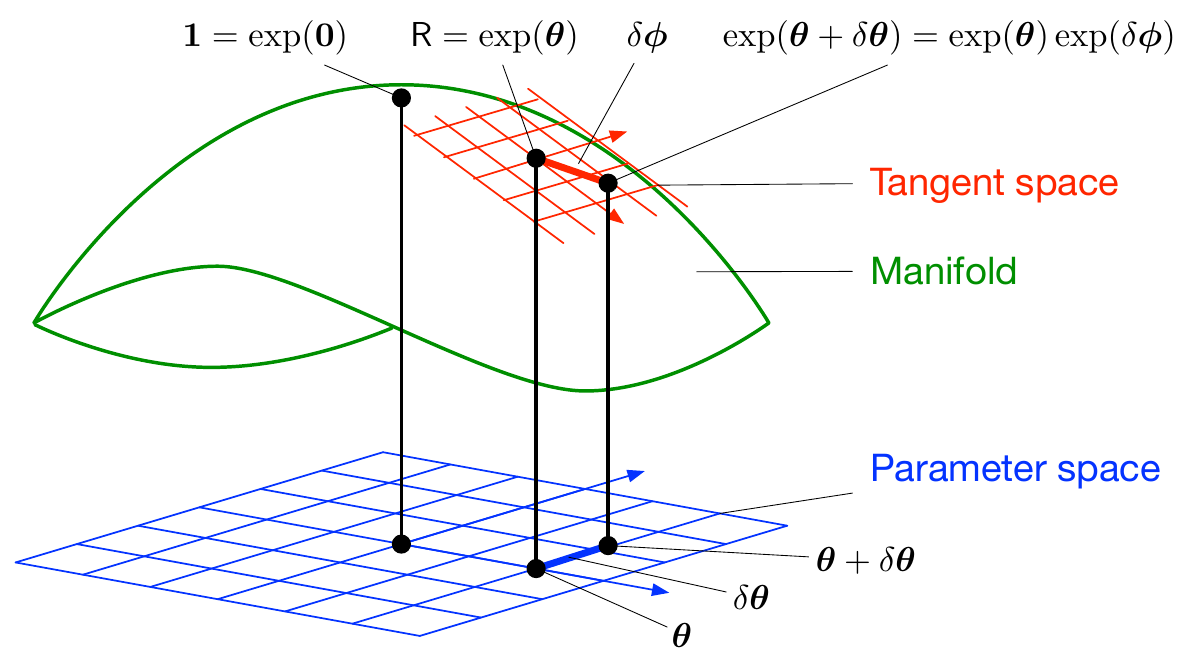}
\caption{The right Jacobian $\bfJ_r=\dparil{\delta\bfphi}{\delta\bftheta}$ maps variations $\delta\bftheta$ around the parameter $\bftheta$ into variations $\delta\bfphi$ on the vector space tangent to the manifold at the point $\Exp{\bftheta}$. }
\label{fig:right_jac}
\end{center}
\end{figure}

Let us consider (see \figRef{fig:right_jac}) an element $\sR\in SO(3)$ and a rotation vector $\bth\in\bbR^3$ such that $\sR=\Exp(\bth)$. When $\bth$ is altered by an amount $\dth$, the element $\sR$ varies. Expressing the variations of $\sR$ in the tangent space of $SO(3)$ at $\sR$ with a rotation vector $\delta\bfphi\in\bbR^3$, we have that (please see the figure, I am not inventing anything here)
\begin{align}
\Exp(\bth)\oplus\delta\bfphi = \Exp(\bth+\dth)
\end{align}
which might be written also as,
\begin{align}
\Exp(\bth)\circ\Exp(\delta\bfphi) &= \Exp(\bth+\dth)
~,
\end{align}
and even
\begin{align}
\delta\bfphi &= \Log\Big(\Exp(\bth)\inv\circ\Exp(\bth+\dth)\Big) = \Exp(\bth+\dth) \ominus \Exp(\bth)
~.
\end{align}

In the limit, the variation of $\delta\bfphi$ as a function of $\dth$ defines a Jacobian matrix
\begin{align}
\dpar{\delta\bfphi}{\dth} 
&= \lim_{\dth\to0}\frac{\delta\bfphi}{\dth} 
= \lim_{\dth\to0}\frac{\Exp(\bth+\dth) \ominus \Exp(\bth)}{\dth} 
~,
\end{align}
whose expression is a particular case of \eqRef{equ:dif_RtoSO3}, that is, it is the derivative of the function $f(\bth)=\Exp(\bth)$, from $\bbR^3$ to $SO(3)$.
This Jacobian matrix is known as the right Jacobian of $SO(3)$, and is defined as,
%
\begin{align}
\bfJ_r(\bth) &\te \dpar{\Exp(\bth)}{\bth} 
~.
\end{align}
Its expression is independent of the parametrization used, though it can indeed be expressed particularly for each parametrization. Using \eqRef{equ:dif_RtoSO3} we have,
\begin{align}
\bfJ_r(\bth) &= \lim_{\dth\to0}\frac{\Exp(\bth+\dth)\ominus\Exp(\bth)}{\dth} \\
 &= \lim_{\dth\to0}\frac{\Log(\Exp(\bth)\tr\Exp(\bth+\dth))}{\dth} && \textrm{if using $\bfR$} \\
 &= \lim_{\dth\to0}\frac{\Log(\Exp(\bth)^*\ot\Exp(\bth+\dth))}{\dth} && \textrm{if using $\bfq$} 
 ~.
\end{align}

The right Jacobian and its inverse can be computed in closed form \citep[page 40]{CHIRIKJIAN-12},
\begin{align}
\bfJ_r(\bth) &= \bfI - \frac{1-\cos\nth}{\nth^2}\hatx{\bth} + \frac{\nth-\sin\nth}{\nth^3}\hatx{\bth}^2 \\
\bfJ_r\inv(\bth) &= \bfI + \frac12\hatx{\bth} + \left(\frac1{\nth^2} - \frac{1+\cos\nth}{2\nth\sin\nth}\right)\hatx{\bth}^2
\end{align}

The right Jacobian of $SO(3)$ has the following properties, for any $\bth$ and small $\dth$,
\begin{align}
\Exp(\bth+\dth) &\approx \Exp(\bth)\Exp(\bfJ_r(\bth)\dth) \label{equ:Jr1} \\
\Exp(\bth)\Exp(\dth) &\approx \Exp(\bth+\bfJ_r\inv(\bth)\,\dth) \\
\Log(\Exp(\bth)\Exp(\dth)) &\approx \bth+\bfJ_r\inv(\bth)\,\dth 
\end{align}

\subsubsection{Jacobian \wrt the rotation vector}

The rotation of a vector $\bfa'=\bfR\{\bftheta\}\,\bfa$ \wrt the rotation vector $\bth$ is a function from $\bbR^3$ to $\bbR^3$. Its derivative \wrt the rotation vector $\bftheta$ uses \eqRef{equ:derivative_vector} and is developed from the previous result, using \eqRef{equ:Jr1},
\begin{align*} 
\dpar{(\bfq\ot\bfa\ot\bfq^*)}{\delta\bftheta} 
= \dpar{(\bfR\,\bfa)}{\delta\bftheta} 
&= \lim_{\delta\bftheta\to 0} \frac{\bfR\{\bftheta+\delta\bftheta\}\,\bfa-\bfR\{\bftheta\}\,\bfa}{\delta\bftheta} &&\gets\eqRef{equ:derivative_vector}\\
&= \lim_{\delta\bftheta\to 0} \frac{(\bfR\{\bftheta\}\Exp(\bfJ_r(\bftheta)\,\delta\bftheta)-\bfR\{\bftheta\})\bfa}{\delta\bftheta} && \gets\eqRef{equ:Jr1} \\
&= \lim_{\delta\bftheta\to 0} \frac{(\bfR\{\bftheta\}(\bfI+\hatx{\bfJ_r(\bftheta)\,\delta\bftheta})-\bfR\{\bftheta\})\bfa}{\delta\bftheta} \\
&= \lim_{\delta\bftheta\to 0} \frac{\bfR\{\bftheta\}\hatx{\bfJ_r(\bftheta)\,\delta\bftheta}\bfa}{\delta\bftheta} \\
&= \lim_{\delta\bftheta\to 0} -\frac{\bfR\{\bftheta\}\hatx{\bfa}\bfJ_r(\bftheta)\,\delta\bftheta}{\delta\bftheta} \\
&= -\bfR\{\bftheta\}\hatx{\bfa}\bfJ_r (\bftheta)
~,
\end{align*}
where $\bfR\{\bftheta\}\te\Exp(\bftheta)$. Summarizing,
\begin{align} \label{equ:drot_da}
\eqbox{\dpar{(\bfq\ot\bfa\ot\bfq^*)}{\delta\bftheta} 
= \dpar{(\bfR\,\bfa)}{\delta\bftheta} 
= -\bfR\{\bftheta\}\hatx{\bfa}\bfJ_r(\bftheta) 
}~.
\end{align}

\subsubsection{Jacobians of the rotation composition}

Consider the SO(3) composition $\sP=\sQ\circ\sR$, which can be implemented in either quaternion or matrix form,
\begin{align}
\bfp &= \bfq_\theta\ot\bfr_\phi & \bfP &= \bfQ_\theta\,\bfR_\phi
\end{align}
where the subindices indicate the name of the vector perturbations in the tangent space. 
These are functions from $SO(3)$ to $SO(3)$, and therefore we use \eqRef{equ:derivative_SO3} to write the drivatives,
\begin{align*}
\dpar{\sQ\circ\sR}{\sQ} 
= \dpar{\bfq_\theta\ot\bfr_\phi}{\bth} 
= \dpar{\bfQ_\theta \bfR_\phi}{\bth} 
&= \lim_{\delta\bftheta\to 0}\frac{((\bfQ_\theta\oplus\dth)\bfR_\phi)\ominus(\bfQ_\theta\bfR_\phi)}{\dth} \\
&= \lim_{\delta\bftheta\to 0}\frac{\Log[(\bfQ_\theta\bfR_\phi)\tr(\bfQ_\theta\Exp(\dth)\bfR_\phi)]}{\dth} \\
&= \lim_{\delta\bftheta\to 0}\frac{\Log[\bfR_\phi\tr\Exp(\dth)\bfR_\phi]}{\dth} \\
&= \lim_{\delta\bftheta\to 0}\frac{\Log[\Exp(\bfR_\phi\tr\dth)]}{\dth} \\
&= \lim_{\delta\bftheta\to 0}\frac{\bfR_\phi\tr\dth}{\dth}  = \bfR_\phi\tr 
\end{align*}
\begin{align*}
\dpar{\sQ\circ\sR}{\sR} 
= \dpar{\bfq_\theta\ot\bfr_\phi}{\bfphi} 
= \dpar{\bfQ_\theta \bfR_\phi}{\bfphi} 
&= \lim_{\delta\bfphi\to 0}\frac{(\bfQ_\theta(\bfR_\phi\oplus\delta\bfphi))\ominus(\bfQ_\theta\bfR_\phi)}{\delta\bfphi} \\
&= \lim_{\delta\bfphi\to 0}\frac{\Log[(\bfQ_\theta\bfR_\phi)\tr(\bfQ_\theta\bfR_\phi\Exp(\delta\bfphi))]}{\delta\bfphi} \\
&= \lim_{\delta\bfphi\to 0}\frac{\Log[\Exp(\delta\bfphi)]}{\delta\bfphi} \\
&= \lim_{\delta\bfphi\to 0}\frac{\delta\bfphi}{\delta\bfphi}  = \bfI 
\end{align*}

\subsection{Perturbations, uncertainties, noise}

\subsubsection{Local perturbations}

A perturbed orientation $\tilde{\bfq}$ may be expressed as the composition of the unperturbed orientation $\bfq$ with a small local perturbation ${\Delta\bfq_\cL}$. 
Because of the Hamilton convention, this local perturbation appears \emph{at the right hand side} of the composition product ---we give also the matrix equivalent for comparison,
\begin{align}
\tilde{\bfq} &= \bfq\ot{\Delta\bfq_\cL}
~, &
\tilde\bfR &= \bfR\,\Delta\bfR_\cL
~.
\end{align}%
These local perturbation $\Delta\bfq_\cL$ (or $\Delta\bfR_\cL$) is easily obtained from its equivalent vector form $\Delta\bfphi_\cL=\bfu\Delta\phi_\cL$, defined in the tangent space, using the exponential map. This gives
\begin{align}
\tilde\bfq_\cL &= \bfq_\cL\ot\Exp(\Delta\bfphi_\cL)
~,& 
\tilde\bfR_\cL &= \bfR_\cL\tdot\Exp(\Delta\bfphi_\cL)
\end{align}
leading to an expression of the local perturbation 
\begin{align}
\Delta\bfphi_\cL = \Log(\bfq_\cL^*\ot\tilde\bfq_\cL) = \Log(\bfR_\cL\tr\tdot\tilde\bfR_\cL)
\end{align}

If the perturbation angle $\Delta\phi_\cL$ is small then the perturbation in quaternion and rotation matrix forms can be approximated by the Taylor expansions of \eqRef{equ:vectoquat} and \eqRef{equ:vectomat} up to the linear terms,
\begin{align}
{\Delta\bfq_\cL}& \approx \begin{bmatrix}
1\\\frac{1}{2}\Delta\bfphi_\cL
\end{bmatrix}
~,
&
\Delta\bfR_\cL& \approx
\bfI+\hatx{\Delta\bfphi_\cL}
~.
\end{align}%
Perturbations can therefore be specified in the local vector space $\Delta\bfphi_\cL$ tangent to the $SO(3)$ manifold at the actual orientation. It is convenient, for example, to express the covariances matrix of these perturbations in this vectorial space, that is, with a regular $3\times 3$ covariance matrix.

\subsubsection{Global perturbations}

It is possible and indeed interesting to consider globally-defined perturbations, and likewise for the related derivatives. 
Global perturbations appear \emph{at the left hand side} of the composition product, namely,
%
%
\begin{align}
\tilde\bfq_\cG &= \Exp(\Delta\bfphi_\cG)\ot\bfq_\cG
~,& 
\tilde\bfR_\cG &= \Exp(\Delta\bfphi_\cG)\tdot\bfR_\cG
\end{align}
leading to an expression of the global perturbation 
\begin{align}
\Delta\bfphi_\cG = \Log(\tilde\bfq_\cG\ot\bfq_\cG^*) = \Log(\tilde\bfR_\cG\tdot\bfR_\cG\tr)
\end{align}

Again, these perturbations can be specified in the vector space $\Delta\bfphi_\cG$ tangent to the $SO(3)$ manifold at the origin.

\subsection{Time derivatives}

Expressing the local perturbations in a vector space we can easily develop expressions for the time-derivatives. 
Just consider $\bfq=\bfq(t)$ as the original state, $\tilde{\bfq}=\bfq(t+\Dt)$ as the perturbed state, and apply the definition of the derivative
\begin{align}
\dif{\bfq(t)}{t} \triangleq \lim_{\Dt\to0} \frac{\bfq(t+\Dt)-\bfq(t)}{\Dt}~, \label{equ:derivative}
\end{align}%
to the above, with
\begin{align}
\bfomega_\cL(t) \triangleq \dif{\bfphi_\cL(t)}{t} \triangleq \lim_{\Dt\to0} \frac{\Delta\bfphi_\cL}{\Dt}~,
\end{align}%
which, being $\Delta\bfphi_\cL$ a local angular perturbation, corresponds to the angular rates vector in the local frame defined by $\bfq$.

The development of the time-derivative of the quaternion follows (an analogous reasoning would be used for the rotation matrix)
\begin{align}
\dot{\bfq} &\triangleq \lim_{\Dt\to0} \frac{\bfq(t+\Dt)-\bfq(t)}{\Dt} \nonumber\\
&= \lim_{\Dt\to0} \frac{\bfq\ot{\Delta\bfq_\cL}-\bfq}{\Dt} \nonumber\\
&= \lim_{\Dt\to0} \frac{\bfq\ot\left(\begin{bmatrix}
1\\\Delta\bfphi_\cL/2
\end{bmatrix}-\begin{bmatrix}
1\\\bf0
\end{bmatrix}\right)}{\Dt} \nonumber\\ 
&= \lim_{\Dt\to0} \frac{\bfq\ot\begin{bmatrix}
0\\\Delta\bfphi_\cL/2
\end{bmatrix}}{\Dt} \nonumber\\ 
&= \frac12\,\bfq\ot\begin{bmatrix}
0\\\bfomega_\cL
\end{bmatrix}
~. \label{equ:lastquatdev}
\end{align}%
Defining
\begin{align}
\bfOmega(\bfomega) 
\triangleq \QR{\bfomega} 
= \begin{bmatrix}
0 & -\bfomega\tr \\
\bfomega & -\hatx{\bfomega}
\end{bmatrix} = \begin{bmatrix}
0        & -\omega_x & -\omega_y & -\omega_z \\
\omega_x & 0         &  \omega_z & -\omega_y \\
\omega_y & -\omega_z & 0         & \omega_x \\
\omega_z &  \omega_y & -\omega_x & 0
\end{bmatrix} ~, \label{equ:Omega}
\end{align}%
we get from \eqRef{equ:lastquatdev} and \eqRef{equ:quatMatProd} (we give also its matrix equivalent)
\begin{empheq}[box=\widefbox]{align}
\label{equ:qdotLocal}
\dot{\bfq} &= \frac{1}{2}\bfOmega(\bfomega_\cL)\,\bfq = \frac{1}{2}\bfq\ot\bfomega_\cL
~,
&
\dot\bfR &= \bfR\hatx{\bfomega_\cL}
~.
\end{empheq}

These expressions are of course identical to \eqRef{equ:qdot} and \eqRef{equ:Rdot}, developed in the framework of the rotation group $SO(3)$. 
Here, however, and interestingly, we are able to clearly refer the angular rate $\bfomega_\cL$ to a particular reference frame, which in this case is the local frame defined by the orientation $\bfq$ or $\bfR$.
This has been possible now because we have given the operators $\bfq$ and $\bfR$ a precise geometrical meaning.
From this viewpoint, \eqRef{equ:qdotLocal} expresses the evolution of the orientation of a reference frame, when the angular rates are expressed locally in this frame.


The time-derivatives associated to global perturbations follow from a development analogous to \eqRef{equ:lastquatdev}, which results in
\begin{empheq}[box=\widefbox]{align}
\label{equ:qdotGlobal}
\dot\bfq &= \frac12\,\bfomega_\cG\ot\bfq~,
&
\dot\bfR &= \hatx{\bfomega_\cG}\bfR~,
\end{empheq}
where
\begin{align}
\bfomega_\cG(t)\triangleq\dif{\bfphi_\cG(t)}{t}
\end{align}
is the angular rates vector expressed in the global frame.
Eq.~\eqRef{equ:qdotGlobal} expresses the evolution of the orientation of a reference frame, when the angular rates are expressed in the global reference frame.
\subsubsection{Global-to-local relations}

From the previous paragraph, it is worth noticing the following relation between local and global angular rates,
\begin{align}
\frac12\,\bfomega_\cG\ot\bfq = \dot\bfq = \frac12\,\bfq\ot\bfomega_\cL ~.
\end{align}
Then, post-multiplying by the conjugate quaternion we have
\begin{align}
\bfomega_\cG = \bfq\ot\bfomega_\cL\ot\bfq^* = \bfR\,\bfomega_\cL~.
\end{align}
Likewise, considering that $\Delta\bfphi_R \approx \bfomega\Dt$ for small $\Dt$, we have that
\begin{align}
\Delta\bfphi_\cG = \bfq\ot\Delta\bfphi_\cL\ot\bfq^* = \bfR\,\Delta\bfphi_\cL~.
\end{align}
That is, we can transform angular rates vectors $\bfomega$ and small angular perturbations $\Delta\bfphi$ via frame transformation, using the quaternion or the rotation matrix, as if they were regular vectors. The same can be seen by posing $\bfomega=\bfu\omega$, or $\Delta\bfphi=\bfu\Delta\phi$, and noticing that the rotation axis vector $\bfu$ transforms normally, with
\begin{align}
\bfu_\cG=\bfq\ot\bfu_\cL\ot\bfq^*=\bfR\,\bfu_\cL ~.
\end{align}

\subsubsection{Time-derivative of the quaternion product}

We use the regular formula for the derivative of the product,
\begin{align}
\dot{({\bfq_1\ot\bfq_2})} &= \dot{\bfq_1}\ot{\bfq_2} + \bfq_1\ot\dot{{\bfq_2}}~, 
&
\dot{(\bfR_1\bfR_2)} &= \dot\bfR_1\bfR_2+\bfR_1\dot\bfR_2~,
\end{align}%
but noticing that, since the products are non commutative, we need to respect the order of the operands strictly.
This means that $\dot{(\bfq^2)}\neq2\,\bfq\ot\dot\bfq
$~, as it would be in the scalar case, but rather
\begin{align}
\dot{(\bfq^2)} = \dot\bfq\ot\bfq + \bfq\ot\dot\bfq 
~.
\end{align}

\subsubsection{Other useful expressions with the derivative}

We can derive an expression for the local rotation rate
\begin{align}
\bfomega_\cL &= 2\,\bfq^*\ot\dot{\bfq}~, 
&
\hatx{\bfomega_\cL} &= \bfR\tr\,\dot\bfR~.
\end{align}%
and the global rotation rate,
\begin{align}
\bfomega_\cG &= 2\,\dot{\bfq}\ot\bfq^*~, 
&
\hatx{\bfomega_\cG} &= \dot\bfR\,\bfR\tr~.
\end{align}%

\subsection{Time-integration of rotation rates}

Accumulating rotation over time in quaternion form is done by integrating the differential equation appropriate to the rotation rate definition, that is, \eqRef{equ:qdotLocal} for a local rotation rate definition, and \eqRef{equ:qdotGlobal} for a global one. 
In the cases we are interested in, the angular rates are measured by local sensors, thus providing local measurements $\bfomega(t_n)$ at discrete times $t_n=n\Dt$. 
We concentrate here on this case only, for which we reproduce the differential equation \eqRef{equ:qdotLocal},
\begin{align}
\dot\bfq(t) 
= \frac12\bfq(t)\ot\bfomega(t) 
~.
\label{equ:intLocal}
\end{align}

\begin{figure}[tb]
\centering
\includegraphics{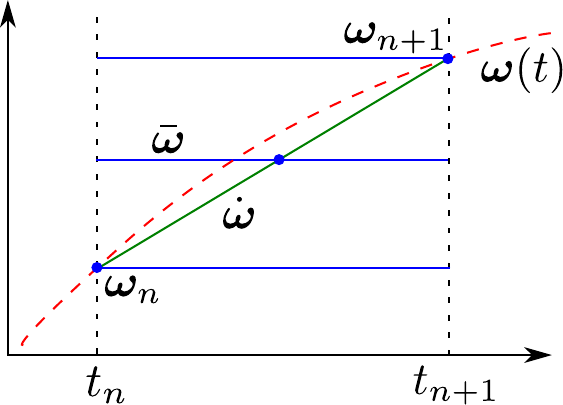}
\caption{Angular velocity approximations for the integral: Red: true velocity. Blue: zero-th order approximations (bottom to top: forward, midward and backward). Green: first order approximation.}
\label{fig:quatInt}
\end{figure}

\begin{figure}[tb]
\begin{center}
\includegraphics{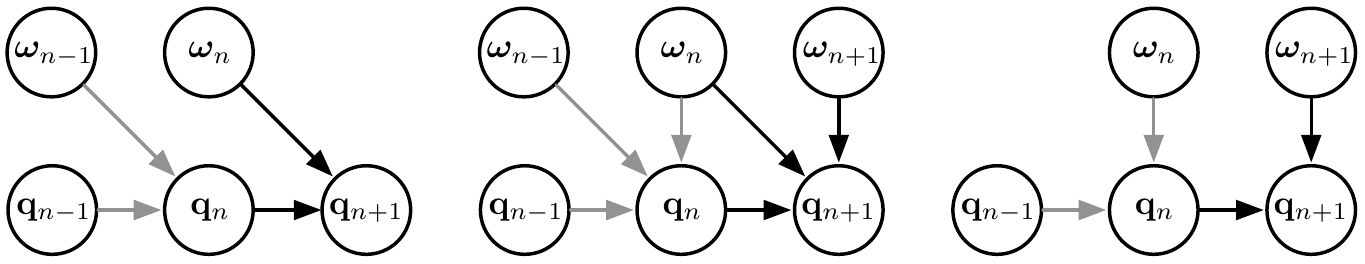}
\caption{Integration schemes for two consecutive time steps (gray and black arrow sets), where variables sharing the same time stamp have been organized in columns. Left: forward integration. Center: midward and first-order integrations. Right: backward integration.}
\label{fig:integrate}
\end{center}
\end{figure}

We develop zeroth- and first- order integration methods~(Figs.~\ref{fig:quatInt} and \ref{fig:integrate}), all based on the Taylor series of $\bfq(t_n+\Dt)$ around the time $t=t_n$. 
We note $\bfq\triangleq\bfq(t)$ and $\bfq_n\triangleq\bfq(t_n)$, and the same for $\bfomega$. 
The Taylor series reads, 
\begin{align}
\q_{n+1} = \q_n + \dq_n\Dt + \frac1{2!}\ddq_n\Dt^2 + \frac1{3!}\dddq_n\Dt^3 + \frac1{4!}\ddddq_n\Dt^4 + \cdots~.
\label{equ:qnTaylor}
\end{align}
The successive derivatives of $\bfq_n$ above are easily obtained by repeatedly applying the expression of the quaternion derivative, \eqRef{equ:intLocal}, with $\ddot\bfomega=0$. We obtain
\begin{subequations}
\begin{align}
\dq_n 
\e \frac12\q_n\w_n \\
\ddq_n 
\e \frac1{2^2}\q_n\w_n^2+\frac12\q_n\dw \\
\dddq_n 
\e \frac1{2^3}\q_n\w_n^3 + \frac14\q_n\dw\w_n + \frac12\q\w_n\dw \\
\q_n^{(i\,\ge\,4)} 
\e \frac1{2^i}\q_n\w_n^i + \cdots~,
\end{align}%
\label{equ:qnDerivatives}%
\end{subequations}%
where we have omitted the $\ot$ signs for economy of notation, that is, all products and the powers of $\w$ must be interpreted in terms of the quaternion product.

\subsubsection{Zeroth order integration}

\paragraph{Forward integration}
In the case where the angular rate $\bfomega_n$ is held constant over the period $[t_n,t_{n+1}]$, 
we have $\dot\bfomega=0$ and \eqRef{equ:qnTaylor} reduces to,
\begin{align}
\q_{n+1} = \q_n\ot\left(
1 
+ \frac12\bfomega_n\Dt 
+ \frac1{2!}\Big(\frac12\bfomega_n\Dt\Big)^2 
+ \frac1{3!}\Big(\frac12\bfomega_n\Dt\Big)^3 
+  \cdots \right)~,
\end{align}
where we identify the Taylor series \eqRef{equ:pureQuatExpSeries} of the exponential $e^{\bfomega_n\Dt/2}$. 
From 
\eqRef{equ:vectoquat}, 
this exponential corresponds to the quaternion representing the incremental rotation $\Delta\theta=\bfomega_n\Dt$, 
\begin{align*}
e^{\bfomega\Dt/2} = \Exp(\bfomega\Dt) = \bfq\{\bfomega\Dt\} = \begin{bmatrix}
\cos(\norm{\bfomega}\Dt/2) \\
\frac{\bfomega}{\norm{\bfomega}}\sin(\norm{\bfomega}\Dt/2)
\end{bmatrix}~,
\end{align*}
therefore,
\begin{align}
\eqbox{
\bfq_{n+1} = \bfq_n\ot\bfq\{\bfomega_n\Dt\} 
} 
~.
\label{equ:intZeroth}
\end{align}

\paragraph{Backward integration}
We can also consider that the constant velocity over the period $\Dt$ corresponds to $\bfomega_{n+1}$, the velocity measured at the end of the period. This can be developed in a similar manner with a Taylor expansion of $\bfq_n$ around $t_{n+1}$, leading to
\begin{align}
{\bfq_{n+1} \approx \bfq_n\ot\bfq\{\bfomega_{n+1}\Dt\}} ~.
\end{align}

We want to remark here that this is the typical integration method when the arriving motion measurements are to be processed in real time, because the integration horizon corresponds to the last measurement (in this case, $t_{n+1}$, see \figRef{fig:integrate}). To make this more salient, we can re-label the time indices to use $\{n-1,n\}$ instead of $\{n,n+1\}$, and write,
\begin{align}
\eqbox{\bfq_n = \bfq_{n-1}\ot\bfq\{\bfomega_n\Dt\}} ~.
\end{align}

\paragraph{Midward integration}

Similarly, if the velocity is considered constant at the median rate over the period $\Dt$ (which is not necessary the velocity at the midpoint of the period),
\begin{align}
\aw = \frac{\w_{n+1}+\w_n}{2}~,
\label{equ:wbar}
\end{align}
we have,
\begin{align}
\eqbox{
\q_{n+1} = \bfq_n\ot\bfq\{\aw\Dt\} 
}
~.
\label{equ:intFirstC}
\end{align}

\subsubsection{First order integration}

The angular rate $\w(t)$ is now linear with time. Its first derivative is constant, and all higher ones are zero,
\begin{align}
\dw \e \frac{\w_{n+1}-\w_n}{\Dt} \label{equ:wdot} \\
\ddot\w = \dddot\w = \cdots \e 0 \label{equ:wddot}
~.
\end{align}%
We can write the median rate $\aw$ in terms of $\w_n$ and $\dw$,
\begin{align}
\aw = \w_n+\frac12\dw\Dt~,
\end{align}
and derive the expression of the powers of $\w_n$ appearing in the quaternion derivatives \eqRef{equ:qnDerivatives}, in terms of the more convenient $\aw$ and $\dw$, 
\begin{subequations}
\begin{align}
\w_n \e \aw-\frac12\dw\Dt \label{equ:wone}\\
\w_n^2 \e \aw^2 - \frac12\aw\dw\Dt - \frac12\dw\aw\Dt + \frac14\dw^2\Dt^2 \\
\w_n^3 \e \aw^3 - \frac32\aw^2\dw\Dt + \frac34\aw\dw^2\Dt^2 + \frac18\dw^3\Dt^3 \\
\w_n^4 \e \aw^4 + \cdots \label{equ:wthree}
~.
\end{align}%
\end{subequations}
Injecting them in the quaternion derivatives, and substituting in the Taylor series \eqRef{equ:qnTaylor}, we have after proper reordering,
\begin{subequations}
\begin{align}
\q_{n+1} 
\e
\q\left(1+\frac12\aw\Dt+\frac1{2!}\left(\frac12\aw\Dt\right)^2+\frac1{3!}\left(\frac12\aw\Dt\right)^3+\cdots\right) \label{equ:exp} \\
&+\: \q\left(-\frac14\dw + \frac14\dw \right)\Dt^2 \label{equ:vanish}\\
&+\: \q\left(-\frac1{16}\aw\dw - \frac1{16}\dw\aw + \frac1{24}\dw\aw + \frac1{12}\aw\dw\right)\Dt^3 \label{equ:bracket} \\
&+\: \q\,\bigg(\:\cdots\:\bigg)\Dt^4 \:+\: \cdots 
\label{equ:neglected}
~,
\end{align}%
\end{subequations}
where in \eqRef{equ:exp} we recognize the exponential series $e^{\aw\Dt/2}= \q\{\aw\Dt\}$, \eqRef{equ:vanish} vanishes, and \eqRef{equ:neglected} represents terms of high multiplicity that we are going to neglect. 
This yields after simplification  (we recover now the normal $\ot$ notation),
\begin{align}
\q_{n+1} = \q_n\ot\q\{\aw\Dt\} + \frac{\Dt^3}{48}\q_n\ot(\aw\ot\dw - \dw\ot\aw) + \cdots ~.
\end{align}
Substituting $\dw$ and $\aw$ by their definitions \eqRef{equ:wdot} and \eqRef{equ:wbar} we get,
\begin{align}
\q_{n+1} = \q_n\ot\q\{\aw\Dt\} + \frac{\Dt^2}{48}\q_n\ot(\w_n\ot\w_{n+1} - \w_{n+1}\ot\w_n) + \cdots ~,
\label{equ:intFirstA}
\end{align}
which is a result equivalent to \citep{TRAWNY-05-QUAT}'s, but using the Hamilton convention, and the quaternion product form instead of the matrix product form. 
Finally, since $\bfa_v\ot\bfb_v-\bfb_v\ot\bfa_v=2\,\bfa_v\tcross\bfb_v$, see \eqRef{equ:quatCommutatorPure}, we have the alternative form,
%
%
\begin{align} \label{equ:intFirstB}
\eqbox{\q_{n+1} \approx \bfq_n\ot\left(\bfq\{\aw\Dt\} + \frac{\Dt^2}{24}\,\begin{bmatrix}
0\\\w_n\tcross\w_{n+1}
\end{bmatrix}\right)
}~.
\end{align}
In this expression, the first term of the sum is the midward zeroth order integrator \eqRef{equ:intFirstC}. 
The second term is a second-order correction that vanishes when $\w_n$ and $\w_{n+1}$ are collinear,%
\footnote{Notice also from \eqRef{equ:intFirstA} that this term would \emph{always} vanish if the quaternion product were commutative, which is not.}
\ie, when the axis of rotation has not changed from $t_n$ to $t_{n+1}$.

\paragraph{Case of fixed rotation axis}

Let us write $\bfomega(t)=\bfu(t)\,\omega(t)$ and call $\bfu$ the axis of rotation.
In the case of a constant rotation axis $\bfu(t) = \bfu$, we have $\w_n\tcross\w_{n+1}=0$ and therefore,
\begin{align}\label{equ:first_order_constant_axis}
\q_{n+1} = \bfq_n\ot\bfq\{\bfu\,\ol\omega\,\Dt\}~.
\end{align}
This result is in fact interesting for cases not limited to first-order derivatives of $\bfomega(t)$. 
In effect, if the axis of rotation is constant, the infinitesimal contributions of rotation into the quaternion commute, \ie, 
$$\exp (\bfu\,\omega_1\,\delta t_1)\exp (\bfu\,\omega_2\,\delta t_2)=\exp (\bfu\,\omega_2\,\delta t_2)\exp (\bfu\,\omega_1\,\delta t_1)=\exp (\bfu\,(\omega_1\delta t_1 + \omega_2\delta t_2))~,$$ 
and thus we have the identity,
\begin{subequations}
\begin{align}
\q_{n+1} 
  &= \bfq_n\ot \exp\left(\frac{\bfu}2\,\int_{t_n}^{t_{n+1}}\omega(t)\,\dt\right) \\
  &= \bfq_n\ot \exp(\bfu\,\Delta\theta_n/2) \\
  &= \bfq_n\ot \bfq\{\bfu\,\Delta\theta_n\}~.
\end{align}
\end{subequations}
with $\Delta\theta_n = \int_{t_n}^{t_{n+1}} \omega(t)dt \in \bbR$ the total angle rotated  during the interval $[t_n,t_{n+1}]$.

\paragraph{Case of varying rotation axis}

Clearly, the second term of the sum in \eqRef{equ:intFirstB} captures through $\w_n\tcross\w_{n+1}\neq0$ the effect that a varying rotation axis has on the integrated orientation.
For its practical usage, we notice that given usual IMU sampling times $\Dt\le0.01s$, and the usual near-collinearity of $\w_n$ and $\w_{n+1}$ due to inertia, this second-order term takes values of the order of $10^{-6}\nw^2$, or easily smaller. 
Terms with higher multiplicities of $\w\Dt$ are even smaller and have been neglected.

Please note also that, while all zeroth-order integrators result in unit quaternions by construction (because they are computed as the product of two unit quaternions), this is not the case for the first-order integrator due to the sum in \eqRef{equ:intFirstB}. Hence, when using the first-order integrator, and even if the summed term is small as stated, users should take care to check the evolution of the quaternion norm over time, and eventually re-normalize the quaternion if needed, using quaternion updates of the form $\bfq\gets\bfq/\norm{\bfq}$. Only if the constant axis assumption holds, then \eqRef{equ:first_order_constant_axis} holds too and this normalization is no longer necessary.


\section{Error-state kinematics for IMU-driven systems}
\label{sec:es-kinematics}

\subsection{Motivation}

We wish to write the error-estate equations of the kinematics of an inertial system integrating accelerometer and gyrometer readings with bias and noise, using the Hamilton quaternion to represent the orientation in space or \emph{attitude}. 

Accelerometer and gyrometer readings come typically from an Inertial Measurement Unit (IMU). 
Integrating IMU readings leads to dead-reckoning positioning systems, which drift with time. 
Avoiding drift is a matter of fusing this information with absolute position readings such as GPS or vision.

The error-state Kalman filter (ESKF) is one of the tools we may use for this purpose. 
Within the Kalman filtering paradigm, these are the most remarkable assets of the ESKF~\citep{MADYASTHA-11}:

\begin{itemize}
\item The orientation error-state is minimal (\ie, it has the same number of parameters as degrees of freedom), avoiding issues related to over-parametrization (or redundancy) and the consequent risk of singularity of the involved covariances matrices, resulting typically from enforcing constraints.
\item The error-state system is always operating close to the origin, and therefore far from possible parameter singularities, gimbal lock issues, or the like, providing a guarantee that the linearization validity holds at all times.
\item The error-state is always small, meaning that all second-order products are negligible. 
This makes the computation of Jacobians very easy and fast. 
Some Jacobians may  even be constant or equal to available state magnitudes.
\item The error dynamics are slow because all the large-signal dynamics have been integrated in the nominal-state. This means that we can apply KF corrections (which are the only means to observe the errors) at a lower rate than the predictions.
\end{itemize}

\subsection{The error-state Kalman filter explained}

In error-state filter formulations, we speak of true-, nominal- and error-state values, the true-state being expressed as a suitable composition (linear sum, quaternion product or matrix product) of the nominal- and the error- states. 
The idea is to consider the nominal-state as large-signal (integrable in non-linear fashion) and the error-state as small signal (thus linearly integrable and suitable for linear-Gaussian filtering). 

The error-state filter can be explained as follows. 
On one side, high-frequency IMU data $\bfu_m$ is integrated into a nominal-state $\bfx$. 
This nominal state does not take into account the noise terms $\bfw$ and other possible model imperfections. 
As a consequence, it will accumulate errors. 
These errors are collected in the error-state $\delta\bfx$ and estimated with the  Error-State Kalman Filter (ESKF), this time incorporating all the noise and perturbations. 
The error-state consists of small-signal magnitudes, and its evolution function is correctly defined by a (time-variant) linear dynamic system, with its dynamic, control and measurement matrices computed from the values of the nominal-state. 
In parallel with integration of the nominal-state, the ESKF predicts a Gaussian estimate of the error-state. 
It only predicts, because by now no other measurement is available to correct these estimates. 
The filter correction is performed at the arrival of information other than IMU (\eg~GPS, vision, etc.), which is able to render the errors observable and which  happens generally at a much lower rate than the integration phase. 
This correction provides a posterior Gaussian estimate of the error-state. 
After this, the error-state's mean is injected into the nominal-state, then reset to zero. 
The error-state's covariances matrix is conveniently updated to reflect this reset. 
The system goes on like this forever.

\subsection{System kinematics in continuous time}

The definition of all the involved variables is summarized in \tabRef{tab:errorstatevar}. 
Two important decisions regarding conventions are worth mentioning:
\begin{itemize}
\item
The angular rates $\bfomega$ are defined \emph{locally} with respect to the nominal quaternion. 
This allows us to use the gyrometer measurements $\bfomega_m$ directly, as they provide body-referenced angular rates.
\item
The angular error $\delta\bftheta$ is also defined \emph{locally} with respect to the nominal orientation. 
This is not necessarily the optimal way to proceed, but it corresponds to the choice in most IMU-integration works ---what we could call the \emph{classical approach}. 
There exists evidence \citep{LI-2012} that a globally-defined angular error has better properties. 
This will be explored too in the present document, \secRef{sec:ESKFglobal}, but most of the developments, examples and algorithms here are based in this locally-defined angular error.
\end{itemize}

\begin{table*}[tb]
\renewcommand{\arraystretch}{1.3}
\caption{All variables in the error-state Kalman filter. }
\centering
\vspace{1ex}
\begin{tabular}{|l|c|c|c|c|c|c|}
\hline
Magnitude & True & Nominal & Error & Composition & Measured & Noise \\
\hline
\hline
Full state ($ ^1$)& $\bfx_t$ & $\bfx$ & $\delta\bfx$ & $\bfx_t = \bfx\oplus\delta\bfx$ & & \\
\hline
\hline
Position & $\bfp_t$ & $\bfp$ & $\delta\bfp$ & $\bfp_t = \bfp+\delta\bfp$ & & \\
Velocity & $\bfv_t$ & $\bfv$ & $\delta\bfv$ &$\bfv_t = \bfv+\delta\bfv$& & \\
Quaternion ($ ^{2,3}$)& $\bfq_t$ & $\bfq$ & ${\delta\bfq}$ &$\bfq_t = \bfq\ot{\delta\bfq}$& & \\
Rotation matrix ($ ^{2,3}$)& $\bfR_t$ & $\bfR$ & $\delta\bfR$ &$\bfR_t = \bfR\,\delta\bfR$& & \\
 \multirow{2}{*}{Angles vector ($ ^{4}$)} &  &  &  \multirow{2}{*}{$\delta\bftheta$} & 
$\delta\bfq = e^{\delta\bftheta/2} 
	$ & & \\
& & & & 
$\delta\bfR = e^{\hatx{\delta\bftheta}} 
	$ & &\\
\hline 
Accelerometer bias & $\bfa_{bt}$ & $\bfa_b$ & $\delta\bfa_b$ &$\bfa_{bt} = \bfa_b+\delta\bfa_b$& & $\bfa_w$ \\
Gyrometer bias & $\bfomega_{bt}$ & $\bfomega_b$ & $\delta\bfomega_b$ &$\bfomega_{bt} = \bfomega_b+\delta\bfomega_b$& & $\bfomega_w$ \\
Gravity vector & $\bfg_t$ & $\bfg$ & $\delta\bfg$ & $\bfg_t = \bfg+\delta\bfg$ & & \\
\hline\hline
Acceleration & $\bfa_t$ & 
&&& $\bfa_m$ & $\bfa_n$ \\
Angular rate & $\bfomega_t$ & 
&&& $\bfomega_m$ & $\bfomega_n$ \\
\hline
\multicolumn{7}{l}{($ ^1$) the symbol $\oplus$ indicates a generic composition} \\
\multicolumn{7}{l}{($ ^2$) indicates non-minimal representations} \\
\multicolumn{7}{l}{($ ^3$) see \tabRef{tab:local_to_global} for the composition formula in case of globally-defined angular errors}\\
\multicolumn{7}{l}{($ ^4$) exponentials defined as in (\ref{equ:vectoquat}) and (\ref{equ:vectomat},\,\ref{equ:rodrigues})}
\end{tabular}
\label{tab:errorstatevar}
\end{table*}%

\subsubsection{The true-state kinematics}

The true kinematic equations are
\begin{subequations}
\begin{align}
\dot\bfp_t &= \bfv_t \\
\dot\bfv_t &= \bfa_t \\
\dot{\bfq_t} &= \frac{1}{2}\bfq_t\ot\bfomega_t \\
\dot\bfa_{bt} &= \bfa_w \\
\dot\bfomega_{bt} &= \bfomega_w \\
\dot\bfg_t &= 0
\end{align}%
\end{subequations}
Here, the true acceleration $\bfa_t$ and angular rate $\bfomega_t$ are obtained from an IMU in the form of noisy sensor readings  $\bfa_m$ and $\bfomega_m$ in body frame, namely\footnote{It is common practice to neglect the Earth's rotation rate $\bfomega_\cE$ in the rotational kinematics described in \eqRef{equ:gyroModel}, which would otherwise be $\bfomega_m = \bfomega_t + \bfR_t\tr\bfomega_\cE + \bfomega_{bt} + \bfomega_n$. 
Considering a non-null Earth rotation rate is, in the vast majority of practical cases, unjustifiably complicated.
However, we notice that when employing high-end IMU sensors with very small noises and biases, a value of $\omega_\cE=15^\circ$/h $\approx 7.3\cdot10^{-5}\,$rad/s might become directly measurable; in such cases, in order to keep the IMU error model valid, the rate $\bfomega_\cE$ should not be neglected in the formulation.}
\begin{align}
\bfa_m &= \bfR_t\tr(\bfa_t - \bfg_t) + \bfa_{bt} + \bfa_n \\
\bfomega_m &= \bfomega_t + \bfomega_{bt} + \bfomega_n \label{equ:gyroModel}
\end{align}%
with $\bfR_t\triangleq\bfR\{\bfq_t\}$. With this, the true values can be isolated (this means that we have inverted the measurement equations),
\begin{align}
\bfa_t &= \bfR_t(\bfa_m - \bfa_{bt} - \bfa_n) + \bfg_t \\
\bfomega_t &= \bfomega_m - \bfomega_{bt} - \bfomega_n.
\end{align}%
Substituting above yields the kinematic system
\begin{subequations}
\begin{align}
\dot\bfp_t &= \bfv_t \label{equ:pos} \\
\dot\bfv_t &= \bfR_t(\bfa_m - \bfa_{bt} - \bfa_n) + \bfg_t \label{equ:vel} \\
\dot{\bfq_t} &= \frac{1}{2}\bfq_t\ot(\bfomega_m - \bfomega_{bt} - \bfomega_n) \label{equ:quat}\\
\dot\bfa_{bt} &= \bfa_w \label{equ:abias}\\
\dot\bfomega_{bt} &= \bfomega_w \label{equ:wbias}\\
\dot\bfg_t &= 0 \label{equ:grav}
\end{align}%
\end{subequations}
which we may name $\dot\bfx_t=f_t(\bfx_t,\bfu,\bfw)$. 
This system has state $\bfx_t$, is governed by IMU noisy readings $\bfu_m$, and is perturbed by white Gaussian noise $\bfw$, all defined by
\begin{equation}
\bfx_t = \begin{bmatrix}
\bfp_t \\ \bfv_t \\ \bfq_t \\ \bfa_{bt} \\ \bfomega_{bt} \\ \bfg_t
\end{bmatrix} 
\Quad
\bfu = \begin{bmatrix}
\bfa_m - \bfa_n \\ \bfomega_m - \bfomega_n
\end{bmatrix}
\Quad
\bfw = \begin{bmatrix}
\bfa_w \\ \bfomega_w
\end{bmatrix}~.
\end{equation}

\bigskip
It is to note in the above formulation that the gravity vector $\bfg_t$ is going to be estimated by the filter. 
It has a constant evolution equation, \eqRef{equ:grav}, as corresponds to a magnitude that is known to be constant. 
The system starts at a fixed and arbitrarily known initial orientation $\bfq_t(t=0)=\bfq_0$, which, being generally not in the horizontal plane, makes the initial gravity vector generally unknown. 
For simplicity it is usually taken $\bfq_0=(1, 0, 0, 0)$ and thus $\bfR_0=\bfR\{\bfq_0\}=\bfI$. 
We estimate $\bfg_t$ expressed in frame $\bfq_0$, and not $\bfq_t$ expressed in a horizontal frame, so that the initial uncertainty in orientation is transferred to an initial uncertainty on the gravity direction. 
We do so to improve linearity: indeed, equation \eqRef{equ:vel} is now linear in $\bfg$, which carries all the uncertainty, and the initial orientation $\bfq_0$ is known without uncertainty, so that $\bfq$ starts with no uncertainty. 
Once the gravity vector is estimated the horizontal plane can be recovered and, if desired, the whole state and recovered motion trajectories can be re-oriented to reflect the estimated horizontal. 
See \citep{LUPTON-09} for further justification. 
This is of course optional, and the reader is free to remove all equations related to graviy from the system and adopt a more classical approach of considering $\bfg\triangleq(0,0,-9.8xx)$, with $xx$ the appropriate decimal digits of the gravity vector on the site of the experiment, and an uncertain initial orientation $\bfq_0$.

\subsubsection{The nominal-state kinematics}

The nominal-state kinematics corresponds to the modeled system without noises or perturbations,
\begin{subequations}
\begin{align}
\dot\bfp &= \bfv \label{equ:pdot}\\
\dot\bfv &= \bfR(\bfa_m - \bfa_b) + \bfg \label{equ:vdot}\\
\dot{\bfq} &= \frac{1}{2}\bfq\ot(\bfomega_m - \bfomega_b) \\
\dot\bfa_b &= 0 \\
\dot\bfomega_b &= 0 \\
\dot\bfg &= 0 .
\end{align}%
\end{subequations}
%

\subsubsection{The error-state kinematics}

The goal is to determine the linearized dynamics of the error-state. 
For each state equation, we write its composition (in \tabRef{tab:errorstatevar}), solving for the error state and simplifying all second-order infinitesimals. 
We give here the full error-state dynamic system and proceed afterwards with comments and proofs.
\begin{subequations}\label{equ:efull}
\begin{align}
\dot{\delta\bfp} &= \delta\bfv \label{equ:epos}\\
\dot{\delta\bfv} &= -\bfR\hatx{\bfa_m-\bfa_b}\delta\bftheta - \bfR\delta\bfa_b + \delta\bfg - \bfR\bfa_n \label{equ:evel}\\
\dot{\delta\bftheta} &= -\hatx{\bfomega_m-\bfomega_b}\delta\bftheta - \delta\bfomega_b - \bfomega_n \label{equ:equat}\\
\dot{\delta\bfa_b} &= \bfa_w \label{equ:eabias}\\
\dot{\delta\bfomega_b} &= \bfomega_w \label{equ:ewbias}\\
\dot{\delta\bfg} &= 0 .\label{equ:egrav}
\end{align}%
\end{subequations}
Equations \eqRef{equ:epos}, \eqRef{equ:eabias}, \eqRef{equ:ewbias} and \eqRef{equ:egrav}, respectively of position, both biases, and gravity errors, are derived from linear equations and their error-state dynamics is trivial. 
As an example, consider the true and nominal position equations \eqRef{equ:pos} and \eqRef{equ:pdot}, their composition $\bfp_t=\bfp+\delta\bfp$ from \tabRef{tab:errorstatevar}, and solve for $\dot{\delta \bfp}$ to obtain \eqRef{equ:epos}.

Equations \eqRef{equ:evel} and \eqRef{equ:equat}, of velocity and orientation errors, require some non-trivial manipulations of the non-linear equations \eqRef{equ:vel} and \eqRef{equ:quat} to obtain the linearized dynamics
. 
Their proofs are developed in the following two sections.

\paragraph{Equation \eqRef{equ:evel}: The linear velocity error.}

We wish to determine $\dot{\delta\bfv}$, the dynamics of the velocity errors. 
We start with the following relations
\begin{align}
\bfR_t &= \bfR(\bfI+\hatx{\delta\bftheta})  + O(\norm{\delta\bftheta}^2) \label{equ:Rt}\\
\dot\bfv &= \bfR\bfa_\cB + \bfg, \label{equ:vdot2}
\end{align}%
where \eqRef{equ:Rt} is the small-signal approximation of $\bfR_t$, and in \eqRef{equ:vdot2} we rewrote \eqRef{equ:vdot} but introducing $\bfa_\cB$ and $\delta\bfa_\cB$, defined as the large- and small-signal accelerations in body frame,
\begin{align}
\bfa_\cB &\triangleq \bfa_m - \bfa_b \label{equ:nomacc}\\
\delta\bfa_\cB &\triangleq -\delta\bfa_b - \bfa_n  \label{equ:pertacc}
\end{align}%
so that we can write the true acceleration in inertial frame as a composition of large- and small-signal terms,
\begin{equation}
\bfa_t = \bfR_t(\bfa_\cB+\delta\bfa_\cB) + \bfg + \delta\bfg.
\end{equation}%

We proceed by writing the expression \eqRef{equ:vel} of $\dot\bfv_t$ in two different forms (left and right developments), where the terms $O(\norm{\delta\bftheta}^2)$ have been ignored,
\begin{align*}
\dot\bfv+\dot{\delta\bfv} =& \eqbox{\dot\bfv_t} = \bfR(\bfI+\hatx{\delta\bftheta})(\bfa_\cB+\delta\bfa_\cB)+\bfg + \delta\bfg \\
\bfR\bfa_\cB+\bfg+\dot{\delta\bfv} =&~~~~~~= \bfR\bfa_\cB+\bfR\delta\bfa_\cB+\bfR\hatx{\delta\bftheta}\bfa_\cB+\bfR\hatx{\delta\bftheta}\delta\bfa_\cB+\bfg+\delta\bfg 
\end{align*}%
This leads after removing $\bfR\bfa_\cB+\bfg$ from left and right to
\begin{equation}
\dot{\delta\bfv} = \bfR(\delta\bfa_\cB+\hatx{\delta\bftheta}\bfa_\cB) + \bfR\hatx{\delta\bftheta}\delta\bfa_\cB + \delta\bfg
\end{equation}%
Eliminating the second order terms and reorganizing some cross-products (with $\hatx{\bfa}\bfb=-\hatx{\bfb}\bfa$), we get
\begin{equation}
\dot{\delta\bfv} = \bfR(\delta\bfa_\cB - \hatx{\bfa_\cB}\delta\bftheta) + \delta\bfg,
\end{equation}%
then, recalling \eqRef{equ:nomacc} and \eqRef{equ:pertacc},
\begin{equation}
{\dot{\delta\bfv} = \bfR(-\hatx{\bfa_m-\bfa_b}\delta\bftheta - \delta\bfa_b - \bfa_n) + \delta\bfg}
\end{equation}%
which after proper rearranging leads to the dynamics of the linear velocity error,
\begin{equation}
\eqbox{\dot{\delta\bfv} = -\bfR\hatx{\bfa_m-\bfa_b}\delta\bftheta - \bfR\delta\bfa_b + \delta\bfg - \bfR\bfa_n}\ .
\end{equation}%
To further clean up this expression, we can often times assume that the accelerometer noise is white, uncorrelated and isotropic\footnote{This assumption cannot be made in cases where the three $XYZ$ accelerometers are not identical.}, 
\begin{equation}
\bbE[\bfa_n] = 0 \Quad \bbE[\bfa_n\bfa_n\tr]=\sigma_a^2\bfI,
\end{equation}%
that is, the covariance ellipsoid is a sphere centered at the origin, which means that its mean and covariances matrix are invariant upon rotations (\emph{Proof:} $\bbE[\bfR\bfa_n] = \bfR\bbE[\bfa_n] = 0$ and $\bfE[(\bfR\bfa_n)(\bfR\bfa_n)\tr] = \bfR\bbE[\bfa_n\bfa_n\tr]\bfR\tr = \bfR\sigma_a^2\bfI\bfR\tr = \sigma_a^2\bfI$). 
Then we can redefine the accelerometer noise vector, with absolutely no consequences, according to
\begin{equation}
\bfa_n \gets \bfR\bfa_n
\end{equation}%
which gives
\begin{equation}
\eqbox{\dot{\delta\bfv} = -\bfR\hatx{\bfa_m-\bfa_b}\delta\bftheta - \bfR\delta\bfa_b + \delta\bfg - \bfa_n}\ .
\end{equation}%

\paragraph{Equation \eqRef{equ:equat}: The orientation error.}

We wish to determine $\dot{\delta\bftheta}$, the dynamics of the angular errors. We start with the following relations
\begin{align}
\dot{\bfq_t} &= \frac{1}{2}\bfq_t\ot\bfomega_t \\
\dot{\bfq} &= \frac{1}{2}\bfq\ot\bfomega ,
\end{align}%
which are the true- and nominal- definitions of the quaternion derivatives.

As we did with the acceleration, we group large- and small-signal terms in the angular rate for clarity,
\begin{align}
\bfomega &\triangleq \bfomega_m - \bfomega_b \label{equ:nomangrate}\\
\delta\bfomega &\triangleq -\delta\bfomega_b - \bfomega_n, \label{equ:pertangrate}
\end{align}%
so that $\bfomega_t$ can be written with a nominal part and an error part,
\begin{equation}
\bfomega_t = \bfomega + \delta\bfomega .
\end{equation}%

We proceed by computing $\dot{\bfq_t}$ by two different means (left and right developments)
\begin{align*}
\dot{{(\bfq\ot{\delta\bfq})}} =& \eqbox{\dot{\bfq_t}} = \frac{1}{2}\bfq_t\ot\bfomega_t \\
\dot{\bfq}\ot{\delta\bfq} + \bfq\ot\dot{{\delta\bfq}} =&~~~~~~= \frac{1}{2}\bfq\ot{\delta\bfq}\ot\bfomega_t \\
\frac{1}{2}\bfq\ot\bfomega\ot{\delta\bfq}+\bfq\ot\dot{{\delta\bfq}} =&& 
\end{align*}%
simplifying the leading $\bfq$ and isolating $\dot{{\delta\bfq}}$ we obtain
\begin{align}
\begin{bmatrix}
0\\\dot{\delta\bftheta}
\end{bmatrix} = \eqbox{2\dot{{\delta\bfq}}} &= {\delta\bfq}\ot\bfomega_t - \bfomega\ot{\delta\bfq} \nonumber \\
&= [\bfq]_R(\bfomega_t){\delta\bfq} - [\bfq]_L(\bfomega){\delta\bfq} \nonumber \\
&= \begin{bmatrix}
0 & -(\bfomega_t-\bfomega)\tr \\
(\bfomega_t-\bfomega) & -\hatx{\bfomega_t+\bfomega} 
\end{bmatrix}\begin{bmatrix}
1 \\
\delta\bftheta/2
\end{bmatrix} + O(\norm{\delta\bftheta}^2) \nonumber \\
&= \begin{bmatrix}
0 & -\delta\bfomega\tr \\
\delta\bfomega & -\hatx{2\bfomega+\delta\bfomega} 
\end{bmatrix}\begin{bmatrix}
1 \\
\delta\bftheta/2
\end{bmatrix} + O(\norm{\delta\bftheta}^2) 
\end{align}%
which results in one scalar- and one vector- equalities
\begin{subequations}
\begin{align}
0 &= \delta\bfomega\tr\delta\bftheta + O(|\delta\bftheta|^2) \\
\dot{\delta\bftheta} &= \delta\bfomega - \hatx\bfomega\delta\bftheta - \frac{1}{2}\hatx{\delta\bfomega}\delta\bftheta + O(\norm{\delta\bftheta}^2).
\end{align}%
\end{subequations}
The first equation leads to $\delta\bfomega\tr\delta\bftheta = O(\norm{\delta\bftheta}^2)$, which is formed by second-order infinitesimals, not very useful. 
The second equation yields, after neglecting all second-order terms,
\begin{equation}
{\dot{\delta\bftheta} = -\hatx\bfomega\delta\bftheta} + \delta\bfomega 
\end{equation}%
and finally, recalling \eqRef{equ:nomangrate} and \eqRef{equ:pertangrate}, we get the linearized dynamics of the angular error,
\begin{equation}
\eqbox{\dot{\delta\bftheta} = -\hatx{\bfomega_m-\bfomega_b}\delta\bftheta - \delta\bfomega_b - \bfomega_n}\ .
\end{equation}%

\subsection{System kinematics in discrete time}

The differential equations above need to be integrated into differences equations to account for discrete time intervals $\Dt>0$. 
The integration methods may vary. 
In some cases, one will be able to use exact closed-form solutions. 
In other cases, numerical integration of varying degree of accuracy may be employed. 
Please refer to the Appendices for pertinent details on integration methods.

Integration needs to be done for the following sub-systems:
\begin{enumerate}
\item The nominal state.
\item The error-state.
\begin{enumerate}
\item The deterministic part: state dynamics and control.
\item The stochastic part: noise and perturbations.
\end{enumerate}
\end{enumerate}

\subsubsection{The nominal state kinematics}

We can write the differences equations of the nominal-state as
\begin{subequations}
\begin{align}
\bfp &\gets \bfp + \bfv\, \Dt + \frac{1}{2}(\bfR(\bfa_{m}-\bfa_{b})+\bfg)\, \Dt ^2\\
\bfv &\gets \bfv + (\bfR(\bfa_{m}-\bfa_{b})+\bfg)\, \Dt \\
\bfq &\gets \bfq\ot \bfq\{(\bfomega_{m} - \bfomega_{b})\, \Dt\} \\
\bfa_{b} &\gets \bfa_{b} \\
\bfomega_{b} &\gets \bfomega_{b} \\
\bfg &\gets \bfg~,
\end{align}%
\end{subequations}%
where $x\gets f(x,\bullet)$ stands for a time update of the type $x_{k+1} = f(x_k,\bullet_k)$, $\bfR\triangleq\bfR\{\bfq\}$ is the rotation matrix associated to the current nominal orientation $\bfq$, and $\bfq\{v\}$ is the quaternion associated to the rotation $v$, according to \eqRef{equ:vectoquat}.

We can also use more precise integration, please see the Appendices for more information.

\subsubsection{The error-state kinematics}

The deterministic part is integrated normally (in this case we follow the methods in \appRef{sec:BlockWiseTruncation}), and the integration of the stochastic part results in random impulses (see \appRef{sec:IntNoise}), thus,
\begin{subequations}
\begin{align}
\delta\bfp &\gets \delta\bfp + \delta\bfv\,\Dt \\
\delta\bfv &\gets \delta\bfv + (-\bfR\hatx{\bfa_{m}-\bfa_{b}}\delta\bftheta - \bfR\delta\bfa_{b} + \delta\bfg)\Dt + \bfv_\bfi \\
\delta\bftheta &\gets \bfR\tr\{(\bfomega_{m}-\bfomega_{b})\Dt\}\delta\bftheta - \delta\bfomega_{b} \Dt + \bftheta_\bfi \\
\delta\bfa_{b} &\gets \delta\bfa_{b} + \bfa_\bfi \\
\delta\bfomega_{b} &\gets \delta\bfomega_{b} + \bfomega_\bfi \\
\delta\bfg &\gets \delta\bfg ~.
\end{align}%
\end{subequations}

Here, $\bfv_\bfi$, $\bftheta_\bfi$, $\bfa_\bfi$ and $\bfomega_\bfi$ are the random impulses applied to the velocity, orientation and bias estimates, modeled by white Gaussian processes. 
Their mean is zero, and their covariances matrices are obtained by integrating the covariances of $\bfa_n$, $\bfomega_n$, $\bfa_w$ and $\bfomega_w$ over the step time $\Dt$ (see \appRef{sec:IntNoise}),
\begin{align}
\bfV_\bfi &= \sigma_{\tilde\bfa_n}^2\Dt^2\bfI \quad &&[m^2/s^2]\\
\Theta_\bfi &= \sigma_{\tilde\bfomega_n}^2\Dt^2\bfI \quad &&[rad^2] \\
\bfA_\bfi &= \sigma_{\bfa_w}^2\Dt\bfI \quad &&[m^2/s^4] \\
\bfOmega_\bfi &= \sigma_{\bfomega_w}^2\Dt\bfI \quad &&[rad^2/s^2] 
\end{align}%
where $\sigma_{\tilde\bfa_n}[m/s^2]$, $\sigma_{\tilde\bfomega_n}[rad/s]$, $\sigma_{\bfa_w}[m/s^2\sqrt{s}]$ and $\sigma_{\bfomega_w}[rad/s\sqrt{s}]$ are to be determined from the information in the IMU datasheet, or from experimental measurements.

\subsubsection{The error-state Jacobian and perturbation matrices}

The Jacobians are obtained by simple inspection of the error-state differences equations in the previous section. 

To write these equations in compact form, we consider the nominal state vector $\bfx$, the error state vector $\delta\bfx$, the input vector $\bfu_m$, and the perturbation impulses vector $\bfi$, as follows (see \appRef{sec:pertImpulses} for details and justifications),
\begin{equation}
\bfx=\begin{bmatrix}\bfp\\\bfv\\\bfq\\\bfa_b\\\bfomega_b\\\bfg\end{bmatrix} \quad, \quad
\delta\bfx=\begin{bmatrix}\delta\bfp\\\delta\bfv\\\delta\bftheta\\\delta\bfa_b\\\delta\bfomega_b\\\delta\bfg\end{bmatrix} \quad, \quad
\bfu_m = \begin{bmatrix}
\bfa_m \\
\bfomega_m
\end{bmatrix} 
\quad,  \quad
\bfi = \begin{bmatrix}
\bfv_\bfi \\
\bftheta_\bfi \\
\bfa_\bfi \\
\bfomega_\bfi
\end{bmatrix}
\end{equation}

\bigskip
The error-state system is now
\begin{equation}
\delta\bfx \gets f(\bfx,\delta\bfx,\bfu_m,\bfi)=\bfF_\bfx(\bfx, \bfu_m)\tdot \delta\bfx+\bfF_\bfi\tdot\bfi,
\end{equation}
The ESKF prediction equations are written:
\begin{align}
\hat{\delta\bfx} &\gets \bfF_\bfx(\bfx, \bfu_m)\tdot \hat{\delta\bfx} \label{equ:errorMeanPred} \\
\bfP &\gets \bfF_\bfx\,\bfP\,\bfF_\bfx\tr + \bfF_\bfi\,\bfQ_\bfi\,\bfF_\bfi\tr \label{equ:errorCovPred} ~,
\end{align}%
where $\delta\bfx\sim\cN\{\hat{\delta\bfx},\bfP\}$\footnote{$x\sim\cN\{\mu,\Sigma\}$ means that $x$ is a Gaussian random variable with mean and covariances matrix specified by $\mu$ and $\Sigma$.}; $\bfF_\bfx$ and $\bfF_\bfi$ are the Jacobians of $f()$ \wrt the error and perturbation vectors; and $\bfQ_\bfi$ is the covariances matrix of the perturbation impulses. 

The expressions of the Jacobian and covariances matrices above are detailed below. 
All state-related values appearing herein are extracted directly from the nominal state.
\begin{equation} \label{equ:Fx_local_euler}
\bfF_\bfx = \pjac{f}{\delta\bfx}{\bfx,\bfu_m} = \begin{bmatrix}
\bfI & \bfI\Dt & 0                             & 0               & 0                     & 0 \\
0 & \bfI    & -\bfR\hatx{\bfa_m-\bfa_b}\Dt     & -\bfR\Dt            & 0                     & \bfI\Dt \\
0 & 0    & \bfR\tr\{(\bfomega_m-\bfomega_b)\Dt\}   & 0               & -\bfI\Dt                  & 0 \\
0 & 0    & 0                             & \bfI & 0                     & 0 \\
0 & 0    & 0                             & 0               & \bfI  & 0 \\
0 & 0    & 0                             & 0               & 0                     & \bfI \\
\end{bmatrix}
\end{equation}%
\begin{equation}
\bfF_\bfi = \pjac{f}{\bfi}{\bfx,\bfu_m} = \begin{bmatrix}
0 & 0 & 0 & 0 \\
\bfI & 0 & 0 & 0 \\
0 & \bfI & 0 & 0 \\
0 & 0 & \bfI & 0 \\
0 & 0 & 0 & \bfI \\
0 & 0 & 0 & 0 
\end{bmatrix}  
\quad,\quad
\bfQ_\bfi = \begin{bmatrix}
\bfV_\bfi & 0        & 0      & 0 \\ 
0      & \bfTheta_\bfi & 0      & 0 \\ 
0      & 0        & \bfA_\bfi & 0 \\ 
0      & 0        & 0      & \bfOmega_\bfi 
\end{bmatrix}~.
\end{equation}%

Please note particularly that $\bfF_\bfx$ is the system's transition matrix, which can be approximated to different levels of precision in a number of ways. 
We showed here one of its simplest forms (the Euler form). 
Se Appendices \ref{sec:ClosedFormInt} to \ref{sec:TranMatRK} for further reference.

Please note also that, being the mean of the error $\delta\bfx$ initialized to zero, the linear equation \eqRef{equ:errorMeanPred} always returns zero. 
You should of course skip line \eqRef{equ:errorMeanPred} in your code. 
I recommend that you write it, though, but that you comment it out so that you are sure you did not forget anything.

And please note, finally, that you should NOT skip the covariance prediction \eqRef{equ:errorCovPred}!! In effect, the term $\bfF_\bfi\,\bfQ_\bfi\,\bfF_\bfi\tr$ is not null and therefore this covariance grows continuously -- as it must be in any prediction step.

\section{Fusing IMU with complementary sensory data}
\label{sec:fusion}

At the arrival of other kind of information than IMU, such as GPS or vision, we proceed to correct the ESKF. 
In a well-designed system, this should render the IMU biases observable and allow the ESKF to correctly estimate them. 
There are a myriad of possibilities, the most popular ones being GPS + IMU, monocular vision + IMU, and stereo vision + IMU. 
In recent years, the combination of visual sensors with IMU has attracted a lot of attention, and thus generated a lot of scientific activity. 
These vision + IMU setups are very interesting for use in GPS-denied environments, and can be implemented on mobile devices (typically smart phones), but also on UAVs and other small, agile platforms.

\bigskip

While the IMU information has served so far to make predictions to the ESKF, this other information is used to correct the filter, and thus observe the IMU bias errors. 
The correction consists of three steps:
\begin{enumerate}
\item
 observation of the error-state via filter correction, 
\item
 injection of the observed errors into the nominal state, and
\item	
 reset of the error-state. 
\end{enumerate}%
These steps are developed in the following sections.

\subsection{Observation of the error state via filter correction}

Suppose as usual that we have a sensor that delivers information that depends on the state, such as
\begin{equation}
\bfy = h(\bfx_t) + v~,
\end{equation}
where $h()$ is a general nonlinear function of the system state (the true state), and $v$ is a white Gaussian noise with covariance $\bfV$,
\begin{equation}
v\sim\cN\{0,\bfV\}~.
\end{equation}

Our filter is estimating the error state, and therefore the filter correction equations\footnote{We give the simplest form of the covariance update, $\bfP \gets
 (\bfI-\bfK\bfH)\bfP$. 
 This form is known to have poor numerical stability, as its outcome is not guaranteed to be symmetric nor positive definite. 
 The reader is free to use more stable forms such as 1) the symmetric form $\bfP\gets\bfP-\bfK(\bfH\bfP\bfH\tr+\bfV)\bfK\tr$ and 2) the symmetric and positive \emph{Joseph} form $\bfP \gets (\bfI-\bfK\bfH)\bfP(\bfI-\bfK\bfH)\tr+\bfK\bfV\bfK\tr$.}, 
\begin{align}
\bfK &= \bfP\bfH\tr(\bfH\bfP\bfH\tr+\bfV)\inv \\
\hat{\delta\bfx} &\gets \bfK(\bfy-h(\hat\bfx_t)) \\
\bfP &\gets (\bfI-\bfK\bfH)\bfP
\end{align}%
require the Jacobian matrix $\bfH$ to be defined \wrt the error state $\delta\bfx$, and evaluated at the best true-state estimate $\hat\bfx_t=\bfx\oplus\hat{\delta\bfx}$. 
As the error state mean is zero at this stage (we have not observed it yet), we have $\hat\bfx_t = \bfx$ and we can use the nominal error $\bfx$ as the evaluation point, leading to
\begin{equation}
\bfH \equiv \pjac{h}{\delta\bfx}{\bfx}~.
\end{equation}

\subsubsection{Jacobian computation for the filter correction}

The Jacobian above might be computed in a number of ways. 
The most illustrative one is by making use of the chain rule, 
\begin{equation}
\bfH \triangleq \pjac{h}{\delta\bfx}{\bfx} 
= \pjac{h}{\bfx_t}{\bfx}\pjac{\bfx_t}{\delta\bfx}{\bfx} = \bfH_\bfx\ \bfX_{\delta\bfx}~.
\end{equation}
Here, $\bfH_\bfx\triangleq\pjac{h}{\bfx_t}{\bfx}$ is the standard Jacobian of $h()$ with respect to its own argument (\ie, the Jacobian one would use in a regular EKF). 
This first part of the chain rule depends on the measurement function of the particular sensor used, and is not presented here. 

The second part, $\bfX_{\delta\bfx}\triangleq\pjac{\bfx_t}{\delta\bfx}{\bfx}$, is the Jacobian of the true state \wrt the error state. 
This part can be derived here as it only depends on the ESKF composition of states. We have the derivatives,
\begin{equation}
\bfX_{\delta\bfx} = 
\begin{bmatrix}
\dpar{(\bfp+\delta\bfp)}{\delta\bfp} &&&&& \\ 
& \dpar{(\bfv+\delta\bfv)}{\delta\bfv} &&&0& \\ 
&& \dpar{(\bfq\otimes\delta\bfq)}{\delta\bftheta} &&& \\ 
&&& \dpar{(\bfa_b+\delta\bfa_b)}{\delta\bfa_b} && \\ 
&0&&& \dpar{(\bfomega_b+\delta\bfomega_b)}{\delta\bfomega_b} & \\ 
&&&&& \dpar{(\bfg+\delta\bfg)}{\delta\bfg} 
\end{bmatrix}
\end{equation}
which results in all identity $3\times3$ blocks (for example, $\dpar{(\bfp+\delta\bfp)}{\delta\bfp}=\bfI_3$) except for the $4\times3$ quaternion term $\bfQ_{\delta\bftheta} = \dparil{(\bfq\otimes\delta\bfq)}{\delta\bftheta}$. Therefore we have the form,
\begin{equation}
\bfX_{\delta\bfx}\triangleq\pjac{\bfx_t}{\delta\bfx}{\bfx} = \begin{bmatrix}
\bfI_6 & 0 & 0 \\
0 &  \bfQ_{\delta\bftheta} & 0 \\
0 & 0 & \bfI_9
\end{bmatrix}
\end{equation}

Using the chain rule, equations \eqsRef{equ:quatMatProd}{equ:quatMatrix}, and the  limit $\delta\bfq \underset{\delta\bftheta\to 0}{\longrightarrow} \begin{bmatrix}
1 \\ \frac12\delta\bftheta
\end{bmatrix}$, the quaternion term $\bfQ_{\delta\bftheta}$ may be derived as follows,
{
\setlength{\arraycolsep}{2pt}
\begin{align*}
\bfQ_{\delta\bftheta} \triangleq \pjac{(\bfq\otimes\delta\bfq)}{\delta\bftheta}{\bfq} 
&=\pjac{(\bfq\otimes\delta\bfq)}{\delta\bfq}{\bfq} \pjac{\delta\bfq}{\delta\bftheta}{\hat{\delta\bftheta}=0} \\
&= \pjac{([\bfq]_L\delta\bfq)}{\delta\bfq}{\bfq} \pjac{\begin{bmatrix}
1 \\ \frac12\delta\bftheta
\end{bmatrix}}{\delta\bftheta}{\hat{\delta\bftheta}=0} \\
&= [\bfq]_L\,\frac12\begin{bmatrix}
0 & 0 & 0 \\
1 & 0 & 0 \\
0 & 1 & 0 \\
0 & 0 & 1 \\
\end{bmatrix} ~,
\end{align*}
which leads to
\begin{equation}
\bfQ_{\delta\bftheta}
= \frac{1}{2}\begin{bmatrix}
-q_x &-q_y &-q_z\\
 q_w &-q_z & q_y\\
 q_z & q_w &-q_x\\
-q_y & q_x & q_w\\
\end{bmatrix}~.
\end{equation}•

\subsection{Injection of the observed error into the nominal state}

After the ESKF update, the nominal state gets updated with the observed error state using the appropriate compositions (sums or quaternion products, see \tabRef{tab:errorstatevar}),
\begin{equation}
\bfx \gets \bfx \oplus \hat{\delta\bfx}~, \label{equ:errorInjection}
\end{equation}
that is,
\begin{subequations}
\begin{align}
\bfp &\gets \bfp + \hat{\delta\bfp} \\
\bfv &\gets \bfv + \hat{\delta\bfv} \\
\bfq &\gets \bfq \otimes \bfq\{\hat{\delta\bftheta}\} \label{equ:quatErrorInjection}\\
\bfa_b &\gets \bfa_b + \hat{\delta\bfa_b} \\
\bfomega_b &\gets \bfomega_b + \hat{\delta\bfomega_b} \\
\bfg &\gets \bfg + \hat{\delta\bfg} 
\end{align}%
\end{subequations}%

\subsection{ESKF reset}

After error injection into the nominal state, the error state mean $\hat{\delta\bfx}$ gets reset. 
This is especially relevant for the orientation part, as the new orientation error will be expressed locally \wrt the orientation frame of the new nominal state. 
To make the ESKF update complete, the covariance of the error needs to be updated according to this modification.

\bigskip

Let us call the error reset function $g()$. 
It is written as follows,
\begin{equation}
\delta\bfx \gets g(\delta\bfx) = \delta\bfx \ominus \hat{\delta\bfx}~, \label{equ:resetFcn}
\end{equation}
where $\ominus$ stands for the composition inverse of $\oplus$. 
The ESKF error reset operation is thus,
\begin{align}
\hat{\delta\bfx} &\gets 0 \\
\bfP &\gets \bfG\,\bfP\,\bfG\tr~.
\end{align}
where $\bfG$ is the Jacobian matrix defined by,
\begin{equation}
\bfG \triangleq \pjac{g}{\delta\bfx}{\hat{\delta\bfx}}~.
\end{equation}

Similarly to what happened with the update Jacobian above, this Jacobian is the identity on all diagonal blocks except in the orientation error. 
We give here the full expression and proceed in the following section with the derivation of the orientation error block, $\partial\delta\bftheta^+/\partial\delta\bftheta = \bfI - \hatx{\frac12\hat{\delta\bftheta}}$,
\begin{equation}
\bfG = \begin{bmatrix}
\bfI_6 & 0 & 0 \\
0 & \bfI - \hatx{\frac12\hat{\delta\bftheta}} & 0 \\
0 & 0 & \bfI_9
\end{bmatrix}~.
\end{equation}

In major cases, the error term $\hat{\delta\bftheta}$ can be neglected, leading simply to a Jacobian $\bfG=\bfI_{18}$, and thus to a trivial error reset. 
This is what most implementations of the ESKF do.
The expression here provided should produce more precise results, which might be of interest for reducing long-term error drift in odometry systems.

\subsubsection{Jacobian of the reset operation \wrt the orientation error}

We want to obtain the expression of the new angular error $\delta\bftheta^+$ \wrt the old error $\delta\bftheta$ and the observed error $\hat{\delta\bftheta}$. Consider these facts:
\begin{itemize}
\item
The true orientation does not change on error reset, \ie, $\bfq_t^+ = \bfq_t$. This gives:
\begin{equation}
\bfq^+\ot\delta\bfq^+ = \bfq\ot \delta\bfq~.
\end{equation}
\item
The observed error mean has been injected into the nominal state (see \eqRef{equ:quatErrorInjection} and \eqRef{equ:rotComposition}):
\begin{equation}
\bfq^+ = \bfq\ot \hat{\delta\bfq}~.
\end{equation}
\end{itemize}

Combining both identities we obtain an expression of $\delta\bfq^+$,
\begin{equation}
\delta\bfq^+ 
= (\bfq^+)^* \ot \bfq \ot \delta\bfq 
= (\bfq \ot \hat{\delta\bfq})^* \ot \bfq \ot \delta\bfq 
= \hat{\delta\bfq}^* \ot \delta\bfq 
= [\hat{\delta\bfq}^*]_L\cdot\delta\bfq~.
\end{equation}
Considering that
$\hat{\delta\bfq}^* \approx \begin{bmatrix}
1 \\ -\frac12\hat{\delta\bftheta}
\end{bmatrix}$,
the identity above can be expanded as
\begin{equation}
\begin{bmatrix}
1 \\ \frac12\delta\bftheta^+
\end{bmatrix}
=
\begin{bmatrix}
1                  & \frac12\hat{\delta\bftheta}\tr \\
-\frac12\hat{\delta\bftheta} & \bfI - \hatx{\frac12\hat{\delta\bftheta}}
\end{bmatrix}
\cdot
\begin{bmatrix}
1 \\ \frac12\delta\bftheta
\end{bmatrix} + \cO(\norm{\delta\bftheta}^2) ~,
\end{equation}
which gives one scalar- and one vector- equations,
\begin{subequations}
\begin{align}
\frac14\hat{\delta\bftheta}\tr\delta\bftheta &= \cO(\norm{\delta\bftheta}^2)\\
\delta\bftheta^+ &= -\hat{\delta\bftheta} + \left(\bfI - \hatx{\frac12\hat{\delta\bftheta}}\right)\delta\bftheta + \cO(\norm{\delta\bftheta}^2)~,
\end{align}%
\end{subequations}
among which the first one is not very informative in that it is only a relation of infinitesimals. 
One can show from the second equation that $\hat{\delta\bftheta}^+ = 0$, which is what we expect from the reset operation. 
The Jacobian is obtained by simple inspection,
\begin{equation}
\eqbox{\dpar{\delta\bftheta^+}{\delta\bftheta} = \bfI - \hatx{\frac12\hat{\delta\bftheta}}}~.
\end{equation}


\section{The ESKF using global angular errors}
\label{sec:ESKFglobal}

We explore in this section the implications of having the angular error defined in the global reference, as opposed to the local definition we have used so far. 
We retrace the development of sections \ref{sec:es-kinematics} and \ref{sec:fusion}, and particularize the subsections that present changes \wrt the new definition.

A global definition of the angular error $\delta\bftheta$ implies a composition \emph{on the left hand side}, \ie,
\begin{equation*}
\bfq_t = \delta\bfq\ \ot \bfq = \bfq\{\delta\bftheta\} \ot \bfq~.
\end{equation*}

We remark for the sake of completeness that we keep the local definition of the angular rates vector $\bfomega$, \ie, $\dot\bfq=\frac12\bfq\ot\bfomega$ in continuous time, and therefore $\bfq \gets \bfq\ot\bfq\{\bfomega\Dt\}$ in discrete time, regardless of the angular error being defined globally. 
This is so for convenience, as the measure of the angular rates provided by the gyrometers is in body frame, that is, local.

\subsection{System kinematics in continuous time}

\subsubsection{The true- and nominal-state kinematics}

True and nominal kinematics do not involve errors and their equations are unchanged.

\subsubsection{The error-state kinematics}

We start by writing the equations of the error-state kinematics, and proceed afterwards with comments and proofs.
\begin{subequations}\label{equ:efullglobal}
\begin{align}
\dot{\delta\bfp} &= \delta\bfv \label{equ:epglobal} \\
\dot{\delta\bfv} &= -\hatx{\bfR(\bfa_m-b\bfa_b)}\delta\bftheta - \bfR\delta\bfa_b + \delta\bfg - \bfR\bfa_n \label{equ:evglobal}\\
\dot{\delta\bftheta} &= -\bfR\delta\bfomega_b - \bfR\bfomega_n \label{equ:etglobal}\\
\dot{\delta\bfa_b} &= \bfa_w \label{equ:eabglobal}\\
\dot{\delta\bfomega_b} &= \bfomega_w \label{equ:ewbglobal}\\
\dot{\delta\bfg} &= 0 ~, \label{equ:egglobal}
\end{align}%
\end{subequations}%
where, again, all equations except those of $\dot{\delta\bfv}$ and $\dot{\delta\bftheta}$ are trivial. 
The non-trivial expressions are developed below.

\paragraph{Equation \eqRef{equ:evglobal}: The linear velocity error.}

We wish to determine $\dot{\delta\bfv}$, the dynamics of the velocity errors. 
We start with the following relations
\begin{align}
\bfR_t &= (\bfI+\hatx{\delta\bftheta})\bfR  + O(\norm{\delta\bftheta}^2) \label{equ:Rtglobal}\\
\dot\bfv &= \bfR\bfa_\cB + \bfg~, \label{equ:vdot2global}
\end{align}%
where \eqRef{equ:Rtglobal} is the small-signal approximation of $\bfR_t$ using a globally defined error, and in \eqRef{equ:vdot2global} we introduced $\bfa_\cB$ and $\delta\bfa_\cB$ as the large- and small- signal accelerations in body frame, defined in\eqRef{equ:nomacc} and \eqRef{equ:pertacc}, as we did for the locally-defined case.

We proceed by writing the expression \eqRef{equ:vel} of $\dot\bfv_t$ in two different forms (left and right developments), where the terms $O(\norm{\delta\bftheta}^2)$ have been ignored,
\begin{align*}
\dot\bfv+\dot{\delta\bfv} =& \eqbox{\dot\bfv_t} = (\bfI + \hatx{\delta\bftheta})\bfR(\bfa_\cB+\delta\bfa_\cB) + \bfg + \delta\bfg \\
\bfR\bfa_\cB+\bfg+\dot{\delta\bfv} =&
~~~~~~= \bfR\bfa_\cB+\bfR\delta\bfa_\cB + \hatx{\delta\bftheta}\bfR\bfa_\cB + \hatx{\delta\bftheta}\bfR\delta\bfa_\cB + \bfg + \delta\bfg 
\end{align*}%
This leads after removing $\bfR\bfa_\cB+\bfg$ from left and right to
\begin{equation}
\dot{\delta\bfv} = \bfR\delta\bfa_\cB + \hatx{\delta\bftheta}\bfR(\bfa_\cB + \delta\bfa_\cB) + \delta\bfg
\end{equation}%
Eliminating the second order terms and reorganizing some cross-products (with $\hatx{\bfa}\bfb=-\hatx{\bfb}\bfa$), we get
\begin{equation}
\dot{\delta\bfv} = \bfR\delta\bfa_\cB - \hatx{\bfR\bfa_\cB}\delta\bftheta + \delta\bfg~,
\end{equation}%
and finally, recalling \eqRef{equ:nomacc} and \eqRef{equ:pertacc} and rearranging, we obtain  the expression of the derivative of the velocity error,
\begin{equation}
\eqbox{{\dot{\delta\bfv} = -\hatx{\bfR(\bfa_m-\bfa_b)}\delta\bftheta - \bfR\delta\bfa_b + \delta\bfg - \bfR\bfa_n }}
\end{equation}%

\paragraph{Equation \eqRef{equ:etglobal}: The orientation error.}

We start by writing the true- and nominal- definitions of the quaternion derivatives,
\begin{align}
\dot{\bfq_t} &= \frac{1}{2}\bfq_t\ot\bfomega_t \\
\dot{\bfq} &= \frac{1}{2}\bfq\ot\bfomega~,
\end{align}%
and remind that we are using a globally-defined angular error, \ie,
\begin{equation}
\bfq_t = \delta\bfq\ot\bfq~.
\end{equation}
As we did for the locally-defined error case, we also group large- and small-signal angular rates \eqsRef{equ:nomangrate}{equ:pertangrate}. 
We proceed by computing $\dot{\bfq_t}$ by two different means (left and right developments),
\begin{align*}
\dot{{({\delta\bfq}\ot\bfq)}} =& \eqbox{\dot{\bfq_t}} = \frac{1}{2}\bfq_t\ot\bfomega_t \\
\dot{\delta\bfq}\ot{\bfq} + \delta\bfq\ot\dot{{\bfq}} =&~~~~~~= \frac{1}{2}\delta\bfq\ot\bfq\ot\bfomega_t \\
\dot{\delta\bfq}\ot{\bfq} + \frac{1}{2}\delta\bfq\ot\bfq\ot\bfomega =&&  
\end{align*}%
Having $\bfomega_t = \bfomega + \delta\bfomega$, this reduces to
\begin{equation}
\dot{\delta\bfq}\ot{\bfq} = \frac{1}{2}\delta\bfq\ot\bfq\ot\delta\bfomega~.
\end{equation}
Right-multiplying left and right terms by $\bfq^*$, and recalling that $\bfq\ot\delta\bfomega\ot\bfq^* \equiv \bfR\delta\bfomega$, we can further develop as follows,
\begin{align}
\dot{\delta\bfq} &= \frac{1}{2}\delta\bfq\ot\bfq\ot\delta\bfomega\ot\bfq^* \nonumber \\
&= \frac{1}{2}\delta\bfq\ot(\bfR\delta\bfomega) \nonumber \\
&= \frac{1}{2}\delta\bfq\ot\delta\bfomega_G~,
\end{align}%
with $\delta\bfomega_G \triangleq \bfR\delta\bfomega$ the small-signal angular rate expressed in the global frame. Then,
\begin{align}
\begin{bmatrix}
0\\\dot{\delta\bftheta}
\end{bmatrix} = \eqbox{2\dot{{\delta\bfq}}} 
&= {\delta\bfq}\ot\delta\bfomega_G \nonumber \\
&= \Omega(\delta\bfomega_G)\,\delta\bfq \nonumber \\
&= \begin{bmatrix}
0 & -\delta\bfomega_G\tr \\
\delta\bfomega_G & -\hatx{\delta\bfomega_G} 
\end{bmatrix}\begin{bmatrix}
1 \\
\delta\bftheta/2
\end{bmatrix} + O(\norm{\delta\bftheta}^2) ~,
\end{align}%
which results in one scalar- and one vector- equalities
\begin{subequations}
\begin{align}
0 &= \delta\bfomega_G\tr\delta\bftheta + O(|\delta\bftheta|^2) \\
\dot{\delta\bftheta} &= \delta\bfomega_G - \frac{1}{2}\hatx{\delta\bfomega_G}\delta\bftheta + O(\norm{\delta\bftheta}^2).
\end{align}%
\end{subequations}
The first equation leads to $\delta\bfomega_G\tr\delta\bftheta = O(\norm{\delta\bftheta}^2)$, which is formed by second-order infinitesimals, not very useful. 
The second equation yields, after neglecting all second-order terms,
\begin{equation}
\dot{\delta\bftheta} = \delta\bfomega_G = \bfR\delta\bfomega ~.
\end{equation}%
Finally, recalling 
\eqRef{equ:pertangrate}, we obtain the linearized dynamics of the global angular error,
\begin{equation}
\eqbox{\dot{\delta\bftheta} = - \bfR\delta\bfomega_b - \bfR\bfomega_n}\ .
\end{equation}%

\subsection{System kinematics in discrete time}
\subsubsection{The nominal state}
The nominal state equations do not involve errors and are therefore the same as in the case where the orientation error is defined locally. 

\subsubsection{The error state}

Using Euler integration, we obtain the following set of differences equations,
\begin{subequations}
\begin{align}
\delta\bfp &\gets \delta\bfp + \delta\bfv\,\Dt \\
\delta\bfv &\gets \delta\bfv + (-\hatx{\bfR(\bfa_{m}-\bfa_{b})}\delta\bftheta - \bfR\delta\bfa_{b} + \delta\bfg)\Dt + \bfv_\bfi \\
\delta\bftheta &\gets \delta\bftheta - \bfR\delta\bfomega_{b} \Dt + \bftheta_\bfi \\
\delta\bfa_{b} &\gets  \delta\bfa_{b} + \bfa_\bfi \\
\delta\bfomega_{b} &\gets \delta\bfomega_{b} + \bfomega_\bfi \\
\delta\bfg &\gets \delta\bfg .
\end{align}%
\end{subequations}

\subsubsection{The error state Jacobian and perturbation matrices}

The Transition matrix is obtained by simple inspection of the equations above,
\begin{equation}
\bfF_\bfx = 
\begin{bmatrix}
\bfI & \bfI\Dt & 0                             & 0               & 0                     & 0 \\
0 & \bfI    & \eqbox{-\hatx{\bfR(\bfa_m-\bfa_b)}\Dt}     & -\bfR\Dt            & 0                     & \bfI\Dt \\
0 & 0    & \eqbox{\bfI}   & 0               & \eqbox{-\bfR\Dt}                  & 0 \\
0 & 0    & 0                             & \bfI & 0                     & 0 \\
0 & 0    & 0                             & 0               & \bfI  & 0 \\
0 & 0    & 0                             & 0               & 0                     & \bfI \\
\end{bmatrix}~.
\end{equation}

We observe three changes \wrt the case with a locally-defined angular error (compare the boxed terms in the Jacobian above to the ones in \eqRef{equ:Fx_local_euler}); these changes are summarized in \tabRef{tab:local_to_global}.

The perturbation Jacobian and the perturbation matrix are unchanged after considering isotropic noises and the developments of \appRef{sec:IntNoise},
\begin{equation}
\bfF_\bfi = 
\begin{bmatrix}
0 & 0 & 0 & 0 \\
\bfI & 0 & 0 & 0 \\
0 & \bfI & 0 & 0 \\
0 & 0 & \bfI & 0 \\
0 & 0 & 0 & \bfI \\
0 & 0 & 0 & 0 
\end{bmatrix}  
\quad,\quad
\bfQ_\bfi = \begin{bmatrix}
\bfV_\bfi & 0        & 0      & 0 \\ 
0      & \bfTheta_\bfi & 0      & 0 \\ 
0      & 0        & \bfA_\bfi & 0 \\ 
0      & 0        & 0      & \bfOmega_\bfi 
\end{bmatrix}~.
\end{equation}%
%

\subsection{Fusing with complementary sensory data}

The fusing equations involving the ESKF machinery vary only slightly when considering global angular errors. 
We revise these variations in the error state observation via ESKF correction, the injection of the error into the nominal state, and the reset step.

\subsubsection{Error state observation}

The only difference \wrt the local error definition is in the Jacobian block of the observation function that relates the orientation to the angular error. 
This new block is developed below.

Using \eqsRef{equ:quatMatProd}{equ:quatMatrix} and the first-order approximation $\delta\bfq\rightarrow \begin{bmatrix}
1 \\ \frac12\delta\bftheta
\end{bmatrix}$, the quaternion term $\bfQ_{\delta\bftheta}$ may be derived as follows,
\begin{subequations}
\begin{align}
\bfQ_{\delta\bftheta} \triangleq \pjac{(\delta\bfq\otimes\bfq)}{\delta\bftheta}{\bfq} 
&=\pjac{(\delta\bfq\otimes\bfq)}{\delta\bfq}{\bfq} \pjac{\delta\bfq}{\delta\bftheta}{\hat{\delta\bftheta}=0} \\
&= [\bfq]_R\,\frac12\begin{bmatrix}
0 & 0 & 0 \\
1 & 0 & 0 \\
0 & 1 & 0 \\
0 & 0 & 1 \\
\end{bmatrix} \\
&= \frac{1}{2}\begin{bmatrix}
-q_x &-q_y &-q_z\\
 q_w & q_z &-q_y\\
-q_z & q_w & q_x\\
 q_y &-q_x & q_w\\
\end{bmatrix}~.
\end{align}
\end{subequations}

\subsubsection{Injection of the observed error into the nominal state}

The composition $\bfx \gets \bfx\oplus\hat{\delta\bfx}$ of the nominal and error states is depicted as follows,
\begin{subequations}
\begin{align}
\bfp &\gets \bfp + \delta\bfp \\
\bfv &\gets \bfv + \delta\bfv \\
\bfq &\gets \bfq\{\hat{\delta\bftheta}\} \ot \bfq \\ 
\bfa_b &\gets \bfa_b + \delta\bfa_b \\
\bfomega_b &\gets \bfomega_b + \delta\bfomega_b \\
\bfg &\gets \bfg + \delta\bfg ~.
\end{align}%
\end{subequations}
where only the equation for the quaternion update has been affected. 
This is summarized in \tabRef{tab:local_to_global}.

\subsubsection{ESKF reset}

The ESKF error mean is reset, and the covariance updated, according to,
\begin{align}
\hat{\delta\bfx} &\gets 0 \\
\bfP &\gets \bfG\bfP\bfG\tr
\end{align}%
with the Jacobian
\begin{equation}
\bfG = \begin{bmatrix}
\bfI_6 & 0 & 0 \\
0 & \bfI+\hatx{\hat{\frac12\delta\bftheta}} & 0 \\
0 & 0 & \bfI_9
\end{bmatrix}
\end{equation}
whose non-trivial term is developed as follows. 
Our goal is to obtain the expression of the new angular error $\delta\bftheta^+$ \wrt the old error $\delta\bftheta$. 
We consider these facts:
\begin{itemize}
\item
The true orientation does not change on error reset, \ie, $\bfq_t^+ \equiv \bfq_t$. This gives:
\begin{equation}
\delta\bfq^+\ot\bfq^+ = \delta\bfq\ot \bfq~.
\end{equation}
\item
The observed error mean has been injected into the nominal state (see \eqRef{equ:quatErrorInjection} and \eqRef{equ:rotComposition}):
\begin{equation}
\bfq^+ = \hat{\delta\bfq} \ot \bfq ~.
\end{equation}
\end{itemize}

Combining both identities we obtain an expression of the new orientation error \wrt the old one and the observed error $\hat{\delta\bfq}$,
\begin{equation}
\delta\bfq^+ = \delta\bfq \ot \hat{\delta\bfq}^*  = [\hat{\delta\bfq}^*]_R\cdot\delta\bfq~.
\end{equation}
Considering that
$\hat{\delta\bfq}^* \approx \begin{bmatrix}
1 \\ -\frac12\hat{\delta\bftheta}
\end{bmatrix}$,
the identity above can be expanded as
\begin{equation}
\begin{bmatrix}
1 \\ \frac12\delta\bftheta^+
\end{bmatrix}
=
\begin{bmatrix}
1                  & \frac12\hat{\delta\bftheta}\tr \\
-\frac12\hat{\delta\bftheta} & \bfI + \hatx{\frac12\hat{\delta\bftheta}}
\end{bmatrix}
\cdot
\begin{bmatrix}
1 \\ \frac12\delta\bftheta
\end{bmatrix} + \cO(\norm{\delta\bftheta}^2)
\end{equation}
which gives one scalar- and one vector- equations,
\begin{subequations}
\begin{align}
\frac14\hat{\delta\bftheta}\tr\delta\bftheta &= \cO(\norm{\delta\bftheta}^2)\\
\delta\bftheta^+ &= -\hat{\delta\bftheta} + \left(\bfI + \hatx{\frac12\hat{\delta\bftheta}}\right)\delta\bftheta + \cO(\norm{\delta\bftheta}^2)
\end{align}%
\end{subequations}
among which the first one is not very informative in that it is only a relation of infinitesimals. 
One can show from the second equation that $\hat{\delta\bftheta}^+ = 0$, which is what we expect from the reset operation. 
The Jacobian is obtained by simple inspection,
\begin{equation}
\eqbox{\dpar{\delta\bftheta^+}{\delta\bftheta} = \bfI + \hatx{\frac12\hat{\delta\bftheta}}}~.
\end{equation}

The difference \wrt the local error case is summarized in \tabRef{tab:local_to_global}.

\begin{table*}[tb]
\renewcommand{\arraystretch}{1.7}
\caption{Algorithm modifications related to the definition of the orientation errors.}
\label{tab:local_to_global}
\vspace{1ex}
\centering
\begin{tabular}{|c|c|c|c|}
\hline
Context & Item & local angular error & global angular error \\
\hline
\hline
Error composition & $\bfq_t$ & $\bfq_t = \bfq\ot\delta\bfq$ & $\bfq_t = \delta\bfq\ot\bfq$ \\
\hline
\hline
\multirow{3}{*}{Euler integration} & $\dparil{\delta\bfv^+}{\delta\bftheta}$ & $-\bfR\hatx{\bfa_m-\bfa_b}\Dt$ & $-\hatx{\bfR(\bfa_m-\bfa_b)}\Dt$ \\
&$\dparil{\delta\bftheta^+}{\delta\bftheta}$ & $\bfR\tr\{(\bfomega_m-\bfomega_b)\Dt\}$ & $\bfI$ \\
&$\dparil{\delta\bftheta^+}{\delta\bfomega_b}$ & $-\bfI\Dt$ &  $-\bfR\Dt$ \\
\hline
\hline
Error observation & $
\bfQ_{\delta\bftheta}
$ & $\frac12\begin{bmatrix}
-q_x &-q_y &-q_z\\
 q_w &-q_z & q_y\\
 q_z & q_w &-q_x\\
-q_y & q_x & q_w\\
\end{bmatrix}$ & $\frac12\begin{bmatrix}
-q_x &-q_y &-q_z\\
 q_w & q_z &-q_y\\
-q_z & q_w & q_x\\
 q_y &-q_x & q_w\\
\end{bmatrix}$ \\
\hline
Error injection &  
& $\bfq \gets \bfq \ot \bfq\{\hat{\delta\bftheta}\}$ & $\bfq \gets \bfq\{\hat{\delta\bftheta}\} \ot \bfq$ \\ 
\hline
Error reset & $\dparil{\delta\bftheta^+}{\delta\bftheta}$ & $\bfI - \hatx{\frac12\hat{\delta\bftheta}}$ & $\bfI + \hatx{\frac12\hat{\delta\bftheta}}$ \\
\hline
\end{tabular}
\end{table*}


\appendix


\section{Runge-Kutta numerical integration methods}
\label{sec:NumInt}

We aim at integrating nonlinear differential equations of the form
\begin{equation}
\dot\bfx = f(t,\bfx)
\end{equation}
over a limited time interval $\Dt$, in order to convert them to a differences equation, \ie,
\begin{equation}
\bfx(t+\Dt) =  \bfx(t)+\int_t^{t+\Dt}f(\tau,\bfx(\tau))d\tau~,
\end{equation}
or equivalently, if we assume that $t_n = n\Dt$ and $\bfx_n\triangleq \bfx(t_n)$,
\begin{equation}
\bfx_{n+1} =  \bfx_n+\int_{n\Dt}^{(n+1)\Dt}f(\tau,\bfx(\tau))d\tau~.
\end{equation}

One of the most utilized family of methods is the Runge-Kutta methods (from now on, RK). 
These methods use several iterations to estimate the derivative over the interval, and then use this derivative to integrate over the step $\Dt$.

In the sections that follow, several RK methods are presented, from the simplest one to the most general one, and are named according to their most common name.

\bigskip
NOTE: All the material here is taken from the \emph{Runge-Kutta method} entry in the English Wikipedia.

\subsection{The Euler method}
\label{sec:Euler}

The Euler method assumes that the derivative $f(\cdot)$ is constant over the interval, and therefore
\begin{equation}
\eqbox{\bfx_{n+1} =  \bfx_n + \Dt\tdot f(t_n,\bfx_n)~.}
\end{equation}

Put as a general RK method, this corresponds to a single-stage method, which can be depicted as follows. 
Compute the derivative at the initial point,
\begin{equation}
k_1 = f(t_n,\bfx_n)~,
\end{equation}
and use it to compute the integrated value at the end point,
\begin{equation}
\bfx_{n+1} = \bfx_n + \Dt\tdot k_1~.
\end{equation}

\subsection{The midpoint method}

The midpoint method assumes that the derivative is the one at the midpoint of the interval, and makes one iteration to compute the value of $\bfx$ at this midpoint, \ie,
\begin{equation}
\eqbox{\bfx_{n+1} =  \bfx_n + \Dt\tdot f \Big( t_n + \frac{1}{2}\Dt \ ,\ \bfx_n + \frac{1}{2} \Dt \tdot f(t_n , \bfx_n) \Big)}~.
\end{equation}

The midpoint method can be explained as a two-step method as follows. 
First, use the Euler method to integrate until the midpoint, using $k_1$ as defined previously,
\begin{align}
k_1 &= f(t_n,\bfx_n) \\
\bfx(t_n+\tfrac12\Dt) &=  \bfx_n + \frac{1}{2} \Dt\tdot k_1~.
\end{align}%
Then use this value to evaluate the derivative at the midpoint, $k_2$, leading to the integration
\begin{align}
k_2 &= f(t_n+\tfrac12\Dt \ ,\ \bfx(t_n+\tfrac12\Dt)) \\
\bfx_{n+1} &=  \bfx_n + \Dt\tdot k_2~.
\end{align}%

\subsection{The RK4 method}

This is usually referred to as simply the Runge-Kutta method. 
It assumes evaluation values for $f()$ at the start, midpoint and end of the interval. 
And it uses four stages or iterations to compute the integral, with four derivatives, $k_1\dots k_4$, that are obtained sequentially. 
These derivatives, or \emph{slopes}, are then weight-averaged to obtain the 4th-order estimate of the derivative in the interval. 

The RK4 method is better specified as a small algorithm than a one-step formula like the two methods above. The RK4 integration step is,

\begin{equation}
\eqbox{\bfx_{n+1} = \bfx_n + \frac{\Dt}{6} \Big( k_1 + 2k_2 + 2k_3 + k_4 \Big) }~,
\end{equation}
that is, the increment is computed by assuming a slope which is the weighted average of the slopes $k_1,k_2,k_3,k_4$, with
\begin{align}
k_1 &= f(t_n, \bfx_n) \\
k_2 &= f\Big(t_n + \frac{1}{2}\Dt \ ,\ \bfx_n + \frac{\Dt}{2} k_1\Big) \\
k_3 &= f\Big(t_n + \frac{1}{2}\Dt \ ,\ \bfx_n + \frac{\Dt}{2} k_2\Big) \\
k_4 &= f\Big(t_n + \Dt \ ,\ \bfx_n + \Dt \tdot k_3\Big) ~.
\end{align}%

The different slopes have the following interpretation:
\begin{itemize}
\item $k_1$ is the slope at the beginning of the interval, using $\bfx_n$ , (Euler's method);
\item   $ k_2$ is the slope at the midpoint of the interval, using $\bfx_n + \tfrac12 \Dt\tdot k_1$, (midpoint method);
\item    $k_3$ is again the slope at the midpoint, but now using $\bfx_n + \tfrac12 \Dt\tdot k_2$;
\item    $k_4$ is the slope at the end of the interval, using $\bfx_n + \Dt\tdot k_3 $.

\end{itemize}

\subsection{General Runge-Kutta method}

More elaborated RK methods are possible. 
They aim at either reduce the error and/or increase stability. 
They take the general form
\begin{equation}
\eqbox{\bfx_{n+1} = \bfx_n + \Dt \sum_{i=1}^s b_i k_i} ~,
\end{equation}
where
\begin{equation}
k_i = f\Big( t_n+\Dt \tdot c_i ,  \bfx_n + \Dt \sum_{j=1}^s a_{ij}  k_j \Big)~,
\end{equation}
that is, the number of iterations (the order of the method) is $s$, the averaging weights are defined by $b_i$, the evaluation time instants by $c_i$, and the slopes $k_i$ are determined using the values $a_{ij}$. 
Depending on the structure of the terms $a_{ij}$, one can have \emph{explicit} or \emph{implicit} RK methods. 

\begin{itemize}
\item In explicit methods, all $k_i$ are computed sequentially, \ie, using only previously computed values. 
This implies that the matrix $[a_{ij}]$ is lower triangular with zero diagonal entries (\ie, $a_{ij}=0$ for $j\ge i$). 
Euler, midpoint and RK4 methods are explicit. 

\item Implicit methods have a full $[a_{ij}]$ matrix and require the solution of a linear set of equations to determine all $k_i$. 
They are therefore costlier to compute, but they are able to improve on accuracy and stability \wrt explicit methods.
\end{itemize}

Please refer to specialized documentation for more detailed information.


\section{Closed-form integration methods}
\label{sec:ClosedFormInt}

In many cases it is possible to arrive to a closed-form expression for the integration step. 
We consider now the case of a first-order linear differential equation,
\begin{equation}
\dot\bfx(t)=\bfA\tdot\bfx(t)~,
\end{equation}
that is, the relation is linear and constant over the interval. 
In such cases, the integration over the interval $[t_n , t_n+\Dt]$ results in
\begin{equation}
\bfx_{n+1}= e^{\bfA\tdot\Dt}\bfx_n = \Phi\bfx_n~,
\end{equation}
where $\Phi$ is known as the transition matrix. 
The Taylor expansion of this transition matrix is
\begin{equation}
\Phi= e^{\bfA\tdot\Dt}=\bfI + \bfA\Dt + \frac{1}{2} \bfA^2\Dt^2 + \frac{1}{3!}\bfA^3\Dt^3 + \dots   = \sum_{k=0}^\infty \frac{1}{k!}\bfA^k\Dt^k ~.
\label{equ:TaylorExp}
\end{equation}

When writing this series for known instances of $\bfA$, it is sometimes possible to identify known series in the result. 
This allows writing the resulting integration in closed form. 
A few examples follow.

\subsection{Integration of the angular error}
\label{sec:ClosedFormAngle}

For example, consider the angular error dynamics without bias and noise (a cleaned version of Eq.~\eqRef{equ:equat}),
\begin{equation}
\dot{\delta\bftheta} = -\hatx\bfomega\delta\bftheta 
\end{equation}
Its transition matrix can be written as a Taylor series,
\begin{align}
\Phi &= e^{-\hatx\bfomega\Dt} \label{equ:phiexp}\\
&= \bfI -\hatx\bfomega\Dt + \frac12\hatx\bfomega^2\Dt^2 - \frac{1}{3!}\hatx\bfomega^3\Dt^3 + \frac{1}{4!}\hatx\bfomega^4\Dt^4 - \dots 
\end{align}%
Now defining $\bfomega\Dt \triangleq \bfu \Delta\theta$, the unitary axis of rotation and the rotated angle, and applying \eqRef{equ:prop2}, we can group terms and get
\begin{align}
\Phi 
&= \bfI -\hatx{\bfu}\Delta\theta + \frac12\hatx{\bfu}^2\Delta\theta^2 - \frac{1}{3!}\hatx{\bfu}^3\Delta\theta^3 + \frac{1}{4!}\hatx{\bfy}^4\Delta\theta^4 - \dots \nonumber\\
&= \bfI - \hatx{\bfu}\left(\Delta\theta - \frac{\Delta\theta^3}{3!} + \frac{\Delta\theta^5}{5!} - \cdots\right) + \hatx{\bfu}^2\left(\frac{\Delta\theta^2}{2!} - \frac{\Delta\theta^4}{4!}  + \frac{\Delta\theta^6}{6!} - \cdots\right) \nonumber\\
&= \bfI - \hatx{\bfu}\sin\Delta\theta + \hatx{\bfu}^2(1-\cos\Delta\theta) ~,
\end{align}%
which is a closed-form solution.

\bigskip
This solution corresponds to a rotation matrix, $\Phi=\bfR\{-\bfu\Delta\theta\} = \bfR\{\bfomega\Dt\}\tr$, according to the Rodrigues rotation formula \eqRef{equ:rodrigues}, a result that could be obtained by direct inspection of \eqRef{equ:phiexp} and recalling \eqRef{equ:vectomat}. 
Let us therefore write this as the final closed-form result,
\begin{equation}
\eqbox{\Phi=\bfR\{\bfomega\Dt\}\tr}~.
\end{equation}

\subsection{Simplified IMU example}
\label{sec:IMUexample}

Consider the simplified, IMU driven system with error-state dynamics governed by,
\begin{subequations}
\begin{align}
\dot{\delta\bfp}   &= \delta\bfv \\
\dot{\delta\bfv}   &= -\bfR\hatx{\bfa}\,\delta\bftheta \\
\dot{\delta\bftheta} &= -\hatx\bfomega\delta\bftheta~,
\end{align}
\end{subequations}%
where $(\bfa,\bfomega)$ are the IMU readings, and we have obviated gravity and sensor biases. 
This system is defined by the state vector and the dynamic matrix,
\begin{equation}
\bfx = \begin{bmatrix}
\delta\bfp \\ \delta\bfv \\ \delta\bftheta
\end{bmatrix}
\qquad
\bfA = \begin{bmatrix}
0 & \bfP_\bfv & 0 \\
0 & 0 & \bfV_\bftheta \\
0 & 0 & \Theta_\bftheta
\end{bmatrix}~.
\end{equation}
with
\begin{align}
\bfP_\bfv &= \bfI \\
\bfV_\bftheta &= -\bfR\hatx{\bfa} \\
\Theta_\bftheta &= -\hatx\bfomega \label{equ:ThetaThetaDef}
\end{align}%

Its integration with a step time $\Dt$ is $\bfx_{n+1}= e^{(\bfA\Dt)}\tdot\bfx_n=\Phi\tdot\bfx_n$. 
The transition matrix $\Phi$ admits a Taylor development \eqRef{equ:TaylorExp}, in increasing powers of $\bfA\Dt$.
We can write a few powers of $\bfA$ to get an illustration of their general form,
\begin{equation}
\bfA \!=\! \begin{bmatrix}
0 & \bfP_\bfv & 0 \\
0 & 0 & \bfV_\bftheta \\
0 & 0 & \Theta_\bftheta
\end{bmatrix}
\!,
\bfA^2 \!=\! 
\begin{bmatrix} 
0 & 0 & \bfP_v\bfV_\bftheta \\ 
0 & 0 & \bfV_\bftheta\Theta_\bftheta \\
0 & 0 & \Theta_\bftheta^2
\end{bmatrix}
\!,
\bfA^3 \!=\! 
\begin{bmatrix} 
0 & 0 & \bfP_v\bfV_\bftheta\Theta_\bftheta \\ 
0 & 0 & \bfV_\bftheta\Theta_\bftheta^2 \\
0 & 0 & \Theta_\bftheta^3
\end{bmatrix}
\!,
\bfA^4 \!=\! 
\begin{bmatrix} 
0 & 0 & \bfP_v\bfV_\bftheta\Theta_\bftheta^2 \\ 
0 & 0 & \bfV_\bftheta\Theta_\bftheta^3 \\
0 & 0 & \Theta_\bftheta^4
\end{bmatrix}
\!,
\end{equation}
from which it is now visible that, for $k>1$,
\begin{equation}
\bfA^{k>1} = 
\begin{bmatrix} 
0 & 0 & \bfP_v\bfV_\bftheta\Theta_\bftheta^{k-2} \\ 
0 & 0 & \bfV_\bftheta\Theta_\bftheta^{k-1} \\
0 & 0 & \Theta_\bftheta^k
\end{bmatrix}
\end{equation}

We can observe that the terms in the increasing powers of $\bfA$ have a fixed part and an increasing power of $\Theta_\bftheta$. 
These powers can lead to closed form solutions, as in the previous section.

Let us partition the matrix $\Phi$ as follows,
\begin{equation}
\Phi = \begin{bmatrix}
\bfI & \Phi_{\bfp\bfv} & \Phi_{\bfp\bftheta} \\
0 & \bfI & \Phi_{\bfv\bftheta} \\
0 & 0 & \Phi_{\bftheta\bftheta}
\end{bmatrix}~,
\end{equation}
and let us advance step by step, exploring all the non-zero blocks of $\Phi$ one by one. 

\paragraph{Trivial diagonal terms}
Starting by the two upper terms in the diagonal, they are the identity as shown. 

\paragraph{Rotational diagonal term}
Next is the rotational diagonal term $\Phi_{\bftheta\bftheta}$, relating the new angular error to the old angular error. 
Writing the full Taylor series for this term leads to
\begin{equation}
\Phi_{\bftheta\bftheta} = 
\sum_{k=0}^\infty \frac{1}{k!}\Theta_\bftheta^k \Dt^k = \sum_{k=0}^\infty \frac{1}{k!}\hatx{-\bfomega}^k \Dt^k ~, \label{equ:thetaThetaSeries}
\end{equation}
which corresponds, as we have seen in the previous section, to our well-known rotation matrix,
\begin{equation}
\eqbox{\Phi_{\bftheta\bftheta} = \bfR\{\bfomega\Dt\}\tr}~.
\end{equation}

\paragraph{Position-vs-velocity term}
The simplest off-diagonal term is $\Phi_{\bfp\bfv}$, which is
\begin{equation}
\eqbox{\Phi_{\bfp\bfv} = \bfP_\bfv\Dt = \bfI\Dt}~.
\end{equation}

\paragraph{Velocity-vs-angle term}
Let us now move to the term $\Phi_{\bfv, \bftheta}$, by writing its series,
\begin{equation}
\Phi_{\bfv\bftheta} = \bfV_\bftheta\Dt 
+ \frac{1}{2}\bfV_\bftheta\Theta_\bftheta\Dt^2 
+ \frac{1}{3!}\bfV_\bftheta\Theta_\bftheta^2\Dt^3
+\cdots
\end{equation}
\begin{equation}
\Phi_{\bfv\bftheta} = \Dt\bfV_\bftheta( \bfI 
+ \frac{1}{2}\Theta_\bftheta\Dt 
+ \frac{1}{3!}\Theta_\bftheta^2\Dt^2
+\cdots)
\end{equation}
which reduces to
\begin{equation}
\Phi_{\bfv\bftheta} = \Dt\bfV_\bftheta\left( \bfI 
+ \sum_{k\ge1}\frac{(\Theta_\bftheta\Dt)^k}{(k+1)!} 
\right)
\end{equation}

At this point we have two options. 
We can truncate the series at the first significant term, obtaining $\Phi_{\bfv\bftheta} = \bfV_\bftheta\Dt$, but this would not be a closed-form. 
See next section for results using this simplified method.
Alternatively, let us factor $\bfV_\bftheta$ out and write
\begin{equation}
\Phi_{\bfv\bftheta} = \bfV_\bftheta\Sigma_1
\end{equation}
with
\begin{equation}
\Sigma_1 = \bfI\Dt 
+ \frac{1}{2}\Theta_\bftheta\Dt^2 
+ \frac{1}{3!}\Theta_\bftheta^2\Dt^3
+\cdots~.
\end{equation}
The series $\Sigma_1$ ressembles the series we wrote for $\Phi_{\bftheta\bftheta}$, \eqRef{equ:thetaThetaSeries}, with two exceptions:
\begin{itemize}
\item The powers of $\Theta_\bftheta$ in $\Sigma_1$ do not match with the rational coefficients $\tfrac{1}{k!}$ and with the powers of 
$\Dt$. 
In fact, we remark here that the subindex ``1" in $\Sigma_1$ denotes the fact that one power of $\Theta_\bftheta$ is missing in each of the members.

\item Some terms at the start of the series are missing. 
Again, the subindex ``1" indicates that one such term is missing.
\end{itemize}

The first issue may be solved by applying \eqRef{equ:prop2} to \eqRef{equ:ThetaThetaDef}, which yields the identity
\begin{equation}
\Theta_\bftheta
 = \frac{\hatx\bfomega^3}{\norm\bfomega^2}
 = \frac{-\Theta_\bftheta^3 }{\norm\bfomega^2} ~. \label{equ:skewPowerAugment2}
\end{equation}
This expression allows us to increase the exponents of $\Theta_\bftheta$ in the series by two, and write, if $\bfomega \neq 0$,
\begin{equation}
\Sigma_1 =
  \bfI\Dt 
- \frac{\Theta_\bftheta}{\norm\bfomega^2}\left(\frac12\Theta_\bftheta^2\Dt^2 + \frac1{3!}\Theta_\bftheta^3\Dt^3 + \dots \right) ~,
\end{equation}
and $\Sigma_1=\bfI\Dt$ otherwise. 
All the powers in the new series match with the correct coefficients. 
Of course, and as indicated before, some terms are missing. 
This second issue can be solved by adding and substracting the missing terms, and substituting the full series by its closed form. 
We obtain
\begin{equation}
\Sigma_1 =
  \bfI\Dt 
- \frac{\Theta_\bftheta}{\norm\bfomega^2}\left(\bfR\{\bfomega\Dt\}\tr - \bfI - \Theta_\bftheta\Dt \right) ~,
\end{equation}
which is a closed-form solution valid if $\bfomega\neq 0$.  
Therefore we can finally write
\begin{subequations}
\begin{empheq}
 [left={\Phi_{\bfv\bftheta}=\empheqlbrace},box=\fbox]
 {alignat=2}
 &-\bfR\hatx{\bfa}\Dt   &&  \bfomega\to 0 \\
 &-\bfR\hatx{\bfa} \left(\bfI\Dt 
+ \frac{\hatx\bfomega}{\norm\bfomega^2}\left(\bfR\{\bfomega\Dt\}\tr - \bfI + \hatx\bfomega\Dt \right)
\right)
 &\quad & \bfomega \neq 0
\end{empheq}
\end{subequations}

\paragraph{Position-vs-angle term}
Let us finally board the term $\Phi_{\bfp\bftheta}$. 
Its Taylor series is,
\begin{equation}
\Phi_{\bfp\bftheta} = 
  \frac{1}{2 }\bfP_\bfv\bfV_\bftheta\Dt^2
+ \frac{1}{3!}\bfP_\bfv\bfV_\bftheta\Theta_\bftheta\Dt^3 
+ \frac{1}{4!}\bfP_\bfv\bfV_\bftheta\Theta_\bftheta^2\Dt^4
+\cdots
\end{equation}
We factor out the constant terms and get,
\begin{equation}
\Phi_{\bfp\bftheta} = \bfP_\bfv\bfV_\bftheta\ \Sigma_2~,
\end{equation}
with
\begin{equation}
\Sigma_2 = 
  \frac{1}{2 }\bfI\Dt^2
+ \frac{1}{3!}\Theta_\bftheta\Dt^3 
+ \frac{1}{4!}\Theta_\bftheta^2\Dt^4
+ \cdots ~.
\end{equation}
where we note the subindex ``2" in $\Sigma_2$ admits the following interpretation:
\begin{itemize}
\item Two powers of $\Theta_\bftheta$ are missing in each term of the series,
\item The first two terms of the series are missing.
\end{itemize}

Again, we use \eqRef{equ:skewPowerAugment2} to increase the exponents of $\Theta_\bftheta$, yielding
\begin{equation}
\Sigma_2 = 
  \frac{1}{2}\bfI \Dt^2
- \frac{1}{\norm\bfomega^2}\left(
  \frac{1}{3!}\Theta_\bftheta^3\Dt^3 
+ \frac{1}{4!}\Theta_\bftheta^4\Dt^4
+ \cdots \right)~.
\end{equation}
We substitute the incomplete series by its closed form,
\begin{equation}
\Sigma_2 =
  \frac12\bfI \Dt^2
- \frac{1}{\norm\bfomega^2}
\left(
  \bfR\{\bfomega\Dt\}\tr 
	- \bfI 
	- \Theta_\bftheta\Dt
	- \frac{1}{2}\Theta_\bftheta^2\Dt^2
\right)~,
\end{equation}
which leads to the final result
\begin{subequations}
\begin{empheq}[left={\Phi_{\bfp\bftheta}=\empheqlbrace},box=\fbox]{alignat=2}
 & -\bfR\hatx{\bfa} \frac{\Dt^2}{2} && \bfomega\to 0  \\
 & -\bfR\hatx{\bfa} 
\left(
\frac12\bfI \Dt^2
- \frac{1}{\norm\bfomega^2}
\left(
  \bfR\{\bfomega\Dt\}\tr 
	- \sum_{k=0}^2\frac{(-\hatx\bfomega\Dt)^k}{k!}
\right)\right)
 &\quad &\omega\neq 0 
\end{empheq}
\end{subequations}

\subsection{Full IMU example}
\label{sec:closedFormFullImu}

In order to give means to generalize the methods exposed in the simplified IMU example, we need to examine the full IMU case from a little closer.

Consider the full IMU system \eqRef{equ:efull}, which can be posed as
\begin{equation}
\dot{\delta\bfx} = \bfA\delta\bfx + \bfB\bfw~,
\end{equation}
whose discrete-time integration requires the transition matrix 
\begin{equation}
\Phi=\sum_{k=0}^\infty\frac{1}{k!}\bfA^k\Dt^k = \bfI + \bfA\Dt + \frac12\bfA^2\Dt^2 + \dots~, \label{equ:tranmatseries}
\end{equation}
which we wish to compute. 
The dynamic matrix $\bfA$ is block-sparse, and its blocks can be easily determined by examining the original equations \eqRef{equ:efull},

\begin{equation}
\bfA = \begin{bmatrix}
  0& \bfP_\bfv&  0&  0&  0&  0\\
  0&  0& \bfV_\bftheta& \bfV_\bfa&  0& \bfV_\bfg\\
  0&  0& \Theta_\bftheta&  0& \Theta_\bfomega
&  0\\
  0&  0&  0&  0&  0&  0\\
  0&  0&  0&  0&  0&  0\\
  0&  0&  0&  0&  0&  0
\end{bmatrix}~. \label{equ:FullIMU_Fmat}
\end{equation}

As we did before, let us write a few powers of $\bfA$,
\begin{align*}
\bfA^2 &= \begin{bmatrix}
  ~~0~~& ~~0~~& ~\bfP_\bfv \bfV_\bftheta~& \bfP_\bfv \bfV_\bfa&          0& \bfP_\bfv \bfV_\bfg\\
  0& 0& ~\bfV_\bftheta \Theta_\bftheta~&    0&      ~\bfV_\bftheta \Theta_\bfomega~&    0\\
  0& 0&  \Theta_\bftheta^2&     0& \Theta_\bftheta \Theta_\bfomega&     0\\
  0& 0&     0&     0&          0&     0\\
  0& 0&     0&     0&          0&     0\\
  0& 0&     0&     0&          0&     0
\end{bmatrix}\\
\bfA^3 &= \begin{bmatrix}
  ~~0~~& ~~0~~& \bfP_\bfv \bfV_\bftheta \Theta_\bftheta& ~0~&             ~\bfP_\bfv \bfV_\bftheta \Theta_\bfomega~& ~~0~~\\
  0& 0&  \bfV_\bftheta \Theta_\bftheta^2&    0&     \bfV_\bftheta \Theta_\bftheta \Theta_\bfomega&    0\\
  0& 0&     \Theta_\bftheta^3&     0& \Theta_\bftheta^2\Theta_\bfomega&     0\\
  0& 0&        0&     0&                    0&     0\\
  0& 0&        0&     0&                    0&     0\\
  0& 0&        0&     0&                    0&     0
\end{bmatrix}\\
\bfA^4 &= \begin{bmatrix}
  ~~0~~& ~~0~~& \bfP_\bfv \bfV_\bftheta \Theta_\bftheta^2 & ~0~&         \bfP_\bfv \bfV_\bftheta\Theta_\bftheta \Theta_\bfomega& ~~0~~\\
  0& 0&    \bfV_\bftheta \Theta_\bftheta^3&    0&  \bfV_\bftheta \Theta_\bftheta^2 \Theta_\bfomega&    0\\
  0& 0&       \Theta_\bftheta^4&     0& \Theta_\bftheta^3\Theta_\bfomega&     0\\
  0& 0&          0&     0&                              0&     0\\
  0& 0&          0&     0&                              0&     0\\
  0& 0&          0&     0&                              0&     0
\end{bmatrix}~.%
\end{align*}%
Basically, we observe the following,

\begin{itemize}
\item
The only term in the diagonal of $\bfA$, the rotational term $\Theta_\bftheta$, propagates right and up in the sequence of powers $\bfA^k$. 
All terms not affected by this propagation vanish. 
This propagation afects the structure of the sequence $\{\bfA^k\}$ in the three following aspects:
\item 
The sparsity of the powers of $\bfA$ is stabilized after the 3rd power. 
That is to say, there are no more non-zero blocks appearing or vanishing for powers of $\bfA$ higher than 3.
\item 
The upper-left $3\times 3$ block, corresponding to the simplified IMU model in the previous example, has not changed with respect to that example. 
Therefore, its closed-form solution developed before holds.
\item 
The terms related to the gyrometer bias error (those of the fifth column) introduce a similar series of powers of $\Theta_\bftheta$, which can be solved with the same techniques we used in the simplified example. 
\end{itemize}

We are interested at this point in finding a generalized method to board the construction of the closed-form elements of the transition matrix $\Phi$. 
Let us recall the remarks we made about the series $\Sigma_1$ and $\Sigma_2$,
\begin{itemize}
\item The subindex coincides with the lacking powers of $\Theta_\bftheta$ in each of the members of the series.
\item The subindex coincides with the number of terms missing at the beginning of the series.
\end{itemize}

Taking care of these properties, let us introduce the series $\Sigma_n(\bfX,y)$, defined by\footnote{Note that, being $\bfX$ a square matrix that is not necessarily invertible (as it is the case for $\bfX=\Theta_\bftheta$), we are not allowed to rearrange the definition of $\Sigma_n$ with $\Sigma_n=\bfX^{-n}\sum_{k=n}^\infty\frac{1}{k!}(y\bfX)^{k}$.}
\begin{equation}
\Sigma_n(\bfX,y) \triangleq \sum_{k=n}^\infty \frac{1}{k!}\bfX^{k-n}y^k = \sum_{k=0}^\infty \frac{1}{(k+n)!}\bfX^{k}y^{\,k+n}= y^n\sum_{k=0}^\infty \frac{1}{(k+n)!}\bfX^k y^k
\end{equation}
in which the sum starts at term $n$ and the terms lack $n$ powers of the matrix $\bfX$. 
It follows immediately that $\Sigma_1$ and $\Sigma_2$ respond to 
\begin{equation}
\Sigma_n=\Sigma_n(\Theta_\bftheta,\Dt)~,
\end{equation}
and that $\Sigma_0 = \bfR\{\bfomega\Dt\}\tr$. 
We can now write the transition matrix \eqRef{equ:tranmatseries} as a function of these series,
\begin{equation}
\Phi = \begin{bmatrix}
\bfI & \bfP_\bfv\Dt & \bfP_\bfv\bfV_\bftheta\Sigma_2 &  \tfrac12\bfP_\bfv\bfV_\bfa\Dt^2 & \bfP_\bfv\bfV_\bftheta\Sigma_3\bftheta_\bfomega & \tfrac12\bfP_\bfv\bfV_\bfg\Dt^2 \\
0 & \bfI & \bfV_\bftheta\Sigma_1 &  \bfV_\bfa\Dt & \bfV_\bftheta\Sigma_2\bftheta_\bfomega & \bfV_\bfg\Dt \\
0 & 0 & \Sigma_0 &  0 & \Sigma_1\bftheta_\bfomega & 0 \\
0 & 0 & 0 & \bfI & 0 & 0 \\
0 & 0 & 0 & 0 & \bfI & 0 \\
0 & 0 & 0 & 0 & 0 & \bfI \\
\end{bmatrix}~.\label{equ:fullIMUtranmat}
\end{equation}�

Our problem has now derived to the problem of finding a general, closed-form expression for $\Sigma_n$. 
Let us observe the closed-form results we have obtained so far,
\begin{align}
\Sigma_0 &= \bfR\{\bfomega\Dt\}\tr \\
\Sigma_1 &=
  \bfI\Dt 
- \frac{\Theta_\bftheta}{\norm\bfomega^2}\left(\bfR\{\bfomega\Dt\}\tr - \bfI - \Theta_\bftheta\Dt \right) \\
\Sigma_2 &= 
  \frac12 \bfI\Dt^2
- \frac{1}{\norm\bfomega^2}\left(
  \bfR\{\bfomega\Dt\}\tr - \bfI - \Theta_\bftheta\Dt - \frac12\Theta_\bftheta^2\Dt^2
 \right)~.
\end{align}%

In order to develop $\Sigma_3$, we need to apply the identity \eqRef{equ:skewPowerAugment2} twice (because we lack three powers, and each application of \eqRef{equ:skewPowerAugment2} increases this number by only two), getting
\begin{equation}
\Sigma_3 = 
  \frac1{3!}\bfI\Dt^3 
+ \frac{\Theta_\bftheta}{\norm\bfomega^4}
	\left(
		  \frac1{4!}\Theta_\bftheta^4\Dt^4
		+ \frac1{5!}\Theta_\bftheta^5\Dt^5
		+ \dots
	\right)~,
\end{equation}%
which leads to
\begin{equation}
\Sigma_3 = 
  \frac1{3!}\bfI\Dt^3 
+ \frac{\Theta_\bftheta}{\norm\bfomega^4}
	\left(
		\bfR\{\bfomega\Dt\}\tr - \bfI - \Theta_\bftheta\Dt - \frac12\Theta_\bftheta^2\Dt^2 - \frac1{3!}\Theta_\bftheta^3\Dt^3
	\right)~.
\end{equation}%
By careful inspection of the series $\Sigma_0\dots\Sigma_3$, we can now derive a general, closed-form expression for $\Sigma_n$, as follows,
\begin{subequations}
\begin{empheq}[left={\Sigma_n=\empheqlbrace},box=\widefbox]{alignat=2}
 & \frac1{n!}\bfI\Dt^n && \bfomega \to 0 \\
 & \bfR\{\bfomega\Dt\}\tr && n = 0 \\
 &  \frac1{n!}\bfI\Dt^n
- \frac{(-1)^{\tfrac{n+1}{2}}\hatx\bfomega}{\norm\bfomega^{n+1}}
	\left(
		\bfR\{\bfomega\Dt\}\tr - \sum_{k=0}^n \frac{(-\hatx\bfomega\Dt)^k}{k!}
	\right) && n \text{ odd}  \\
 &   \frac1{n!}\bfI\Dt^n
+ \frac{(-1)^{\tfrac{n}{2}}}{\norm\bfomega^n}\left(
  \bfR\{\bfomega\Dt\}\tr - \sum_{k=0}^n \frac{(-\hatx\bfomega\Dt)^k}{k!}
	\right) &\quad & n \text{ even} 
\end{empheq}
\end{subequations}

The final result for the transition matrix $\Phi$ follows immediately by substituting the appropriate values of $\Sigma_n,\ n\in\{0,1,2,3\}$, in the corresponding positions of \eqRef{equ:fullIMUtranmat}.

It might be worth noticing that the series now appearing in these new expressions of $\Sigma_n$ have a finite number of terms, and thus that they can be effectively computed. 
That is to say, the expression of $\Sigma_n$ is a closed form as long as $n<\infty$, which is always the case. 
For the current example, we have $n\leq 3$ as can be observed in \eqRef{equ:fullIMUtranmat}.


\section{Approximate methods using truncated series}

In the previous section, we have devised closed-form expressions for the transition matrix of complex, IMU-driven dynamic systems written in their linearized, error-state form $\dot{\delta\bfx}=\bfA\delta\bfx$. 
Closed form expressions may always be of interest, but it is unclear up to which point we should be worried about high order errors and their impact on the performance of real algorithms. 
This remark is particularly relevant in systems where IMU integration errors are observed (and thus compensated for) at relatively high rates, such as visual-inertial or GPS-inertial fusion schemes.

In this section we devise methods for approximating the transition matrix. 
They start from the same assumption that the transition matrix can be expressed as a Taylor series, and then truncate these series at the most significant terms. 
This truncation can be done system-wise, or block-wise.

\subsection{System-wise truncation}

\subsubsection{First order truncation: the finite differences method}

A typical, widely used integration method for systems of the type
\begin{equation*}
\dot\bfx = f(t,\bfx) 
\end{equation*}
is based on the finite-differences method for the computation of the derivative, \ie,
\begin{equation}
\dot\bfx \triangleq \lim_{\delta t\to 0} \frac{\bfx(t+\delta t)-\bfx(t)}{\delta t} \approx \frac{\bfx_{n+1}-\bfx_n}{\Dt}ª.
\end{equation}
This leads immediately to
\begin{equation}
\bfx_{n+1} \approx \bfx_n + \Dt\,f(t_n,\bfx_n)~,
\end{equation}
which is precisely the Euler method. 
Linearization of the function $f()$ at the beginning of the integration interval leads to
\begin{subequations}
\begin{equation}
\bfx_{n+1} \approx \bfx_n + \Dt\,\bfA\,\bfx_n~,
\end{equation}
where $ \bfA\triangleq\dpar{f}{\bfx}{(t_n,\bfx_n)}$ is a Jacobian matrix. 
This is strictly equivalent to writing the exponential solution to the linearized differential equation and truncating the series at the linear term (\ie, the following relation is identical to the previous one),
\begin{equation}
\bfx_{n+1} = e^{\bfA\Dt}\bfx_n \approx (\bfI + \Dt\,\bfA)\,\bfx_n~.
\end{equation}
\end{subequations}
This means that the Euler method (\appRef{sec:Euler}), the finite-differences method, and the first-order system-wise Taylor truncation method, are all the same. 
We get the approximate transition matrix,
\begin{equation}
\eqbox{\Phi \approx \bfI+\Dt\bfA}~.
\end{equation}

For the simplified IMU example of \secRef{sec:IMUexample}, the finite-differences method results in the approximated transition matrix
\begin{equation}
\Phi \approx \begin{bmatrix}
\bfI & \bfI\Dt & 0 \\
0 & \bfI & -\bfR\hatx{\bfa}\Dt \\
0 & 0 & \bfI-\hatx{\bfomega\Dt}
\end{bmatrix}~.
\end{equation}
However, we already know from \secRef{sec:ClosedFormAngle} that the rotational term has a compact, closed-form solution, $\Phi_{\bftheta\bftheta}=\bfR(\bfomega \Dt)\tr$. 
It is convenient to re-write the transition matrix according to it,
\begin{equation}
\Phi \approx \begin{bmatrix}
\bfI & \bfI\Dt & 0 \\
0 & \bfI & -\bfR\hatx{\bfa}\Dt \\
0 & 0 & \bfR\{\bfomega\Dt\}\tr
\end{bmatrix}~.
\end{equation}

\subsubsection{N-th order truncation}

Truncating at higher orders will increase the precision of the approximated transition matrix. 
A particularly interesting order of truncation is that which exploits the sparsity of the result to its maximum. 
In other words, the order after which no new non-zero terms appear.

For the simplified IMU example of \secRef{sec:IMUexample}, this order is 2, resulting in
\begin{equation}
{\bf\Phi} 
\approx \bfI + \bfA\Dt + \frac12\bfA^2\Dt^2
= \begin{bmatrix}
\bfI & \bfI\Dt & -\frac12\bfR\hatx{\bfa}\Dt^2                           \\
0    & \bfI    &        -\bfR\hatx{\bfa}(\bfI - \frac12\hatx{\bfomega}\Dt)\Dt  \\
0    & 0       &         \bfR\{\bfomega\Dt\}\tr
\end{bmatrix}~.
\end{equation}

In the full IMU example of \secRef{sec:closedFormFullImu}, the is order 3, resulting in
\begin{equation}
{\bf\Phi} \approx \bfI + \bfA\Dt + \frac12\bfA^2\Dt^2 + \frac16\bfA^3\Dt^3~,
\end{equation}
whose full form is not given here for space reasons. 
The reader may consult the expressions of $\bfA$, $\bfA^2$ and $\bfA^3$ in \secRef{sec:closedFormFullImu}.

\subsection{Block-wise truncation}
\label{sec:BlockWiseTruncation}
A fairly good approximation to the closed forms previously explained results from truncating the Taylor series of each block of the transition matrix at the first significant term. 
That is, instead of truncating the series in full powers of $\bfA$, as we have just made above, we regard each block individually. 
Therefore, truncation needs to be analyzed in a per-block basis. 
We explore it with two examples.

For the simplified IMU example of \secRef{sec:IMUexample}, we had series $\Sigma_1$ and $\Sigma_2$, which we can truncate as follows
\begin{empheq}{alignat=6}
\Sigma_1 &= 
   \bfI\Dt 
&+ \frac{1}{2}\Theta_\bftheta\Dt^2 
&+ \cdots 
&{}
&\approx \bfI\Dt \\
\Sigma_2 &= 
{}
&  \frac{1}{2 }\bfI\Dt^2
&+ \frac{1}{3!}\Theta_\bftheta\Dt^3 
&+ \cdots 
&\approx  \  \frac{1}{2}\bfI\Dt^2 
& ~.
\end{empheq}
This leads to the approximate transition matrix
\begin{equation}
\Phi \approx \begin{bmatrix}
\bfI & \bfI\Dt & -\frac12\bfR\hatx{\bfa} \Dt^2 \\
0 & \bfI & -\bfR\hatx{\bfa} \Dt \\
0 & 0 & \bfR(\bfomega \Dt)\tr
\end{bmatrix}~,
\end{equation}
which is more accurate than the one in the system-wide first-order truncation above (because of the upper-right term which has now appeared), yet it remains easy to obtain and compute, especially when compared to the closed forms developed in \secRef{sec:ClosedFormInt}. 
Again, observe that we have taken the closed-form for the lowest term, \ie, $\Phi_{\bftheta\bftheta}=\bfR(\bfomega \Dt)\tr$.

In the general case, it suffices to approximate each $\Sigma_n$ except $\Sigma_0$ by the first term of its series, \ie,
\begin{equation}
\eqbox{
\Sigma_0=\bfR\{\bfomega\Dt\}\tr~,\qquad \Sigma_{n>0} \approx \frac{1}{n!}\bfI\Dt^n}  ~.
\end{equation}%

For the full IMU example, feeding the previous $\Sigma_n$ into \eqRef{equ:fullIMUtranmat} yields the approximated transition matrix,
\begin{equation}
\Phi \approx \begin{bmatrix}
\bfI & \bfI\Dt & -\frac{1}{2}\bfR\hatx{\bfa}\Dt^2 &  -\tfrac12\bfR\Dt^2 & \frac{1}{3!}\bfR\hatx{\bfa}\Dt^3 & \tfrac12\bfI\Dt^2 \\
0 & \bfI & -\bfR\hatx{\bfa}\Dt &  -\bfR\Dt & \frac{1}{2}\bfR\hatx{\bfa}\Dt^2 & \bfI\Dt \\
0 & 0 & \bfR\{\bfomega\Dt\}\tr &  0 & -\bfI\Dt & 0 \\
0 & 0 & 0 & \bfI & 0 & 0 \\
0 & 0 & 0 & 0 & \bfI & 0 \\
0 & 0 & 0 & 0 & 0 & \bfI \\
\end{bmatrix}\label{equ:FullIMU_PhiTrunc}
\end{equation}
with (see \eqRef{equ:efull})
\begin{equation*}
\bfa=\bfa_m-\bfa_b~,\quad\bfomega=\bfomega_m-\bfomega_b~,\quad\bfR=\bfR\{\bfq\}~,
\end{equation*}
and where we have substituted the matrix blocks by their appropriate values (see also \eqRef{equ:efull}),
\begin{equation*}
\bfP_\bfv=\bfI~, \quad 
\bfV_\bftheta=-\bfR\hatx{\bfa}~, \quad 
\bfV_\bfa=-\bfR~, \quad
\bfV_\bfg=\bfI~, \quad
\Theta_\bftheta=-\hatx\bfomega~, \quad
\Theta_\bfomega=-\bfI 
\end{equation*}

A slight simplification of this method is to limit each block in the matrix to a certain maximum order $n$. 
For $n=1$ we have,
\begin{equation}
\Phi \approx \begin{bmatrix}
\bfI & \bfI\Dt & 0 & 0 & 0 & 0 \\
0 & \bfI & -\bfR\hatx{\bfa}\Dt &  -\bfR\Dt & 0 & \bfI\Dt \\
0 & 0 & \bfR\{\bfomega\Dt\}\tr &  0 & -\bfI\Dt & 0 \\
0 & 0 & 0 & \bfI & 0 & 0 \\
0 & 0 & 0 & 0 & \bfI & 0 \\
0 & 0 & 0 & 0 & 0 & \bfI \\
\end{bmatrix}~,\label{equ:FullIMU_PhiTrunc1}
\end{equation}
which is the Euler method, whereas for $n=2$,
\begin{equation}
\Phi \approx \begin{bmatrix}
\bfI & \bfI\Dt & -\frac{1}{2}\bfR\hatx{\bfa}\Dt^2 &  -\tfrac12\bfR\Dt^2 & 0 & \tfrac12\bfI\Dt^2 \\
0 & \bfI & -\bfR\hatx{\bfa}\Dt &  -\bfR\Dt & \frac{1}{2}\bfR\hatx{\bfa}\Dt^2 & \bfI\Dt \\
0 & 0 & \bfR\{\bfomega\Dt\}\tr &  0 & -\bfI\Dt & 0 \\
0 & 0 & 0 & \bfI & 0 & 0 \\
0 & 0 & 0 & 0 & \bfI & 0 \\
0 & 0 & 0 & 0 & 0 & \bfI \\
\end{bmatrix}~.\label{equ:FullIMU_PhiTrunc2}
\end{equation}
For $n\ge3$ we have the full form \eqRef{equ:FullIMU_PhiTrunc}.


\section{The transition matrix via Runge-Kutta integration}
\label{sec:TranMatRK}

Still another way to approximate the transition matrix is to use Runge-Kutta integration. 
This might be necessary in cases where the dynamic matrix $\bfA$ cannot be considered constant along the integration interval, \ie,
\begin{equation}
\dot\bfx(t)=\bfA(t)\bfx(t)~. \label{equ:RKtran1}
\end{equation}

Let us rewrite the following two relations defining the same system in continuous- and discrete-time. 
They involve the dynamic matrix $\bfA$ and the transition matrix $\Phi$,
\begin{align}
\dot\bfx(t) &= \bfA(t)\tdot\bfx(t) \label{equ:ode_dxFx} \\
\bfx(t_n+\tau) &= \Phi(t_n+\tau|t_n)\tdot\bfx(t_n)~.
\end{align}
These equations allow us to develop $\dot\bfx(t_n+\tau)$ in two ways as follows (left and right developments, please note the tiny dots indicating the time-derivatives),
\begin{align}
\dot{(\Phi(t_n+\tau|t_n)\bfx(t_n))} =& \eqbox{\dot\bfx(t_n+\tau)} =  \bfA(t_n+\tau)\bfx(t_n+\tau) \nonumber \\
\dot\Phi(t_n+\tau|t_n)\bfx(t_n)+\Phi(t_n+\tau|t_n)\dot\bfx(t_n) =&~~~~~~~~~~~~~~~=  \bfA(t_n+\tau)\Phi(t_n+\tau|t_n)\bfx(t_n) \nonumber \\
\dot\Phi(t_n+\tau|t_n)\bfx(t_n) =&& \label{equ:last}
\end{align}%
Here, \eqRef{equ:last} comes from noticing that $\dot\bfx(t_n)=\dot\bfx_n=0$, because it is a sampled value. 
Then,
\begin{equation}
\dot\Phi(t_n+\tau|t_n) = \bfA(t_n+\tau)\Phi(t_n+\tau|t_n) \label{equ:ode_dPhiFtPhi}
\end{equation}
which is the same ODE as \eqRef{equ:ode_dxFx}, now applied to the transition matrix instead of the state vector. 
%
Mind that, because of the identity $\bfx(t_n)=\Phi_{t_n|t_n}\bfx(t_n)$, the transition matrix at the beginning of the interval, $t=t_n$, is always the identity,
\begin{equation}
\Phi_{t_n|t_n} = \bfI~.
\end{equation}

Using RK4 with $f(t,\Phi(t))=\bfA(t)\Phi(t)$, we have
\begin{equation}
\Phi\triangleq\Phi(t_n+\Dt|t_n) = \bfI + \frac{\Dt}{6}(\bfK_1+2\bfK_2+2\bfK_3+\bfK_4)
\end{equation}
with
\begin{align}
\bfK_1 &= \bfA(t_n) \\
\bfK_2 &= \bfA\Big(t_n + \frac{1}{2}\Dt \Big)\Big(\bfI + \frac{\Dt}{2} \bfK_1 \Big)\\
\bfK_3 &= \bfA\Big(t_n + \frac{1}{2}\Dt \Big)\Big( \bfI + \frac{\Dt}{2} \bfK_2\Big) \\
\bfK_4 &= \bfA\Big(t_n + \Dt \Big)\Big( \bfI + \Dt \tdot \bfK_3\Big) ~.
\end{align}%

\subsection{Error-state example}

Let us consider the error-state Kalman filter for the non-linear, time-varying system
\begin{equation}
\dot\bfx_t(t) = f(t, \bfx_t(t), \bfu(t))
\end{equation}
where $\bfx_t$ denotes the true state, and $\bfu$ is a control input. 
This true state is a composition, denoted by $\oplus$, of a nominal state $\bfx$ and the error state $\delta\bfx$,
\begin{equation}
\bfx_t(t) = \bfx(t) \oplus \delta\bfx(t)
\end{equation}
where the error-state dynamics admits a linear form which is time-varying depending on the nominal state $\bfx$ and the control $\bfu$, \ie,
\begin{equation}
\dot{\delta\bfx} = \bfA(\bfx(t),\bfu(t))\tdot\delta\bfx
\end{equation}
that is, the error-state dynamic matrix in \eqRef{equ:RKtran1} has the form $\bfA(t) = \bfA(\bfx(t),\bfu(t))$. 
The dynamics of the error-state transition matrix can be written,
\begin{equation}
\dot{\Phi}(t_n+\tau|t_n)=\bfA(\bfx(t),\bfu(t))\tdot\Phi(t_n+\tau|t_n)~.
\end{equation}
In order to RK-integrate this equation, we need the values of $\bfx(t)$ and $\bfu(t)$ at the RK evaluation points, which for RK4 are $\{t_n,t_n+\Dt/2,t_n+\Dt\}$. 
Starting by the easy ones, the control inputs $\bfu(t)$ at the evaluation points can be obtained by linear interpolation of the current and last measurements,
\begin{align}
\bfu(t_n) &= \bfu_n\\
\bfu(t_n+\Dt/2) &= \frac{\bfu_n+\bfu_{n+1}}{2} \\
\bfu(t_n+\Dt) &= \bfu_{n+1}
\end{align}%
The nominal state dynamics should be integrated using the best integration practicable. 
For example, using RK4 integration we have,
\begin{align*}
\bfk_1 &= f(\bfx_n,\bfu_n) \nonumber \\
\bfk_2 &= f(\bfx_n+\frac{\Dt}{2}\bfk_1,\frac{\bfu_n+\bfu_{n+1}}{2}) \nonumber\\
\bfk_3 &= f(\bfx_n+\frac{\Dt}{2}\bfk_2,\frac{\bfu_n+\bfu_{n+1}}{2}) \nonumber\\
\bfk_4 &= f(\bfx_n+\Dt\bfk_3,\bfu_{n+1}) \nonumber\\
\bfk &= (\bfk_1+2\bfk_2+2\bfk_3+\bfk_4)/6 ~,
\end{align*}%
which gives us the estimates at the evaluation points,
\begin{align}
\bfx(t_n) &= \bfx_n \\
\bfx(t_n+\Dt/2) &= \bfx_n+\frac{\Dt}{2}\bfk\\
\bfx(t_n+\Dt) &= \bfx_n+\Dt\,\bfk ~.
\end{align}%
We notice here that $\bfx(t_n+\Dt/2)=\frac{\bfx_n+\bfx_{n+1}}{2}$, the same linear interpolation we used for the control. 
This should not be surprising given the linear nature of the RK update.

Whichever the way we obtained the nominal state values, we can now compute the RK4 matrices for the integration of the transition matrix,
\begin{align*}
\bfK_1 &= \bfA(\bfx_n,\bfu_n) \\
\bfK_2 &= \bfA\Big(\bfx_n+\frac{\Dt}{2}\bfk, \frac{\bfu_n+\bfu_{n+1}}{2}\Big)\Big(\bfI + \frac{\Dt}{2} \bfK_1 \Big)\\
\bfK_3 &= \bfA\Big(\bfx_n+\frac{\Dt}{2}\bfk, \frac{\bfu_n+\bfu_{n+1}}{2}\Big)\Big(\bfI + \frac{\Dt}{2} \bfK_2 \Big)\\
\bfK_4 &= \bfA\Big(\bfx_n+\Dt\bfk, \bfu_{n+1}\Big)\Big(\bfI + \Dt \bfK_3 \Big)\\
\bfK &= (\bfK_1+2\bfK_2+2\bfK_3+\bfK_4)/6 
\end{align*}%
which finally lead to,
\begin{equation}
\eqbox{\Phi\triangleq\Phi_{t_n+\Dt|t_n} = \bfI + \Dt\,\bfK}
\end{equation}


\section{Integration of random noise and perturbations}
\label{sec:IntNoise}

We aim now at giving appropriate methods for the integration of random variables within dynamic systems. 
Of course, we cannot integrate unknown random values, but we can integrate their variances and covariances for the sake of uncertainty propagation. 
This is needed in order to establish the covariances matrices in estimators for systems that are of continuous nature (and specified in continuous time) but estimated in a discrete manner.

Consider the continuous-time dynamic system,
\begin{equation}
\dot\bfx = f(\bfx,\bfu,\bfw)~,
\label{equ:contSysWithNoise}
\end{equation}
where $\bfx$ is the state vector, $\bfu$ is a vector of  control signals containing noise $\tilde\bfu$, so that the control measurements are $\bfu_m=\bfu+\tilde\bfu$, and $\bfw$ is a vector of random perturbations. 
Both noise and perturbations are assumed white Gaussian processes, specified by,
\begin{equation}
\tilde\bfu \sim \cN\{0,\bfU^c\} \quad , \quad \bfw^c\sim\cN\{0,\bfW^c\}~,
\end{equation}
where the super-index $\bullet^c$ indicates a continuous-time uncertainty specification, which we want to integrate. 

There exists an important difference between the natures of the noise levels in the control signals, $\tilde\bfu$, and the random perturbations, $\bfw$:
\begin{itemize}
\item 
On discretization, the control signals are sampled at the time instants $n\Dt$, having $\bfu_{m,n}\triangleq\bfu_m(n\Dt)=\bfu(n\Dt)+\tilde\bfu(n\Dt)$. 
The measured part is obviously considered constant over the integration interval, \ie, $\bfu_m(t)=\bfu_{m,n}$, and therefore the noise level at the sampling time $n\Dt$ is also held constant,
\begin{equation}
\tilde\bfu(t)=\tilde\bfu(n\Dt) = \tilde\bfu_n, \quad n\Dt<t<(n+1)\Dt~. \label{equ:uint}
\end{equation}
\item  
The perturbations $\bfw$ are never sampled. 

\end{itemize}

As a consequence, the integration over $\Dt$ of these two stochastic processes differs. 
Let us examine it. 

\bigskip
The continuous-time error-state dynamics \eqRef{equ:contSysWithNoise} can be linearized to
\begin{equation}
\dot{\delta\bfx} = \bfA\delta\bfx + \bfB\tilde\bfu + \bfC \bfw~, \label{equ:contTime}
\end{equation}
with 
\begin{equation}
\bfA\triangleq\pjac{f}{\delta\bfx}{\bfx,\bfu_m}\quad,\quad
\bfB\triangleq\pjac{f}{\tilde\bfu}{\bfx,\bfu_m}\quad,\quad
\bfC\triangleq\pjac{f}{\bfw}{\bfx,\bfu_m}~,
\end{equation}
and integrated over the sampling period $\Dt$, giving,
\begin{align}
\delta\bfx_{n+1} &= \delta\bfx_n 
+ \int_{n\Dt}^{(n+1)\Dt} \left(
\bfA\delta\bfx(\tau) + \bfB\tilde\bfu(\tau) + \bfC\bfw^c(\tau)
\right) d\tau \\
 &= \delta\bfx_n 
+ \int_{n\Dt}^{(n+1)\Dt} \bfA\delta\bfx(\tau)   d\tau
+ \int_{n\Dt}^{(n+1)\Dt} \bfB\tilde\bfu(\tau)   d\tau 
+ \int_{n\Dt}^{(n+1)\Dt} \bfC\bfw^c(\tau) d\tau 
\end{align}
which has three terms of very different nature. 
They can be integrated as follows:
\begin{enumerate}
\item From \appRef{sec:ClosedFormInt} we know that the dynamic part is integrated giving the transition matrix,
\begin{equation}
\delta\bfx_n 
+ \int_{n\Dt}^{(n+1)\Dt} \bfA\delta\bfx(\tau)   d\tau = \Phi\tdot\delta\bfx_n
\end{equation}
where $\Phi=e^{\bfA\Dt}$ can be computed in closed-form or approximated at different levels of accuracy.

\item From \eqRef{equ:uint} we have
\begin{equation}
\int_{n\Dt}^{(n+1)\Dt} \bfB\tilde\bfu(\tau)   d\tau = \bfB\Dt\tilde\bfu_n
\end{equation}
which means that the measurement noise, once sampled, is integrated in a deterministic manner because its behavior inside the integration interval is known.

\item From Probability Theory we know that the integration of continuous white Gaussian noise over a period $\Dt$ produces a discrete white Gaussian impulse $\bfw_n$ described by
\begin{equation}
\bfw_n\triangleq\int_{n\Dt}^{(n+1)\Dt}\bfw(\tau)d\tau 
\quad,\quad
\bfw_n \sim \cN\{0,\bfW\} 
\quad,\quad
 \text{with } \bfW = \bfW^c\Dt
\end{equation}
We obsereve that, contrary to the measurement noise just above, the perturbation does not have a deterministic behavior inside the integration interval, and hence it must be integrated stochastically.

\end{enumerate}%
Therefore, the discrete-time, error-state dynamic system can be written as
\begin{equation}
\delta\bfx_{n+1} = \bfF_\bfx\delta\bfx_n + \bfF_\bfu\tilde\bfu_n + \bfF_\bfw \bfw_n \label{equ:discTime}
\end{equation}
with  transition, control and perturbation matrices given by
\begin{equation}
\bfF_\bfx=\Phi = e^{\bfA\Dt} \quad , \quad \bfF_\bfu=\bfB\Dt \quad , \quad \bfF_\bfw = \bfC \quad , \quad 
\end{equation}
%
with noise and perturbation levels defined by
\begin{equation}
\tilde\bfu_n\sim\cN\{0,\bfU\} \quad , \quad \bfw_n\sim\cN\{0,\bfW\} \label{equ:discMat}
\end{equation}
with 
\begin{equation}
\bfU = \bfU^c \quad , \quad \bfW = \bfW^c\Dt~. \label{equ:discCov}
\end{equation}
\begin{table*}
\renewcommand{\arraystretch}{1.3}
\caption{Effect of integration on system and covariances matrices.}
\centering
\vspace{1ex}
\begin{tabular}{|c|c|c|}
\hline
Description & Continuous time $t$ & Discrete time $n\Dt$\\
\hline
\hline
state & $\dot\bfx=f^c(\bfx,\bfu,\bfw)$ & $\bfx_{n+1} = f(\bfx_n,\bfu_n,\bfw_n)$ \\
error-state & $\dot{\delta\bfx}=\bfA\delta\bfx+\bfB\tilde\bfu+\bfC\bfw$ & $\delta\bfx_{n+1}=\bfF_\bfx\delta\bfx_n+\bfF_\bfu\tilde\bfu_n+\bfF_\bfw\bfw_n$ \\
\hline
system matrix & $\bfA$ & $\bfF_\bfx=\Phi=e^{\bfA\Dt}$ \\
control matrix & $\bfB$ & $\bfF_\bfu=\bfB\Dt$ \\
perturbation matrix & $\bfC$ & $\bfF_\bfw=\bfC$ \\
\hline
control covariance & $\bfU^c$ & $\bfU=\bfU^c$  \\
perturbation covariance & $\bfW^c$ & $\bfW=\bfW^c\Dt$  \\
\hline
\end{tabular}
\label{tab:IntEffects}
\end{table*}%

These results are summarized in \tabRef{tab:IntEffects}. 
The prediction stage of an EKF would propagate the error state's mean and covariances matrix according to
\begin{align}
\hat{\delta\bfx}_{n+1} &= \bfF_\bfx \hat{\delta\bfx}_n \\
\bfP_{n+1} &= \bfF_\bfx\bfP_n\bfF_\bfx\tr + \bfF_\bfu\bfU\bfF_\bfu\tr + \bfF_\bfw\bfW\bfF_\bfw\tr \nonumber\\
&= e^{\bfA\Dt}\bfP_n(e^{\bfA\Dt})\tr + \Dt^2\bfB\bfU^c\bfB\tr + \Dt\bfC\bfW^c\bfC\tr \label{equ:NoisePertCovUpdate}
\end{align}

It is important and illustrative here to observe the different effects of the integration interval, $\Dt$, on the three terms of the covariance update \eqRef{equ:NoisePertCovUpdate}: the dynamic error term is exponential, the measurement error term is quadratic, and the perturbation error term is linear.

\subsection{Noise and perturbation impulses}
\label{sec:pertImpulses}

One is oftentimes confronted (for example when reusing existing code or when interpreting other authors' documents) with EKF prediction equations of a simpler form than those that we used here, namely,
\begin{equation}
\bfP_{n+1} = \bfF_\bfx\bfP_n\bfF_\bfx\tr + \bfQ~.
\end{equation}
This corresponds to the general discrete-time dynamic system,
\begin{equation}
\delta\bfx_{n+1} = \bfF_\bfx\delta\bfx_n+\bfi
\end{equation}
where
\begin{equation}
\bfi \sim \cN\{0,\bfQ\}
\end{equation}
is a vector of random (white, Gaussian) impulses that are directly added to the state vector at time $t_{n+1}$. 
The matrix $\bfQ$ is simply considered the impulses covariances matrix. 
From what we have seen, we should compute this covariances matrix as follows,
\begin{equation}
\bfQ = \Dt^2\,\bfB\,\bfU^c\,\bfB\tr + \Dt\,\bfC\,\bfW^c\,\bfC\tr~.
\end{equation}

In the case where the impulses do not affect the full state, as it is often the case, the matrix $\bfQ$ is not full-diagonal and may contain a significant amount of zeros. 
One can then write the equivalent form
\begin{equation}
\delta\bfx_{n+1} = \bfF_\bfx\,\delta\bfx_n+\bfF_\bfi\,\bfi
\end{equation}
with
\begin{equation}
\bfi \sim \cN\{0,\bfQ_\bfi\}~,
\end{equation}
where the matrix $\bfF_\bfi$ simply maps each individual impulse to the part of the state vector it affects to. 
The associated covariance $\bfQ_\bfi$ is then smaller and full-diagonal. 
Please refer to the next section for an example. 
In such case the ESKF time-update becomes
\begin{align}
\hat{\delta\bfx}_{n+1} &= \bfF_\bfx\,\hat{\delta\bfx}_n \\
\bfP_{n+1} &= \bfF_\bfx\,\bfP_n\,\bfF_\bfx\tr + \bfF_\bfi\,\bfQ_\bfi\,\bfF_\bfi\tr~.
\end{align}

Obviously, all these forms are equivalent, as it can be seen in the following double identity for the general perturbation $\bfQ$,
\begin{equation}
\bfF_\bfi\,\bfQ_\bfi\,\bfF_\bfi\tr = \eqbox{\bfQ} =  \Dt^2\,\bfB\,\bfU^c\,\bfB\tr + \Dt\,\bfC\,\bfW^c\,\bfC\tr~.
\end{equation}

\subsection{Full IMU example}

We study the construction of an error-state Kalman filter for an IMU. 
The error-state system is defined in \eqRef{equ:efull} and involves a nominal state $\bfx$, an error-state $\delta\bfx$, a noisy control  signal $\bfu_m=\bfu+\tilde\bfu$ and a perturbation $\bfw$, specified by,
\begin{equation}
\bfx = \begin{bmatrix}
\bfp \\
\bfv \\
\bfq \\
\bfa_b \\
\bfomega_b \\
\bfg
\end{bmatrix}
\quad , \quad
\delta\bfx = \begin{bmatrix}
\delta\bfp \\
\delta\bfv \\
\delta\bftheta \\
\delta\bfa_b \\
\delta\bfomega_b \\
\delta\bfg
\end{bmatrix}
\quad , \quad
\bfu_m=\begin{bmatrix}
\bfa_m \\ \bfomega_m
\end{bmatrix} 
\quad , \quad
\tilde\bfu=\begin{bmatrix}
\tilde\bfa \\ \tilde\bfomega
\end{bmatrix} 
\quad , \quad
\bfw = \begin{bmatrix}
\bfa_w \\ \bfomega_w
\end{bmatrix}
\end{equation}

In a model of an IMU like the one we are considering throughout this document, the control noise corresponds to the additive noise in the IMU measurements. 
The perturbations affect the biases, thus producing their random-walk behavior. 
The dynamic, control and perturbation matrices are (see  \eqRef{equ:contTime}, \eqRef{equ:FullIMU_Fmat} and \eqRef{equ:efull}),
\begin{equation}
\label{equ:IMU_ABC}
\bfA = \begin{bmatrix}
  0& \bfP_\bfv&  0&  0&  0&  0\\
  0&  0& \bfV_\bftheta& \bfV_\bfa&  0& \bfV_\bfg\\
  0&  0& \Theta_\bftheta&  0& \Theta_\bfomega&  0\\
  0&  0&  0&  0&  0&  0\\
  0&  0&  0&  0&  0&  0\\
  0&  0&  0&  0&  0&  0
\end{bmatrix}
\quad,\quad
\bfB = \begin{bmatrix}
0 & 0 \\
-\bfR & 0 \\
0 & -\bfI \\
0 & 0 \\
0 & 0 \\
0 & 0 \\
\end{bmatrix}
\quad,\quad
\bfC = \begin{bmatrix}
0 & 0 \\
0 & 0 \\
0 & 0 \\
\bfI & 0 \\
0 & \bfI \\
0 & 0 \\
\end{bmatrix}
\end{equation}

In the regular case of IMUs with accelerometer and gyrometer triplets of the same kind on the three axes, noise and perturbations are isotropic. 
Their standard deviations are specified as scalars as follows
\begin{equation}
\sigma_{\tilde\bfa}\ [m/s^2] \quad,\quad \sigma_{\tilde\bfomega} \  [rad/s] \quad,\quad \sigma_{\bfa_w}\ [m/s^2\sqrt{s}] \quad,\quad \sigma_{\bfomega_w}\ [rad/s\sqrt{s}]
\end{equation}
and their covariances matrices are purely diagonal, giving
\begin{equation}
\bfU^c=\begin{bmatrix}
\sigma_{\tilde\bfa}^2\bfI & 0 \\ 0 & \sigma_{\tilde\bfomega}^2\bfI  
\end{bmatrix} \qquad ,\qquad
\bfW^c=\begin{bmatrix}
\sigma_{\bfa_w}^2\bfI & 0 \\ 0 & \sigma_{\bfomega_w}^2\bfI  
\end{bmatrix} ~.
\end{equation}

The system evolves with sampled measures at intervals $\Dt$, following \eqsRef{equ:discTime}{equ:discCov}, where the transition matrix $\bfF_\bfx=\Phi$ can be computed in a number of ways -- see previous appendices. 

\subsubsection{Noise and perturbation impulses}


In the case of a perturbation specification in the form of impulses $\bfi$, we can re-define our system as follows,
\begin{equation}
\delta\bfx_{n+1} = \bfF_\bfx(\bfx_n, \bfu_m)\tdot \delta\bfx_n+\bfF_\bfi\tdot\bfi
\end{equation}
with the nominal-state, error-state, control, and impulses vectors defined by,
\begin{equation}
\bfx=\begin{bmatrix}\bfp\\\bfv\\\bfq\\\bfa_b\\\bfomega_b\\\bfg\end{bmatrix} \quad, \quad
\delta\bfx=\begin{bmatrix}\delta\bfp\\\delta\bfv\\\delta\bftheta\\\delta\bfa_b\\\delta\bfomega_b\\\delta\bfg\end{bmatrix} \quad, \quad
\bfu_m = \begin{bmatrix}
\bfa_m \\
\bfomega_m
\end{bmatrix} 
\quad,  \quad
\bfi = \begin{bmatrix}
\bfv_\bfi \\
\bftheta_\bfi \\
\bfa_\bfi \\
\bfomega_\bfi
\end{bmatrix}~,
\end{equation}
the transition and perturbations matrices defined by,
\begin{equation}
\bfF_\bfx = \Phi = e^{\bfA\Dt}
\qquad,\qquad
\bfF_\bfi 
= \begin{bmatrix}
 0 &  0 & 0 & 0 \\
 \bfI &  0 & 0 & 0 \\
 0 &  \bfI & 0 & 0 \\
 0 &  0 & \bfI & 0 \\
 0 & 0 & 0 & \bfI \\
 0 &  0 & 0 & 0 
\end{bmatrix} ~,
\end{equation}
and the impulses variances specified by
\begin{equation}
\bfi \sim \cN\{0,\bfQ_\bfi\}
\quad,\quad
\bfQ_\bfi = \begin{bmatrix}
\sigma_{\tilde\bfa}^2\Dt^2\bfI &&0&\\
& \sigma_{\tilde\bfomega}^2\Dt^2\bfI &&\\
&& \sigma_{\bfa_w}^2\Dt\bfI &\\
&0&& \sigma_{\bfomega_w}^2\Dt\bfI 
\end{bmatrix}~.
\end{equation}

The trivial specification of $\bfF_\bfi$ may appear surprising given especially that of $\bfB$ in \eqRef{equ:IMU_ABC}. 
What happens is that the errors are defined isotropic in $\bfQ_\bfi$, and therefore $-\bfR\sigma^2\bfI(-\bfR)\tr = \sigma^2\bfI$ and $-\bfI\sigma^2\bfI(-\bfI)\tr = \sigma^2\bfI$, leading to the expression given for $\bfF_\bfi$. 
This is not possible when considering non-isotropic IMUs, where a proper Jacobian $\bfF_\bfi=\begin{bmatrix}
\bfB & \bfC
\end{bmatrix}$ should be used together with a proper specification of $\bfQ_\bfi$.


\bigskip
We can of course use full-state perturbation impulses,
\begin{equation}
\delta\bfx_{n+1} = \bfF_\bfx(\bfx_n, \bfu_m)\tdot \delta\bfx_n+\bfi
\end{equation}
with
\begin{equation}
\bfi = \begin{bmatrix}
0 \\
\bfv_\bfi \\
\bftheta_\bfi \\
\bfa_\bfi \\
\bfomega_\bfi \\
0
\end{bmatrix}
\quad,\quad
\bfi \sim \cN\{0,\bfQ\}
\quad,\quad
\bfQ = \begin{bmatrix}
0 & \\
& \sigma_{\tilde\bfa}^2\Dt^2\bfI &&&0&\\
&& \sigma_{\tilde\bfomega}^2\Dt^2\bfI &&\\
&&& \sigma_{\bfa_w}^2\Dt\bfI &\\
&0&&& \sigma_{\bfomega_w}^2\Dt\bfI \\
&&&&&0
\end{bmatrix}~.
\end{equation}

\bigskip
\bigskip
\bigskip

Bye bye.


\bibliographystyle{apalike}
\bibliography{../LatexTools/bibSLAM,../LatexTools/quaternion,../LatexTools/filtering}

\end{document}